\documentclass[]{fairmeta}

\usepackage{algorithm}
\usepackage{algorithmic}

\usepackage{hyperref}

 \usepackage{multirow}
 \usepackage[table]{xcolor}
\usepackage{xcolor}
\usepackage{mathtools}
\usepackage{microtype}
\usepackage{graphicx}
\usepackage{comment}
\usepackage{multirow}

\usepackage{makecell}
\usepackage{booktabs} 
\usepackage[utf8]{inputenc} 
\usepackage[T1]{fontenc}    
\usepackage{url}            
\usepackage{booktabs}       
\usepackage{amsfonts}       
\usepackage{nicefrac}       
\usepackage{microtype}      
\usepackage{xcolor}         
\usepackage{graphicx}
\usepackage{wrapfig}
\usepackage{adjustbox}
\usepackage{amsmath,amssymb,amsthm} 
\usepackage{tabularx}
\usepackage{float}

\usepackage{xspace}
\usepackage{subcaption} 
\usepackage{caption}

\usepackage{amsmath,amsfonts,bm}

\def\eqref#1{equation~\ref{#1}}

\def\1{\bm{1}}

\def\vtheta{{\bm{\theta}}}

\def\mD{{\bm{D}}}

\def\mW{{\bm{W}}}

\DeclareMathAlphabet{\mathsfit}{\encodingdefault}{\sfdefault}{m}{sl}
\SetMathAlphabet{\mathsfit}{bold}{\encodingdefault}{\sfdefault}{bx}{n}

\def\gC{{\mathcal{C}}}

\def\gE{{\mathcal{E}}}

\newcommand{\R}{\mathbb{R}}

\theoremstyle{plain}
\newtheorem{theorem}{Theorem}[section]
\newtheorem{proposition}[theorem]{Proposition}

\newtheorem{corollary}[theorem]{Corollary}
\theoremstyle{definition}
\newtheorem{definition}[theorem]{Definition}

\theoremstyle{remark}

\definecolor{darkred}{rgb}{0.55, 0.0, 0.0}

\definecolor{findingblue}{HTML}{0064E0}
\definecolor{takeawayred}{HTML}{C80A28}

\newcounter{fcounter}
\newcommand\finding[1]{%
  \refstepcounter{fcounter}\vspace{-2pt}%
  \begin{tcolorbox}[
    enhanced,
    colback=findingblue!10!white,
    frame hidden,            
    borderline west={2pt}{0pt}{findingblue}, 
    boxsep=1pt,
    left=4pt,right=2pt,top=1pt,bottom=1pt,
    arc=0pt, outer arc=0pt,  
  ]
  \noindent{\textbf{\fontsize{10pt}{12pt}\selectfont Finding \arabic{fcounter}}: %
  \fontsize{10pt}{12pt}\selectfont #1}
  \end{tcolorbox}\vspace{-2pt}%
}
\crefname{fcounter}{Finding}{Findings}

\newcounter{kcounter}
\crefname{kcounter}{Takeaway}{Takeaways}

\newcommand{\bopt}{B_\mathrm{opt}}
\newcommand{\bcrit}{B_\mathrm{crit}}

\definecolor{findingpurple}{HTML}{6442D3}

\newtcolorbox{generaltakeaway}[2][]{
  enhanced,
  colback=white,
  colframe=black,
  colbacktitle=takeawayred,
  coltitle=white,
  size=small,
  arc=8pt,
  boxrule=1.5pt,
  titlerule=0pt,
  left=8pt,
  right=8pt,
  top=10pt,
  bottom=8pt,
  fonttitle=\bfseries,
  title={#2},
  attach boxed title to top left={xshift=3pt, yshift=-\tcboxedtitleheight/2},
  boxed title style={
    arc=8pt,
    boxrule=1.5pt,
    colframe=black,
  },
  #1
}

\title{MuLoCo: Muon is a Practical Inner Optimizer for DiLoCo}

\author[1,2,3,*]{Benjamin Th\'erien}
\author[2,4]{Xiaolong Huang}
\author[1]{Aaron Defazio}
\author[2,3]{Irina Rish}
\author[2,4]{Eugene Belilovsky}

\affiliation[1]{FAIR at Meta}
\affiliation[2]{Mila}
\affiliation[3]{Universit\'e de Montr\'eal}
\affiliation[4]{Concordia University}

\contribution[*]{Work done at Meta}

\abstract{DiLoCo is a powerful framework for training large language models (LLMs), enabling larger optimal batch sizes and increased accelerator utilization under networking constraints. However, DiLoCo's performance has been shown to degrade as the number of workers (K) increases~\citep{scalingdiloco}. In this work, we posit that a related but often overlooked factor in DiLoCo’s behavior is the choice of inner optimizer, which shapes the pseudogradient used by the outer optimizer. Given the recent success of Muon relative to AdamW for data parallel (DP) training, we examine how Muon's normalized optimizer steps can affect the pseudogradient's quality. We find that, relative to AdamW, Muon yields more \emph{directionally correct} pseudogradients as the number of workers ($K$) increases. In our experiments pre-training language models, we conduct extensive hyperparameter tuning across 150M, 416M, 914M, 1.76B, and 3.1B models for DiLoCo, MuLoCo, AdamW DP, and Muon DP. Consistently across all scales, we find that with $K\geq1$ workers, MuLoCo (Muon inner optimizer DiLoCo) achieves superior performance to DiLoCo in absolute terms and for $K>2$ it outperforms DiLoCo relative to their data parallel baselines, while being compatible with quantization, streaming, and long synchronization intervals. At $K=1$, we find that MuLoCo can even outperform the data-parallel gold standard while having larger critical batch sizes. Finally, we extrapolate optimal hyperparameters to 15B scale and train a model with each method (six in total) using $K=1$ and $K=16$ workers. We find that $K=16$ MuLoCo nearly matches single-worker performance at this scale, while MuLoCo $K=1$ matches the best performing baseline while using a much larger $16$M token batch size.
\vspace{-10pt}

}

\correspondence{Benjamin Th\'erien at \email{benjamin.therien@umontreal.ca}}

\metadata[Code]{\url{https://github.com/facebookresearch/MuLoCo}}
\metadata[Website]{\url{https://bentherien.github.io/muloco-1}}

\begin{document}

\maketitle

\section{Introduction}

In Large Language Model (LLM) pre-training, it is now well established that scaling training data and model parameters in tandem at a ratio of roughly $20$ tokens per parameter (TPP) is compute optimal~\citep{chinchilla,hagele2024scaling,besiroglu2024chinchilla,dey2023cerebrasgpt,schaeffer2025robustness}. With the largest existing training runs reaching $10-40$T tokens~\citep{qwen3,kimiteam2025kimik2openagentic,deepseekai2024deepseekv3technicalreport} and estimates of available data reaching $300$T~\citep{villalobos2024run}, it is reasonable to assume that the largest compute optimal models trained in the near future will have 2-15T parameters. Efficiently training at this scale is bottlenecked by two requirements of data-parallel training: (a) the need to take a large number of gradient descent steps sequentially, leading to long wall-clock training times, and (b) the need to communicate among all model replicas at each optimization step, leading to decreased compute utilization across tens of thousands of accelerators. A straightforward remedy to (a) is training at large batch sizes. However, at the largest scales, this is likely to exacerbate (b) since there is a limit to the number of accelerators that are tightly interconnected in typical deployments~\citep{nvidia2024dgxsuperpod}.

The motivation for training at large batch sizes is clear: leveraging more accelerators in parallel leads to faster wall-clock training times for the same amount of data. At the largest model scales, computing gradients on a single training example can be expensive. Without large batch training across enough accelerators in parallel, practitioners are required to (1) accumulate gradients, leading to longer step times, or (2) take more gradient descent steps sequentially. The critical batch size (CBS)~\citep{cbsmccandlish,cbsmerril,cbszhang,bergsma2025power} is known as the largest batch size that can be used to achieve performance nearly as strong as the optimal (best performing) batch size. Therefore, training at the CBS with enough devices in parallel can lead to linear speedups in training time with nearly no performance degradation, as long as communication costs do not grow with the number of devices. However, when training at the largest model scales or across poorly interconnected accelerators, this is rarely the case.

\begin{figure*}[t]
    \centering
    \subfloat[\textbf{MuLoCo outperforms DiLoCo in absolute terms and continues to do so at larger worker counts even after normalizing by their respective data-parallel baselines.}]{\includegraphics[width=0.45\linewidth]{
    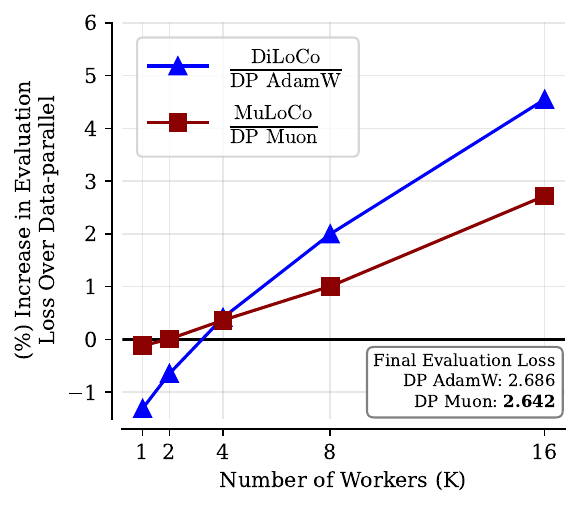
    }}
    \qquad
    \subfloat[\textbf{MuLoCo K=1 acheives a Pareto-Optimal Performance-Time Tradeoff. All datapoints are FLOP-matched.}]{\includegraphics[width=0.45\linewidth]{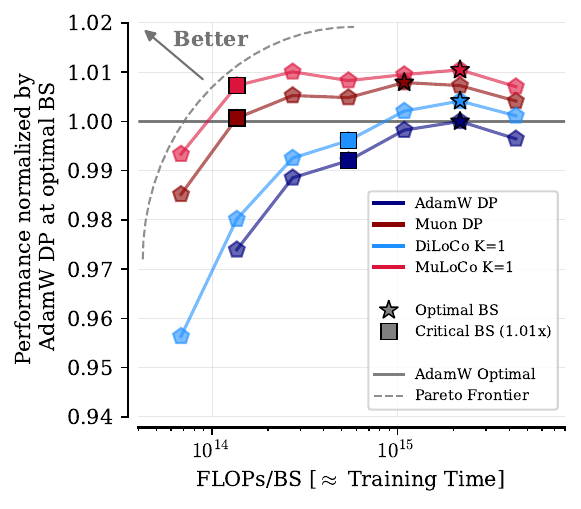}}
    \vspace{-5pt}
\caption{\textbf{Subfigure (a)} reports the percentage increase in loss for DiLoCo and MuLoCo relative to their DP baselines. We observe that the relative performance of each method degrades as K is increased, but that MuLoCo shows clear improvements over DiLoCo relative to their data parallel baselines as workers are increased.
\textbf{ Subfigure (b)} reports the performance of different optimizers in an iso-FLOP setting as a function of FLOPs/Batch Size (a proxy for training time). The optimal batch size ($\bopt$) is the batch size achieving the best performance, while the critical batch size ($\bcrit$) is taken to be the largest batch size achieving a final evaluation loss that is: $L(\bcrit)\leq1.01\times L(\bopt)$. We observe that at the 3.1B scale, our proposed MuLoCo K=1 achieves stronger performance than other optimizers while also being compatible with much larger batch sizes.   }
    \label{fig:worker-scaling}
\end{figure*}

Due to the large communication costs associated with data-parallel (DP) training at large model scales or under bandwidth constraints, a number of communication-efficient training algorithms have been proposed. Most recently, DiLoCo~\citep{douillard2023diloco} has gained attention due to its competitive performance with DP pre-training of LLMs and its larger optimal batch sizes~\citep{scalingdiloco}. DiLoCo independently trains a single model with DP AdamW across $K$ workers in parallel while only communicating across workers every $H$ gradient descent steps. When individual workers span highly interconnected compute pools (e.g., single nodes, racks, adjacent racks, or entire datacenters for the largest models) with significantly faster bandwidth within pools than across them, DiLoCo can lead to significant speedups relative to DP training across the same number of accelerators. However,~\citet{scalingdiloco} shows that the performance of DiLoCo degrades as the number of workers is increased, leading to questions of its utility at large worker counts.

In this work, we seek to address DiLoCo's worker scaling problem while simultaneously increasing its critical batch size. Our core hypothesis is that this degradation is related to the choice of inner optimizer, which shapes the pseudogradient used by the outer optimizer. Given the recent success of Muon relative to AdamW for LLM pre-training, its larger critical batch sizes, and its orthonormalized optimizers steps that are substantially different from those of AdamW~\citep{jordan2024muon, kimimuon, shah2025practical}, we conjecture that, beyond its improved convergence, Muon may offer additional advantages in terms of compressibility, parallelization, and performance within a Local SGD framework. Given the potential of Muon to be a powerful inner optimizer for DiLoCo, in what follows, we systematically compare MuLoCo (Muon inner optimizer) to DiLoCo in settings using a large number of workers (K), long communication intervals (H), when using compressed communication, and when using streaming updates. We further scale compute following a chinchilla optimal prescription and examine whether our most interesting findings hold at up to 15B parameter scale. Our contributions can be summarized as follows: 

\textbf{MuLoCo $K>1$}
\begin{itemize}
  \setlength\itemsep{0em} 
    \item We establish that MuLoCo's performance scales better with workers than DiLoCo. Specifically, beyond outperforming DiLoCo in absolute terms, MuLoCo outperforms DiLoCo at $K>2$ workers even when normalizing by the performance of their respective data parallel baselines. We find that this advantage is maintained at scale.
    \item We demonstrate that MuLoCo's performance at large worker counts (K=16) improves relative to data-parallel baselines as the training scale increases. Indeed, at 15B scale, K=16 MuLoCo matches the final performance of DP AdamW on language model evaluation benchmarks.
    \item We demonstrate that MuLoCo outperforms DiLoCo under long synchronization intervals, under pseudogradient compression, with partitioned communication, and in wall-clock training time, while having lower memory complexity.
    \item At 3.1B scale, we find that MuLoCo’s critical batch size is 2–4$\times$ larger than DiLoCo’s, further reducing communication while increasing usable data parallelism.
\end{itemize}
\textbf{MuLoCo $K=1$}
\begin{itemize}
  \setlength\itemsep{0em} 
    \item At 3.1B scale, we find that K=1 MuLoCo outperforms DP Muon, DP AdamW, and K=1 DiLoCo, while having stronger performance and larger critical batch sizes—placing it on the Pareto frontier of performance vs. training time among the optimizers in our study. 
    \item Our scaling laws predict that K=1 MuLoCo will retain its performance advantage and larger critical batch size over DP Muon at scale, leading to faster training at the largest model sizes. Our experiments at 15B validate this, showing that K=1 MuLoCo matches the best performing baselines while using a (4-16X) larger $16$M token batch size.
\end{itemize}

\textbf{Understanding MuLoCo's improved worker scaling}
\begin{itemize}
  \setlength\itemsep{0em} 
     \item Empirically, we find that Muon's orthonormalized steps lead to a more directionally aligned pseudogradient than AdamW and that the spectrum of AdamW's individual weight differences collapses during averaging.
    \item Theoretically, we link these findings, showing that the pseudogradient’s nuclear norm is governed by its alignment with individual optimizer steps and the magnitude (Frobenius norm) of these steps. Our analysis reveals Muon’s orthonormalized steps stabilize these magnitudes relative to AdamW.
    \item Our findings demonstrate that the structure of the inner-optimizer's step can have a non-trivial effect on pseudogradient quality.
\end{itemize}

\section{Background}\label{sec:background}
We briefly present the necessary background for understanding our empirical evaluation of MuLoCo. This includes implementation details of Muon, a review of DiLoCo, a review of partitioned communication, a review of common compression algorithms for neural networks, and a review of efficient communication collectives used in tandem with these compression algorithms.

\textbf{Muon} is a recently proposed~\citep{jordan2024muon} second-order optimizer that takes orthonormalized steps. Specifically, at each step, Muon updates its momentum accumulator,
\[
m_t = \beta m_{t-1} + g_t,
\]
and applies five steps of \textit{Newton--Schulz} iteration~\citep{bjorck1971orthogonal,kovarik1970orthonormality} to produce an orthogonalized update. Specifically, the iteration refines
\[
X_{t,j} = p(X_{t,j-1} X_{t,j-1}^\top) X_{t,j-1},
\]
where \(p(x) = ax+bx^3+cx^5\) with $(a,b,c)$ empirically tuned to be $3.4445,-4.7750,2.0315$~\citep{jordan2024muon}. This procedure effectively scales and rotates \(m_t\) toward the nearest orthogonal matrix, approximating \(UV^\top\) when \(m_t = U \Sigma V^\top\) is its singular value decomposition. The resulting orthogonalized update \(O_t\) is then rescaled by the learning rate and applied as $W_t = W_{t-1} - \eta O_t -\eta\lambda W_{t-1}$ with optional decoupled weight decay. In this work, we always use decoupled weight decay with Muon as it was shown to be important in maintaining Muon's convergence benefits at larger model sizes~\citep{kimimuon}.

\textbf{DiLoCo}~\citep{douillard2023diloco} is a distributed optimization algorithm that independently trains a single model with DP AdamW across $K$ workers in parallel while only communicating across workers every $H$ gradient descent steps. When individual workers span highly interconnected compute pools (e.g., single nodes, racks, adjacent racks, or entire datacenters for the largest models) with significantly faster bandwidth within pools than across them, DiLoCo can lead to significant speedups relative to DP training across the same number of accelerators. We present the original DiLoCo algorithm and highlight our modifications to it in Algorithm~\ref{algo:muloco} and~\ref{algo:muloco-ef}.

\textbf{Partitioned Communication in Streaming DiLoCo.}~\cite{streamingdiloco} propose Streaming DiLoCo, a variant of DiLoCo which, among other modifications, reduces peak communication bandwidth by partitioning a model's layers into $J$ subsets. With a communication interval $H$ ($J$ must divide $H$), each subset $j$ is communicated at an assigned multiple $j\cdot H/J$ with $1\leq j\leq J$ of the communication interval. In this work we evaluate the performance of MuLoCo under partitioned communication.

\textbf{Compressed Communication.} Compressing neural network gradients and parameters has been well studied in the context of reducing communication costs~\citep{fetchsgd,stitch2018sparse,pmlr-v202-wang23t,peng2024decoupled}. This work studies two common compression algorithms: quantization and sparsification. We now briefly discuss how these compressors are applied to a single weight matrix $\mW\in\R^{m\times n}$:
\vspace{-5pt}
\begin{itemize}
  \setlength\itemsep{0em}
    \item \textbf{Top-$k$} sparsification involves keeping the $k\%$ largest magnitude entries of $\mW$. The remaining entries are set to $0$ and are not communicated. Note that one must still communicate the sparsity pattern~\citep{pmlr-v202-wang23t}.
  \item \textbf{Quantization} involves mapping the entries of $\mW$ to a much smaller representable range, such that they can be efficiently encoded using an offset and a limited number of discrete levels from a codebook. The number of bits determines the size of the codebook, while the codebook itself defines the quantized values used to approximate entries in $\mW$. In \emph{linear} quantization, these levels are uniformly spaced over a fixed range, whereas \emph{statistical} (or non-uniform) quantization allocates levels based on the empirical distribution of $\mW$, typically assigning higher resolution to more frequently occurring values.
    
    
\end{itemize}

\textbf{Collectives for compressed communication.}
A ring all-reduce is bandwidth-optimal in terms of communication volume, achieving the minimum possible per-worker data transfer for dense reductions~\cite{zeropp}. However, exploiting compressed communication generally requires modifying the underlying collective. For example, Top-$k$ sparsification necessitates an all-gather operation, whose bandwidth cost grows linearly with the number of workers. 

For quantized communication, naively applying a ring-based reduce-scatter leads to error accumulation, since each hop requires a dequantize--reduce--quantize operation to preserve compression benefits. A natural remedy is to replace the ring reduce-scatter with an all-to-all reduce-scatter, which achieves the same aggregate bandwidth cost while performing reductions locally after receiving all peers’ quantized pseudogradients~\citep{zeropp}. 

Unlike prior work~\citep{streamingdiloco,basu2019qsparselocalsgd}, which abstracts away the choice of communication primitive, we explicitly model an efficient all-to-all reduce-scatter followed by a ring all-gather in our experiments. This design requires a single dequantize--reduce--quantize operation between the two collectives, resulting in two quantization and two dequantization operations per pseudogradient communication. To the best of our knowledge, this regime has not been empirically evaluated before. 

\textbf{Error feedback (EF)} is a powerful technique for mitigating the bias introduced by lossy compression~\citep{karimireddy2019error}. EF works by persisting
the difference between the original update and its compressed version and
incorporating it into future communications. Concretely, given a compressor
$\gC(\cdot)$ and a residual accumulator $\gE_i$, worker~$i$ updates its EF accumulator $\gE_i \leftarrow \beta\,\gE_i + \Delta_i$, communicates the compressed
update $\tilde{\Delta}_i = \gC(\gE_i)$, and removes the communicated value
$\gE_i \leftarrow \gE_i - \tilde{\Delta}_i$, thereby only persisting what was not communicated. Error feedback has been shown to recover convergence guarantees
comparable to uncompressed optimization for both sparsification and quantization
under standard assumptions~\citep{karimireddy2019error}. In our compression
experiments, we ablate the effect of error feedback.

\section{Related Work}
We now contextualize our paper within the rich body of existing works on distributed and communication-efficient optimization (\ref{sec:local-updates},\ref{sec:bib-compression}), single-worker DiLoCo (\ref{sec:singleworkerdiloco}), and large batch training (\ref{sec:large-batch}).

\subsection{Distributed optimization with local steps}\label{sec:local-updates} 
The idea of taking multiple local steps before communicating originated in the context of federated learning~\citep{mcmahan2017communication}, where the algorithm is known as Federated Averaging (FedAVG). In an i.i.d. setting, a variant of FedAVG known as Local SGD has been shown theoretically lead to communication savings~\citep{stich2018local}. Models trained with LocalSGD were also found to have improved generalization relative to large batch training~\citep{lin2018local}. Follow-up works have proposed more sophisticated server-side optimizers~\citep{wang2019slowmo, joseph2023lo, reddi2021adaptive}, while other research has focused on adaptive client-side optimizers~\citep{zhou2024fedcada,douillard2023diloco,sani2024future}. Notably, in the context of LLM pre-training,~\citet{douillard2023diloco} propose DiLoCO, which uses SGD with Nesterov momentum as the server-side optimizer and AdamW on local workers. DiLoCo was demonstrated to scale to very large numbers of local steps ($H>500$)~\citep{douillard2023diloco} and large models~\citep{scalingdiloco,opendiloco,intellect1}. Other recent works focus on studying DiLoCo variants that also communicate optimizer states, allowing them to prove convergence and improve performance at the cost of additional communication~\citep{cheng2025convergence,iacob2026desloc,iacob2026mtdao,lordo}. In a direct follow-up to the original DiLoCo work,~\citet{scalingdiloco} study the performance of DiLoCo as model size and training data are scaled up. Crucially, the authors find that at each scale, DiLoCo's performance degrades as the number of workers is increased, but that as the model scale is increased, the performance degradation decreases. We complement these findings in our work, by showing that MuLoCo improves on worker scaling over DiLoCo relative to their respective data-parallel baselines and that at large worker counts, MuLoCo also improves with scale.

\subsection{Distributed optimizers with compressed communication}\label{sec:bib-compression} 
Beyond leveraging infrequent communication, many distributed optimizers directly reduce the number of bits communicated through compression. This is typically done by sparsifying~\citep{stitch2018sparse, shi2019topk, peng2024decoupled}, quantizing~\citep{alistarh2017qsgd}, sketching~\citep{fetchsgd}, computing low-rank approximations of the gradient~\citep{zhao2024galore,vogels2019powersgd,ahn2025dion}, or combining any number of these approaches~\citep{pmlr-v202-wang23t}. Among these works, many leverage error feedback for mitigating the bias introduced by lossy compression~\citep{karimireddy2019error}.

Some works combine local SGD or DiLoCo with various compression methods. In the context of image classification,~\citet{basu2019qsparselocalsgd} apply top-k sparsification and quantization to error-corrected moving averages of parameter deltas. Streaming DiLoCo~\citep{streamingdiloco} combines partitioned communication, overlapped communication, and quantization to reduce communication cost in DiLoCo. In concurrent work, SparseLoCo~\citep{sparseloco} demonstrates that using a local error feedback as an outer momentum allows reaching extreme compression ratios in DiLoCo without compromising performance.


\subsection{Single worker DiLoCo}\label{sec:singleworkerdiloco} 
\citet{scalingdiloco} established that the performance of DiLoCo with only a single worker is consistently stronger than AdamW alone. This is partially a rediscovery of the lookahead optimizer~\citep{lookahead} and reptile~\citep{reptile,cptreptile}, except that single-worker DiLoCo uses a Nesterov-style update as opposed to vanilla SGD. In follow-up work,~\citet{snoo} establishes that compared to lookahead, single-worker DiLoCo (or SNOO) performs substantially better. They also show that the performance of single-worker DiLoCo when training MoE and dense LLMs can lead to a substantial speedup over AdamW alone. In concurrent follow-up work,~\cite{gpa} propose GPA, an algorithm that generalizes and improves single-worker DiLoCo by eliminating the outer optimizer and enabling smooth iterate averaging at every step. GPA is shown to outperform single-worker DiLoCo while requiring one less accumulator. In this work, we establish the performance of single-worker MuLoCo at H=30 inner steps, showing that it outperforms data parallel Muon and has larger critical batch sizes. To the best of our knowledge, we are the first to establish K=1 MuLoCo's larger critical batch sizes and show that it obtains a better performance-training-time tradeoff than DP Muon. We expect that the performance improvements we establish will directly apply to GPA with a Muon inner optimizer.

\subsection{Large-Batch Training and Critical Batch Size}\label{sec:large-batch} 
Scaling batch size may be the simplest way to reduce training time, whether it be from increased data parallelism or improved hardware utilization. However, increasing batch size beyond a certain threshold, know as the critical batch size (CBS), leads to significant diminishing returns in performance. As such, many works have studied the large-batch training regime and its performance differences to using smaller batch sizes. Early empirical work highlighted that large batches can alter optimization dynamics and lead to poorer generalization behaviour~\citep{keskarlgb,jastrezbski2018three}. At the same time, practical recipes such as linear learning-rate scaling and warmup demonstrated that large-batch training can match small-batch accuracy when carefully tuned~\citep{goyal2017accurate}. Other work found that the generalization gaps found in earlier work stemmed from insufficient hyperparameter tuning, rather than inherent limitations of batch size~\citep{shallue2019measuring}. In the context of language modeling, \citet{cbsmccandlish} introduced the \emph{gradient noise scale} as a predictor of the CBS, hypothesizing that the CBS coincides with the point at which increasing batch size no longer meaningfully reduces noise in the gradient. Recent work has revisited and refined this perspective. \citet{cbszhang} demonstrate that critical batch size scales primarily with dataset size rather than model size. \citet{cbsmerril} propose a direct empirical measurement of the critical batch size, while \citet{bergsma2025power} derive scaling laws relating optimal and critical batch sizes to dataset size, providing an alternative methodology for predicting the CBS. Finally, several works establish that the critical batch size depends on the optimization algorithm~\citep{zhang2019which,lin2020extrapolation}. For instance, in the LLM regime, \citet{vyas2025soap} show empirically that second-order preconditioned methods such as SOAP tolerate larger batches than AdamW, while \citet{shahmuoncbs} demonstrate that Muon can also reach larger critical batch sizes. Our study further expands on this, showing that MuLoCo achieves a Pareto optimal performance-batch size trade-off among all optimizers in our study (notably including DP Muon).

\section{Method}
Our main methodological contribution is to replace DiLoCo's AdamW inner optimizer with Muon. While this is a simple change, our results (Sec.~\ref{sec:results-ablations}) will show that beyond improving over DiLoCo in absolute terms, MuLoCo improves performance at larger worker counts over DiLoCo \textit{even when normalizing by data-parallel Muon and AdamW, respectively}. That is, MuLoCo's improvements over DiLoCo \textbf{can not} entirely be explained by Muon's performance improvement over AdamW. Our main assumption is that this improvement can be explained by Muon’s pseudogradients being more directionally correct than AdamW’s. In this section, we introduce MuLoCo and theoretically analyse how the alignment of AdamW and Muon optimizer steps affects the pseudogradient’s nuclear norm, offering a speculative explanation for the observed performance difference between MuLoCo and DiLoCo.


\subsection{MuLoCo}
MuLoCo reduces communication by training multiple models in parallel and infrequently communicating weight-space parameter differences after $H$ steps of local optimization. Specifically, each worker $k \in [K]$ samples data from its assigned shard $\mD_k$ and performs local optimization as follows:
\begin{equation}\label{eq:inner-steps}
    \vtheta_{k}^{(t+h)} = \texttt{Muon}\!\left(\vtheta_{k}^{(t+h-1)}, \nabla_\vtheta \mathcal{L}\right), \quad h=1,\ldots,H.
\end{equation}
The worker-level parameter differences, $\Delta_k^{(t)} = \vtheta^{(t-H)} - \vtheta_k^{(t)}$, are then communicated to form a pseudogradient:
\begin{align}\label{eq:pseudogradient}
\Psi^{(t)} = \frac{1}{K}\sum_{k=1}^K \Delta_k^{(t)},
\end{align}
which captures the average weight-space displacement across all workers’ inner optimization steps. The pseudogradient $\Psi^{(t)}$ is then used as input to SGD with Nesterov momentum, which updates its momentum and model weights as:

\begin{align}\label{eq:nesterov}
u^{(t)} &= \mu\, u^{(t-H)} + \eta_{\text{out}}\, \Psi^{(t)} , \nonumber\\
\vtheta^{(t)} &= \vtheta^{(t-1)} - \mu\, u^{(t)} - \eta_{\text{out}}\, \Psi^{(t)}.
\end{align}

\begin{figure*}[t]
    \centering
    \subfloat[MuLoCo]{\includegraphics[width=0.4\linewidth]{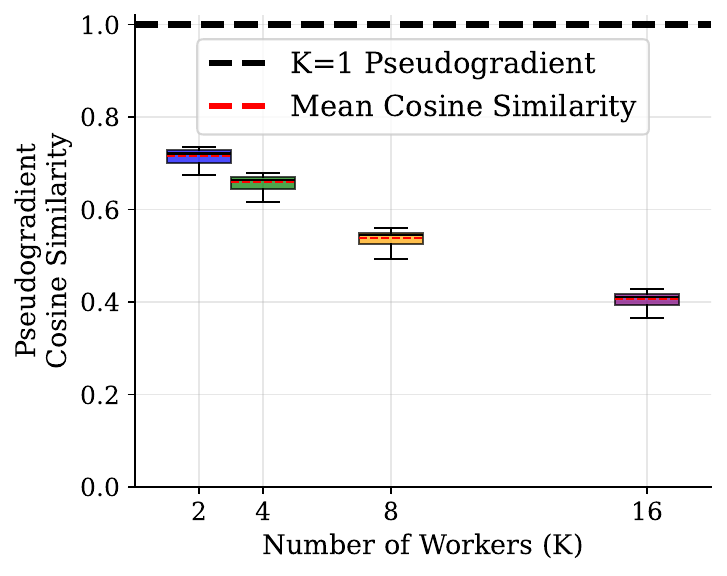}}\qquad\qquad
    \subfloat[DiLoCo]{\includegraphics[width=0.4\linewidth]{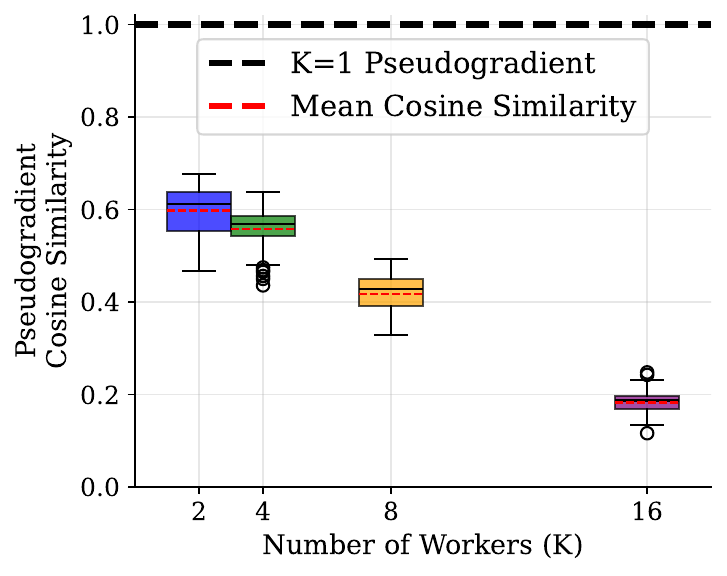}}
    \caption{\textbf{Pseudogradient cosine similarities compared to a FLOP-matched $K=1$ pseudogradient.} Subfigure (a) shows results for Muon, while subfigure (b) shows results for AdamW. The box plots report the spread of cosine similarities across all hidden weight matrices of a 416M transformer. We observe that Muon's pseudogradients are more aligned with the K=1 pseudogradient, and this alignment decreases more slowly as the number of workers is increased.    }
    \label{fig:cosinesim-box} \vspace{-15pt}
\end{figure*}

It should be noted that DiLoCo can be recovered by swapping Muon for AdamW in equation~\ref{eq:inner-steps}. Therefore, the pseudogradients of each method depend on the structure of individual optimizer updates. With this in mind, we will now theoretically connect the alignment of individual optimizer steps to the pseudogradient’s nuclear norm, showing how they differ for Muon and AdamW.

 


\subsection{Analyzing the benefits of Muon pseudogradients over AdamW}
Our results (Sec.~\ref{sec:results-ablations}) will show that beyond improving over DiLoCo in absolute terms, MuLoCo improves performance at larger worker counts over DiLoCo even when normalized by its own data-parallel baseline. 
To better understand the reasons for this improvement we ran a number of analytical experiments targeted towards analyzing the impact of Muon and AdamW on pseudogradient quality, which Section~\ref{sec:inner-opt} describes in depth. In the remainder of this section we will summarize the important results as they relate to our theoretical analysis.  

Figure~\ref{fig:cosinesim-box} reports the average cosine similarity across all hidden layers in a 12-layer transformer (416M in Table~\ref{tab:model_details}) for DiLoCo and MuLoCo at $K\in\{2,4,8,16\}$ workers. The cosine similarity is computed between vectorized tensors and their corresponding K=1 pseudogradients. We find that at each worker count, MuLoCo's pseudogradient is more aligned with the K=1 pseudogradient (equivalent to the data-parallel weight difference). Moreover, the variance in cosine similarity across hidden tensors in the network is much larger for AdamW than Muon. These findings suggest that improved alignment may be a key factor in MuLoCo's improved performance at larger worker counts

\begin{figure*}[t]
    \centering
    \subfloat[Pseudogradient Spectrum]{\includegraphics[width=0.45\linewidth]{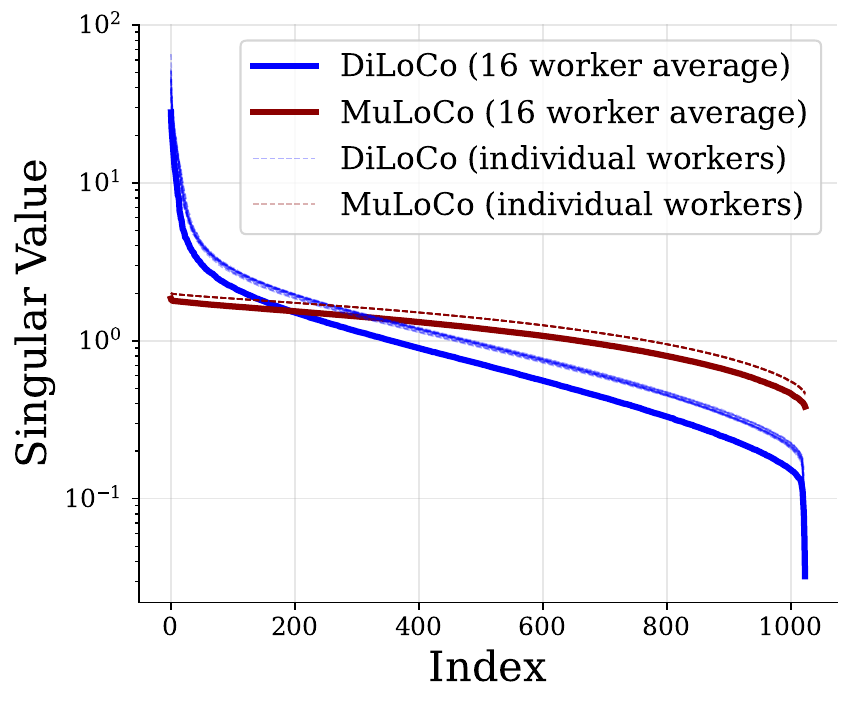}}
    \qquad
    \subfloat[Top-5\% Interference gap]{\includegraphics[width=0.45\linewidth]{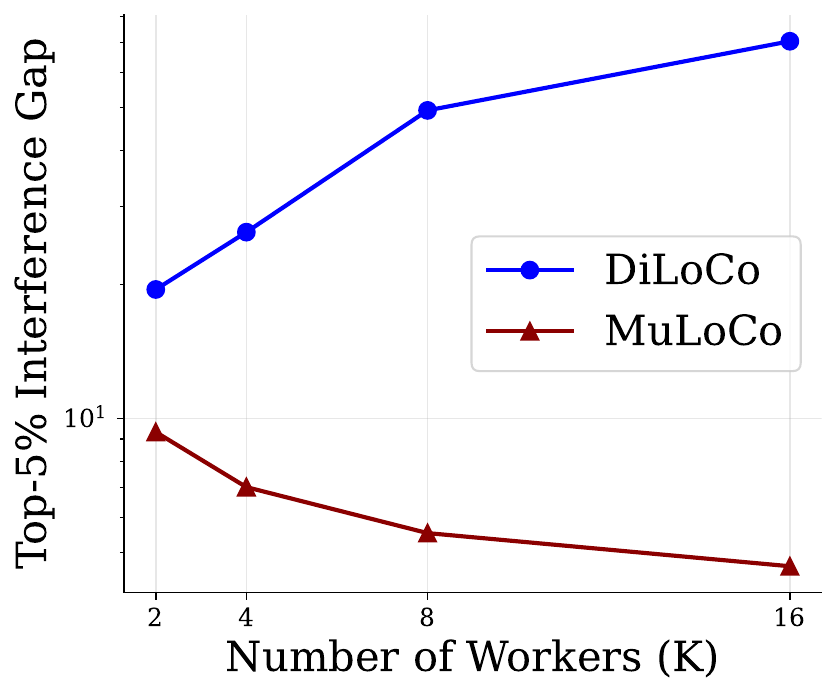}}
\caption{\textbf{DiLoCo's pseudogradient spectrum decays relative to the spectrum of its trajectories as the number of workers increases.}
(\textbf{a}) We plot the singular values of each worker weight difference $\Delta_k$ (dashed) and of the averaged pseudogradient $\Psi$ (solid), where
$\Psi=\frac{1}{K}\sum_{k=1}^K \Delta_k$.
(\textbf{b}) We quantify destructive interference via the \emph{top-$S$ interference gap}
$G_k=\frac{1}{K}\sum_{k=1}^K \sum_{j=1}^S \sigma_j(\Delta_k)\;-\;\sum_{j=1}^S \sigma_j(\Psi)$,
i.e., the difference between the mean top-$S$ Ky--Fan spectral mass of individual worker updates and the top-$S$ spectral mass of their average. We observe that DiLoCo's interference gap grows as the number of workers increases, while Muon's shrinks, providing a possible explanation for DiLoCo's stronger performance degradation at large worker counts.
}
    \label{fig:interferencegap}
\end{figure*}

In an attempt to better understand the exacerbated pseudogradient degradation at higher worker counts for DiLoCo relative to MuLoCo, in figure~\ref{fig:interferencegap} (a) we visualize the pseudogradient spectra of DiLoCo and MuLoCo before and after averaging. We observe that the spectrum of the individual worker's pseudogradients collapse for AdamW during averaging, while the collapse is much less pronounced for muon. To measure this we introduce the interference gap.
\begin{definition}[top-$S$ interference gap]
Let $A_1,\dots,A_N \in \mathbb{R}^{m\times n}$ and define their mean
\[
\bar{A} = \frac{1}{N}\sum_{i=1}^N A_i.
\]
Denote by $\sigma_1(X) \ge \dots \ge \sigma_r(X) \ge 0$ the singular values of a matrix
$X \in \mathbb{R}^{m\times n}$ with $r = \min\{m,n\}$.
For any $k \in \{1,\dots,r\}$, the \emph{top-$S$ interference gap} is
\[
G_S =  \frac{1}{N}\sum_{i=1}^N\sum_{j=1}^S  \sigma_j(A_i)
      -\sum_{j=1}^S\sigma_j(\bar{A}).
\]
Here, $G_S \ge 0$ measures the loss of top-$S$ spectral mass when averaging the matrices:
$G_S = 0$ indicates perfect alignment of dominant singular directions,
while larger $G_S$ implies stronger destructive interference.
\end{definition}
Figure~\ref{fig:interferencegap} (b) reports the top 5\% interference gap observed for MuLoCo and DiLoCo at different worker counts, finding that $G_S$ decreases with worker count for MuLoCo but not DiLoCo. These findings reveal differences in behaviour between AdamW and Muon pseudogradients, suggesting that top-$S$ interference may be related to the observed performance improvements for MuLoCo.

\subsection{Relating Muon and AdamW Pseudogradient Spectra to Per-step Alignment}\label{sec:theory}
Inspired by our empirical results of the previous section, which show a meaningful difference between DiLoCo and MuLoCo pseudogradients in terms of alignment with the K=1 pseudogradient and top-$S$ interference during averaging, in this section, we theoretically link the two. Specifically, we show that the nuclear norm of MuLoCo and DiLoCo pseudogradients is directly related to their alignment with individual optimizer steps.

\textbf{AdamW inner updates and induced pseudogradient.}
For a given worker $k$ and local step $h\in\{1,\ldots,H\}$ with gradient $g_{k}^{(t+h)}$, AdamW maintains first and second moments
\begin{align}
m_{k,h}=\beta_1 m_{k,h-1}+(1-\beta_1)g_{k}^{(t+h)},\qquad\nonumber\\
v_{k,h}=\beta_2 v_{k,h-1}+(1-\beta_2)\,g_{k}^{(t+h)}\odot g_{k}^{(t+h)},\nonumber
\end{align}
and applies decoupled weight-decay leading to the following worker trajectory over $H$ local steps
\begin{align}
\Delta^{(t,k)}_\text{AdamW}
=\sum_{h=1}^{H}\!\Bigg(
\alpha_{t+h}\,\frac{m_{k,h}}{\sqrt{v_{k,h}}+\varepsilon}
+\lambda\,\alpha_{t+h}\,\vtheta_{k}^{(t+h-1)}
\Bigg).\nonumber
\end{align}

\medskip
\textbf{Muon inner updates and induced pseudogradient.}
With decoupled weight decay, Muon's orthonormalized update yield the following worker trajectory:

\begin{align*}
\Delta^{(t,k)}_\text{Muon}=\sum_{h=1}^{H}\!\Big(
\alpha_{t+h}\,O_{k,h}
+\lambda\,\alpha_{t+h}\,\vtheta_{k}^{(t+h-1)}
\Big).\nonumber
\end{align*}
For a more detailed explanation of Muon's update, we refer the reader to section~\ref{sec:background}.

\begin{figure}[t]
  \centering
  \begin{subfigure}[t]{0.49\textwidth}
    \centering
    \includegraphics[width=\linewidth]{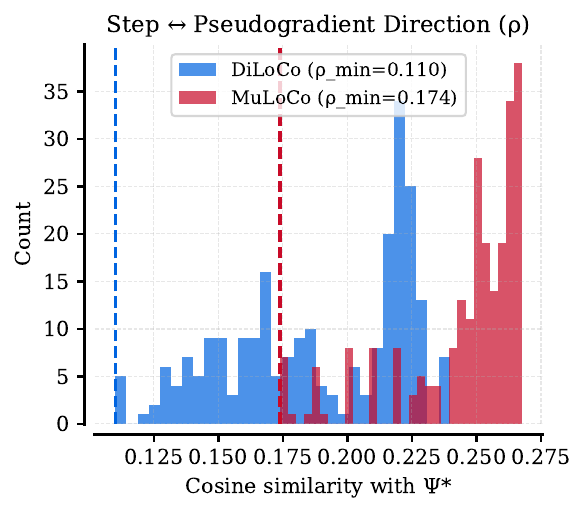}
    \label{fig:step_align}
  \end{subfigure}\hfill
  \begin{subfigure}[t]{0.49\textwidth}
    \centering
    \includegraphics[width=\linewidth]{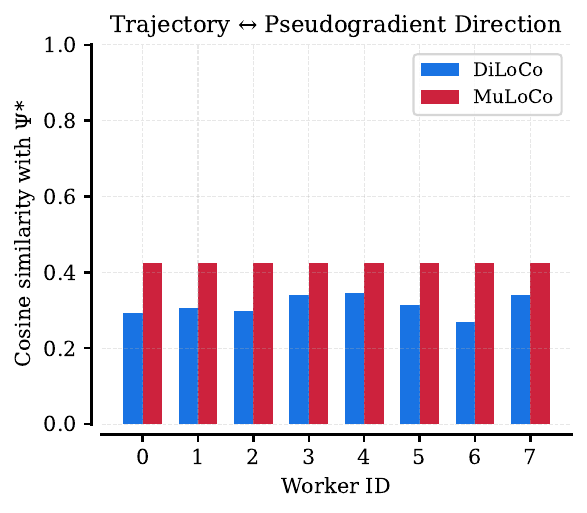}
    \label{fig:traj_align}
  \end{subfigure}
\vspace{-20pt}
  \caption{\textbf{Alignment to the full pseudogradient.}
  We report (left) the cosine similarity between \emph{individual inner-optimizer steps} and the full communicated pseudogradient, and (right) the cosine similarity between \emph{per-worker weight differences} (i.e., worker trajectories) and the full pseudogradient.
  We observe that MuLoCo’s (Muon inner) optimizer steps are more consistently aligned with the final pseudogradient direction.
  Moreover, worker weight differences under Muon exhibit nearly identical cosine similarity to the final pseudogradient across workers, while AdamW shows considerably higher inter-worker variability.}
  \label{fig:step_and_trajectory_alignment}
\end{figure}
Since the weight decay hyperparameter for Muon and AdamW is generally chosen to be $\lambda \leq 10^{-1}$ (this is the case in all our experiments), the contribution of decoupled weight decay to the pseudogradient is negligible. As such, we exclude it from our analysis. We now demonstrate that the nuclear norm of general pseudogradients depends on the alignment of individual optimizer steps with the pseudogradient and on the Frobenius norm of the optimizer step, itself.
\begin{proposition}[Pseudogradient nuclear norm depends on optimizer step alignment]
\label{thm:pseudogradient-spectrum-lowerbound}
Fix $m,n\in\mathbb{N}$ and let $r:=\min\{m,n\}$. Consider a pseudogradient of the form
\[
\Psi \;=\;\frac{1}{K}\sum_{k=1}^K\sum_{h=1}^H \alpha_{t+h}\,\psi^{(t+h,k)}\in\mathbb{R}^{m\times n},
\qquad \alpha_{t+h}\ge 0.
\]
Let $\Psi = U\Sigma V^\top$ be its singular value decomposition and define the orthonormal factor $\Psi^\star := U V^\top.$ Let the cosine similarity between a given optimizer step and $\Psi^\star$ be
\[
\rho^{(t+h,k)} \;:=\;
\frac{\langle \psi^{(t+h,k)},\Psi^\star\rangle_F}{\|\psi^{(t+h,k)}\|_F\;\|\Psi^\star\|_F}.
\]
Then the nuclear norm of $\Psi$ satisfies
\[
\|\Psi\|_\ast
=
\frac{\sqrt{r}}{K}\sum_{k=1}^K\sum_{h=1}^H \rho^{(t+h,k)}\alpha_{t+h}\,\|\psi^{(t+h,k)}\|_F.
\]
\end{proposition}
\proof{The proof is provided in section~\ref{apdx:proof} of the appendix. $\qed$.}

\begin{corollary}[Muon pseudogradient]
\label{cor:muon-spectrum-lowerbound}
In the setting of Proposition~\ref{thm:pseudogradient-spectrum-lowerbound}, assume that the inner-step optimizer updates, $\psi^{(t+h,k)}$, are produced by Newton Shulz Iteration, making them orthonormal;
then
\begin{equation}\label{eq:muon-bound}
\|\Psi\|_\ast
=
\frac{r}{K}\sum_{k=1}^K\sum_{h=1}^H\rho^{(t+h,k)}\alpha_{t+h}.
\end{equation}
\end{corollary}
\proof{This follows from proposition~\ref{thm:pseudogradient-spectrum-lowerbound} and the fact that $\psi^{(t+h,k)}$ is orthonormal, since $\|\psi^{(t+h,k)}\|_F=\sqrt{r}. \qed$}

\begin{figure}[t]
  \centering
  \begin{subfigure}[t]{0.45\textwidth}
    \centering
    \includegraphics[width=\textwidth]{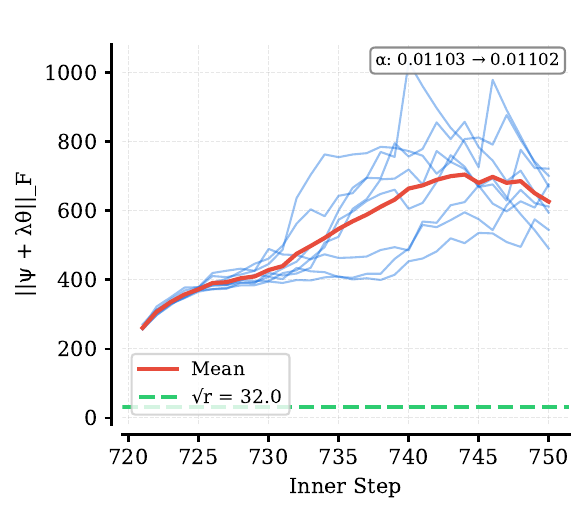}
    \caption{Frobenius norm of \textbf{DiLoCo} inner steps.}
    \label{fig:diloco_step_norms_layer11_w3}
  \end{subfigure}\qquad
  \begin{subfigure}[t]{0.45\textwidth}
    \centering
    \includegraphics[width=\textwidth]{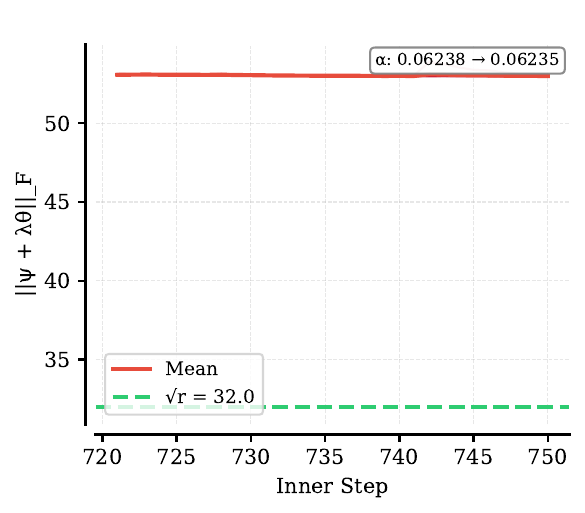}
    \caption{Frobenius norm of \textbf{MuLoCo} inner steps.}
    \label{fig:muloco_step_norms_layer11_w3}
  \end{subfigure}
\vspace{-5pt}
  \caption{\textbf{MuLoCo's orthonormalized inner steps lead to stable Frobenius norms.}
  Light blue (DiLoCo) and light red (MuLoCo) lines correspond to the Frobenius norm of the inner-optimizer step on each individual worker, while the red line reports their mean (MuLoCo's mean overlaps the individual workers). We observe that AdamW’s per-worker step norms are comparatively erratic across workers, whereas Muon’s are stable. Muon’s step Frobenius norms are larger than the predicted $\sqrt{r}$ magnitude (for rank $r$) due to the additional contribution of decoupled weight decay.}
  \vspace{-15pt}
  \label{fig:frobeniusnorm}
\end{figure}

\begin{corollary}[AdamW pseudogradient]
\label{cor:adamw-spectrum-lowerbound}
In the setting of Proposition~\ref{thm:pseudogradient-spectrum-lowerbound}, for general inner-step matrices $\psi^{(t+h,k)}$ (e.g., AdamW steps), one always has
\begin{equation}\label{eq:adamw-bound}
\|\Psi\|_\ast
=
\frac{\sqrt{r}}{K}\sum_{k=1}^K\sum_{h=1}^H \rho^{(t+h,k)}\alpha_{t+h}\,\|\psi^{(t+h,k)}\|_F.
\end{equation}
\end{corollary}
\proof{The proof is identical to proposition~\ref{thm:pseudogradient-spectrum-lowerbound}. $\qed$}

We observe that Muon's orthogonalized optimizer step results in the pseudogradient's nuclear norm naturally varying with the per-step alignment ($\rho$), the number of workers ($K$), and the inner learning rate ($\alpha$). In contrast, the spectrum of AdamW's pseudogradient also depends on $\|\psi^{(t+h,k)}\|_F$, which may be highly variable at higher worker counts, leading to overemphasizing certain steps unnecessarily. This exact behaviour is illustrated in Figure~\ref{fig:frobeniusnorm} where we empirically measure the Frobenius norm of individual optimizer steps, forming the pseudogradient for DiLoCo and MuLoCo. Subfigure (a) shows DiLoCo's pseudogradient Frobenius norm is highly variable across different workers at each inner step, while subfigure (b) shows that MuLoCo's is constant throughout. Intuitively, Muon's orthonormalized optimizer steps allow each individual optimizer step to contribute equally to the pseudogradient's spectrum. While it should be noted that our theory has no direct connection to performance and that all inferences made here are purely speculative, the slower decrease in cosine similarity of MuLoCo in Figure~\ref{fig:cosinesim-box} relative to DiLoCo is correlated with its relatively stronger performance at larger worker counts (Figure~\ref{fig:worker-scaling} (a)). Together with our theory connecting the pseudogradients nuclear norm to the cosine similarity and Frobenius norm of individual optimizer steps, we speculate that the stronger alignment of MuLoCo’s optimizer steps and its stable Frobenius norm across them may lead to a larger pseudogradient spectrum with less interference during averaging (e.g., as observed in Figure~\ref{fig:interferencegap}), which may improve performance.


\section{Empirical Evaluation}\label{sec:emp-eval}
Our goal is to establish the performance of MuLoCo relative to DiLoCo and their data-parallel baselines in the context of LLM pre-training. In Section~\ref{sec:results-ablations} we start by studying various settings relevant to communication-efficient training on a small 416M language modelling task. Subsequently, in Section~\ref{sec:results-scaling} we study scaling laws (150M-3B) of models pre-trained with these optimizers and scale up to $15B$. We now detail the exact experimental setting for these experiments.


\textbf{Base LM pre-training Task.} For all experiments except scaling laws, we select a $416$M dense transformer language modeling task (see $416$M in Table~\ref{tab:model_details} for exact model dimensions). The architecture follows Gemma3~\citep{gemma} with SwiGLU FFNs, QK norm, and additional RMS normalization layers before residual connections. We tokenize using the Llama3 tokenizer~\cite{grattafiori2024llama3herdmodels}. Unless otherwise specified, we synchronize every $H=30$ local steps and use $K=8$ workers. In all of our experiments, we use a sequence length of $2048$, we decay the learning rate following a cosine decay schedule to $0.1\times$ the maximum value, and we pre-train on the high-quality portion of the Nemotron-CC dataset~\citep{nemotroncc}. For experiments scaling the communication steps H, with compressed updates, or using streaming DiLoCo, we fix the global batch size to $B=1$M tokens. In our scaling experiments, the batch size is swept as described below.

\textbf{Scaling study.} In our scaling experiments, we increase model and dataset size following a ratio of 20 tokens per parameter (TPP)~\citep{chinchilla}. We report exact model sizes and token budgets in Table~\ref{tab:model_details}. At each scale except 15B, we conduct extensive hyperparameter sweeps as described below.

\begin{table}[h]
\centering
\caption{Gemma3-style transformers used in our scaling study. The architecture follows Gemma3~\citep{gemma} with SwiGLU FFNs, QK norm, and additional RMS normalization layers before residual connections. However, we tokenize using the Llama3 tokenizer~\citep{grattafiori2024llama3herdmodels}. All models use a sequence length of $2048$.}
\label{tab:model_details}
\begin{adjustbox}{width=0.7\columnwidth}
\begin{tabular}{ccccccc}
\toprule
\textbf{\makecell{Model \\ Parameters}} & \textbf{\makecell{Transformer \\ Layers}} & \textbf{\makecell{Attention \\ Heads}} & \textbf{\makecell{QKV \\ Dimension}} & \textbf{\makecell{Hidden \\ Dimension}} & \textbf{\makecell{Token \\ Budget}} & \textbf{\makecell{HP \\ Sweep}} \\
\midrule
150M & 6 & 4 & 512 & 1,408 & 3B & \textcolor{green}{$\checkmark$} \\
416M & 12 & 8 & 1,024 & 2,816 & 8.16B & \textcolor{green}{$\checkmark$} \\
914M & 18 & 12 & 1,536 & 4224 & 18.12B & \textcolor{green}{$\checkmark$} \\
1.76B & 24 & 16 & 2,048 & 5,632 & 35.23B & \textcolor{green}{$\checkmark$} \\
3.07B & 30 & 20 & 2,560 & 7,040 & 61.4B & \textcolor{green}{$\checkmark$} \\
15.23B & 54 & 36 & 4,608 & 12,672 & 304.6B & \textcolor{red}{$\times$} \\
\bottomrule
\end{tabular}
\end{adjustbox}
\end{table}

\textbf{Hyperparameter tuning.}  At all model scales, we tune the weight decay ($\lambda$), inner learning rate ($\eta_\text{in}$), global batch size (B), outer learning rate ($\eta_\text{out}$), and outer momentum ($\mu$) according to the following procedure. 
\begin{enumerate}
    \item (Data-Parallel $\lambda$) We first perform a grid search across $\lambda$ within$\{1^{-1},1^{-2},1^{-3},1^{-4}\}$ and $\eta_\text{in}$ values that are integer powers of $\sqrt{2}$ for the AdamW and Muon data-parallel baselines. We fix the $B=1$M at all scales for this initial sweep.
    \item (Data-Parallel $\eta_\text{in},B$) Having established the optimal $\lambda^*$ value at $B=$1M, we now perform a grid-search over integer powers of 2 for $B$ and integer powers of $\sqrt{2}$ for $\eta_\text{in}$, while rescaling $\lambda^*$ according to~\citet{scalewd} as B is varied. 
    \item (DiLoCo/MuLoCo $\lambda,\eta_\text{in},B$) Re-using the existing $\lambda^*$ found above for $B=$1M, at each worker count $K\in\{1,2,4,8,16\}$ we perform a grid-search over powers of 2 $B$ and powers of $\sqrt{2}$ $\eta_\text{in}$, while rescaling $\lambda^*$ according to~\citet{scalewd} \textit{using the per-worker batch size ($B/K$)}.
    \item (DiLoCo/MuLoCo $\eta_\text{out},\mu$)  When sweeping the outer optimizer's hyperparameters, we follow empirical guidance from~\citet{scalingdiloco} who found that for DiLoCo they remain roughly constant across model sizes. As such, we perform an extensive grid search at the $416$M scale sweeping across $\lambda$, $\eta_{\text{in}}$, $\eta_{\text{out}}$, $\mu$, and $B$. During this extensive grid search, we tune $\eta_{\text{out}}$ and $\mu$ by searching within $\{0.1,0.2,0.3,0.4,0.5,0.6,0.7,0.8,0.9,1.0\}$ with other hyperparameters spanning ranges from above. We report the optimal $\mu$ and $\eta_{\text{out}}$ values found in Figure~\ref{fig:outer_opt_grid} and re-use them at larger model scales.
\end{enumerate}

\textbf{DiLoCo} We use AdamW as a baseline inner optimizer for DiLoCo. Following~\citet{scalingdiloco} we set $\beta_1=0.9$ and $\beta_2=0.99$ for all experiments. Unless otherwise specified, we use a synchronization interval of $H=30$.

\textbf{MuLoCo} We use the Muon with a quintic Newton-Shulz iteration~\citep{jordan2024muon}. Muon is applied to hidden layers, while AdamW is used for the embeddings, normalization, and output layers. For hidden matrices of size $\mW \in \R^{m\times n}$ we rescale the learning rate by $\sqrt{\frac{n}{m}}$. We set $\beta_1=0.9$ for both Muon and AdamW. We set AdamW's $\beta_2=0.99$. Unless otherwise specified, we use a synchronization interval of $H=30$.

\textbf{Outer Optimizer} We use SGD with Nesterov momentum as the outer optimizer for all our DiLoCo and MuLoCo experiments with hyperparameters reported in Figure~\ref{fig:outer_opt_grid}.

\textbf{Data Parallel Baselines} We also report the performance of well-tuned DP AdamW and DP Muon baselines.

\textbf{Implementation} For Muon, we use~\citet{ahn2025dion}'s distributed implementation. For AdamW, we used the fused PyTorch implementation. All our pre-training experiments are run using TorchTitan~\citep{liang2025torchtitan}. We use the streaming DiLoCo implementation within TorchFT for all our distributed optimizers. We train across H100 and H200 GPUs. Our largest experiments at 15B scale use HSDP with 16 FSDP ranks and 64 data-parallel ranks. For DiLoCo and MuLoCo, they are accordingly subdivided among workers.

\textbf{Evaluation Loss} During training, we log the validation loss every $15$ steps on 2M token batches from a held-out set of Netotron-CC pre-training data. Following~\citep{chinchilla}, to reduce stochasticity in the final loss estimate, we use a smoothed final loss. Specifically, we first subsample the validation loss trajectory to synchronization boundaries (every $H=30$ steps) and apply a time-weighted EMA with base smoothing parameter $\alpha=0.2$. At the nominal spacing of $H=30$ steps, the adaptive coefficient is $\tilde{\alpha} \approx 0.181$, corresponding to an effective smoothing window of approximately $5$--$6$ synchronization rounds. This procedure yields a stable final loss estimate that is less sensitive to variance across validation batches. The resulting $\hat{L}$ is used consistently across all analyses in this paper, including hyperparameter selection, scaling law fitting, and cross-method comparisons. Further details are reported in Section~\ref{sec:evalloss} of the appendix.

\section{Results---Communication Efficient Ablations \& Wall-Clock Time}\label{sec:results-ablations}
The following section compares the performance of MuLoCo and DiLoCo at 416M scale across a number of settings relevant to communication-efficient training: (\ref{sec:inner-opt}) training across a large number of workers ($K=1-16$), (\ref{sec:moreh}) training with long synchronization intervals ($H=15-240$), (\ref{sec:comp}) compressing the pseudogradient (via quantization and sparsification), (\ref{sec:streaming}) reducing peak bandwidth with partitioned communication from streaming DiLoCo, and (\ref{sec:wallclock}) when accounting for wall-clock training time.

\begin{figure*}[h]
    \centering
    \subfloat[Scaling workers ($K$).]{
        \includegraphics[width=0.48\linewidth]{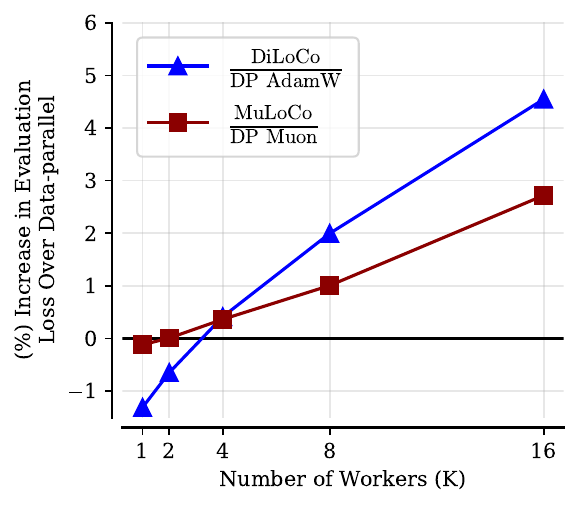}
        \label{fig:worker-scaling-vs-h:a}
    }
    \subfloat[Increasing synchronization interval ($H$).]{
        \includegraphics[width=0.48\linewidth]{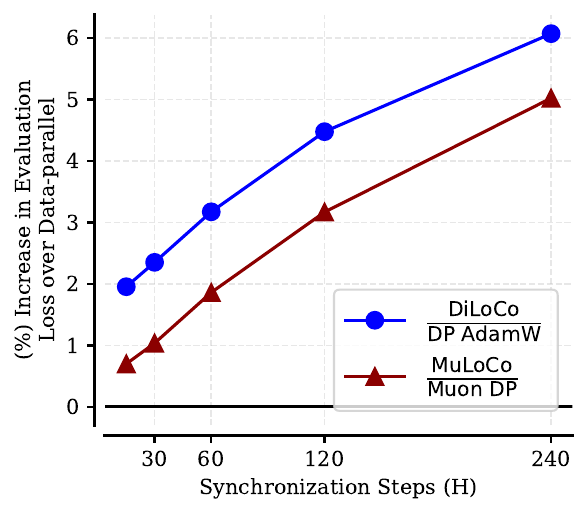}
        \label{fig:worker-scaling-vs-h:b}
    }
    \caption{\textbf{MuLoCo consistently outperforms DiLoCo, including relative to their respective data-parallel (DP) baselines at larger worker counts.}
    \textbf{Subfigure (a)} (same as subfigure (a) in Fig.~\ref{fig:worker-scaling}) reports the percentage increase in evaluation loss for DiLoCo and MuLoCo relative to their DP baselines. We observe that the relative performance of each method degrades as $K$ is increased, but MuLoCo shows clear improvements over DiLoCo \emph{relative to} their data-parallel baselines as the number of workers grows.
    \textbf{Subfigure (b)} varies the synchronization interval $H$ and shows that MuLoCo continues to outperform DiLoCo relative to their DP baselines, maintaining MuLoCo’s improvements at $K{=}8$ workers across different synchronization intervals.}
    \label{fig:worker-scaling-vs-h}
\end{figure*}
\begin{generaltakeaway}{Takeaways:}
  \begin{itemize}
    \item \textbf{Better worker scaling.}  beyond outperforming DiLoCo in absolute terms, MuLoCo outperforms DiLoCo at $K>2$ workers \textit{even} when normalizing by their respective DP baselines.
    \item \textbf{Compatible with reduced communication.} MuLoCo outperforms DiLoCo under long synchronization intervals, pseudogradient compression, and when using with partitioned communication.
    \item \textbf{Better memory complexity.} MuLoCo has a lower memory complexity than DiLoCo (3X v.s. 4X).
    \item \textbf{Low optimizer step overhead.} Despite having a more expensive optimizer step, we find that MuLoCo's overhead in training time is relatively low at less than 1\%. Despite this, we find that it has better wall-clock training times due to its improved optimization.
  \end{itemize}
\end{generaltakeaway}

\subsection{MuLoCo vs DiLoCo: Increasing K}\label{sec:inner-opt}
Figure~\ref{fig:worker-scaling-vs-h} (a) reports the increase (lower is better) in final validation loss relative to data-parallel as a function of the number of workers (k). It should be noted that on an absolute scale, MuLoCo strictly outperforms DiLoCo in each case due to Muon's stronger performance. However, we also observe that both DiLoCo and MuLoCo outperform their respective data-parallel baselines for $K=1$ and $K=2$ workers. At these worker counts, DiLoCo's improvements over data-parallel are larger than MuLoCo's, suggesting that Muon's curvature-aware steps may reduce the gains available from using an outer momentum. For $K > 2$ workers, we see a clear pattern emerge:\textit{ as the number of workers increases the performance of DiLoCo relative to its data parallel baseline decreases at a faster rate than MuLoCo does relative to Muon DP, demonstrating that MuLoCo's performance scales more favourably as workers are increased.}

\finding{Relative to their data-parallel baselines, MuLoCo scales better than DiLoCo as the number of workers increases. (Fig.~\ref{fig:worker-scaling} (a))}

\textbf{Why does MuLoCo scale better with workers?} Our central hypothesis is that Muon's orthonormalized optimizer steps lead to pseudogradients that are more directionally aligned with the data-parallel/K=1 pseudogradient, allowing MuLoCo to more accurately approximate it at higher worker counts. To validate this, we measure the alignment of MuLoCo's and DiLoCo's pseudogradients with their data-parallel/K=1 pseudogradients. To accomplish this, we use the optimal hyperparameters for DP Muon and AdamW found in our sweeps and train a 416M model saving checkpoints along the way. We subsequently resume training from these checkpoints using DiLoCo or MuLoCo with $K\in\{2,4,8,16\}$, including loading optimizer states, and continue to train using the same hyperparameters and global batch size for H=30 steps, after which we save pseudogradients across all workers to disk. We then measure alignment of pseudogradients at $K>1$ with the DP baseline/K=1 pseudogradient. It is calculated by measuring the consine similarity between all vectorized hidden weights in the network and taking the average of these values.

Figure~\ref{fig:cosinesim-box} reports the cosine similarity as a function of worker count for MuLoCo (a) and DiLoCo (b). It clearly demonstrates that MuLoCo's pseudogradients are consistently more directionally aligned with the the data-parallel/K=1 pseudogradient than DiLoCo. This suggests that Muon's inner steps may play a crucial role in effectively shaping the pseudogradient. Pursuing this hypothesis further, in Figure~\ref{fig:step_and_trajectory_alignment} (a), we report the cosine similarity between a $K=8$ pseudogradient (collected as described above) and the individual optimizer steps that make it up. We find that Muon's optimizer steps are much more aligned with the final pseudogradient direction than AdamW's, suggesting that \textit{Muon optimizer steps are naturally more aligned with one another}. Thus far, we have provided evidence of MuLoCo's more \textit{directionally correct} pseudogradients but we have not explained why DiLoCo's pseudogradients degrade at larger worker counts. We hypothesize that the poorer alignment of AdamW steps may lead to exacerbated information loss during averaging. In an effort to better understand this, we investigate the spectrum of individual worker weight differences $\Delta_k$ before and after averaging them to form a pseudogradient. In Figure~\ref{fig:interferencegap} we show that DiLoCo's pseudogradient spectrum collapses during averaging to a much greater extend that MuLoCo's and that the collapse worsens at larger worker counts. This is corroborated by DiLoCo's relatively stronger performance at K=1 and K=2 workers, when interference is low, with a sharp degradation thereafter. While they do not show a direct causal link, \textit{these findings are consistent with our hypothesis that information loss during averaging is the main mechanism by which DiLoCo pseudogradients degrade.}

\textbf{Why does K=1 DiLoCo perform better relative to DP AdamW than MuLoCo does relative to DP Muon?} Despite always outperforming K=1 DiLoCo in absolute terms, in Figure~\ref{fig:worker-scaling-vs-h} (a) K=1 DiLoCo's improvement is larger than K=1 MuLoCo relative to their respective data-parallel baselines. We hypothesize that this is due to the greater alignment with the final pseudogradient observed for Muon optimizer steps than for AdamW's (Fig.~\ref{fig:step_and_trajectory_alignment}). The outer optimizer in DiLoCo/MuLoCo is analogous Reptile meta-learning, which is known to encourage alignment between different optimizer steps~\citep{reptile}. Therefore, since Muon's optimizer's steps are already well aligned, it stands to reason that the outer optimizer would be less effective at improving its performance than for AdamW.
\begin{figure*}[t]
    \centering
    \subfloat[Linear quantization]{%
        \includegraphics[width=0.48\linewidth]{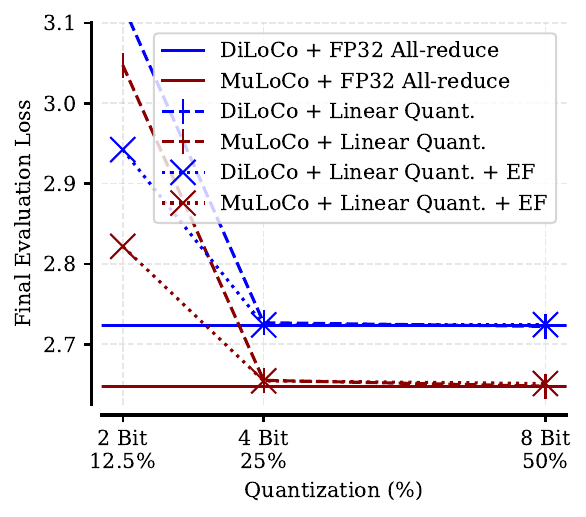}
    }
    \subfloat[Statistical quantization]{%
        \includegraphics[width=0.48\linewidth]{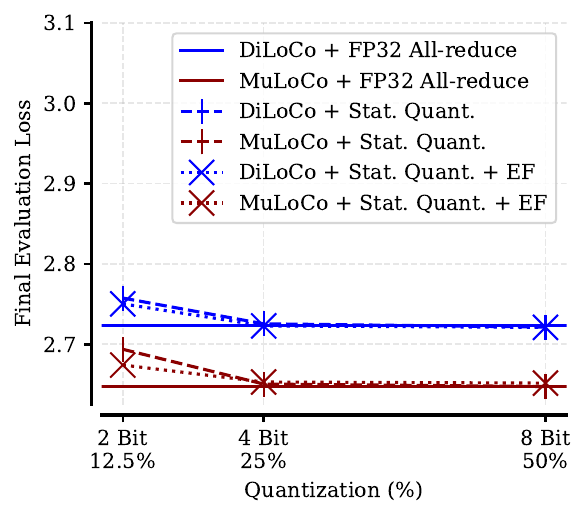}
    }
    \caption{\textbf{Quantized pseudogradient compression: linear vs.\ statistical quantization and the effect of error feedback (EF).}
    Linear quantization (left) and row-wise statistical quantization (right) are shown with and without EF across bitwidths.
    In each case, MuLoCo outperforms DiLoCo. For 4 bits and above, both statistical and linear quantization behave essentially identically with or without EF, suggesting that EF is unnecessary in this regime.
    At 2 bits, statistical quantization is consistently better than linear, indicating that it preserves update quality under more aggressive quantization.
    Overall, EF provides only a modest improvement, except for 2-bit linear quantization, where EF can partially recover performance; however, the remaining degradation is still severe, making 2-bit linear quantization impractical in this setting.}
    \label{fig:quantized_compression}
\end{figure*}

\subsection{MuLoCo vs DiLoCo: Increasing H}
\label{sec:moreh}
In this section, we study the performance of MuLoCo and DiLoCo relative to their data-parallel baselines as the synchronization interval (H) is progressively doubled from $H=15$ to $H=240$. To accomplish this, we train DP baselines in our chinchilla-optimial $8$B token setting for $8160$ steps at batch size $512$ and sequence length $2048$. For DiLoCo and MuLoCo, we fix the number of workers at $K=8$ and vary $H$. Figure~\ref{fig:worker-scaling-vs-h} (b) reports the performance of both methods relative to data-parallel. We observe that for all synchronization intervals tested, MuLoCo outperforms DiLoCo relative to their respective data-parallel baselines. Furthermore, the improvements of MuLoCo over DiLoCo remain mostly constant across $H$ values, showing that the communication frequency no effect on performance differences between DiLoCo and MuLoCo relative to DP.

\finding{MuLoCo maintains a \emph{relative} performance advantage over DiLoCo as the synchronization interval is increased. (Fig.~\ref{fig:worker-scaling-vs-h} (b))}

\label{sec:comp}
\subsection{MuLoCo vs DiLoCo: Pseudogradient Compression}\label{sec:compression}

\begin{figure*}[t]
    \centering
    \subfloat[\textsc{Top-$k$} sparsification]{\includegraphics[width=0.48\linewidth]{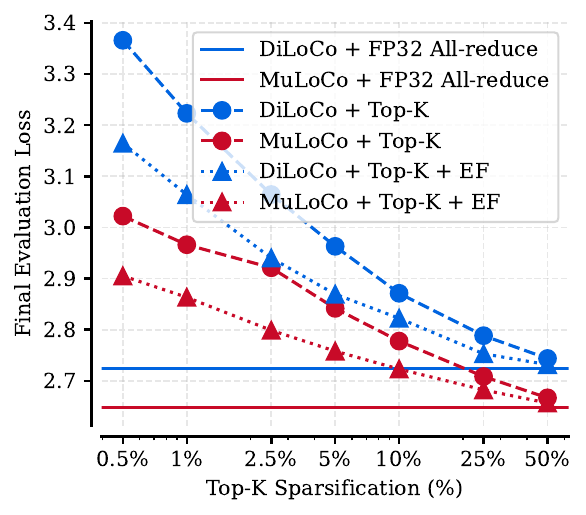}}
    \hfill
    \subfloat[Streaming updates]{\includegraphics[width=0.48\linewidth]{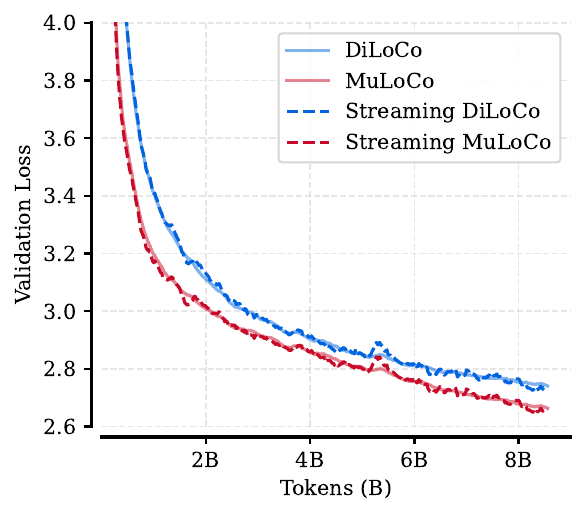}}
    \caption{\textbf{Top-$k$ compression and streaming.}
    \textbf{Left (Top-$k$):} MuLoCo remains more robust than DiLoCo at higher sparsities (i.e., smaller communicated fractions), achieving lower final validation loss at aggressive sparsification levels. Error feedback (EF) improves both methods, but does not make extreme sparsities viable: performance degrades rapidly as sparsity increases.
    \textbf{Right (Streaming):} Streaming and non-streaming variants match closely for both DiLoCo and MuLoCo, indicating no degradation from partitioned (streamed) communication. In particular, MuLoCo tracks DiLoCo under streaming and remains fully compatible with streaming updates.}
    \label{fig:topk_streaming}
\end{figure*}

In this section, we evaluate the performance of MuLoCo and DiLoCo when compression is applied to further reduce communication costs. Specifically, we quantize or sparsify the parameter delta in line 15 of Algorithm~\ref{algo:muloco-ef} before communicating it and ablate using error feedback~\citep{karimireddy2019error}. 

\textbf{Quantization} Figure~\ref{fig:quantized_compression} reports the final evaluation loss of models trained using DiLoCo and MuLoCo with 8-bit, 4-bit, and 2-bit quantization in a K=8, H=30 setting. We experiment with global linear quantization (a) and statistical quantization (b) (see Fig.~\ref{fig:appendix_quantization} for row-wise versions of each algorithm). Our experiments assume that communication is performed using an efficient all-to-all reduce-scatter~\citep{zeropp} and subsequent all-gather to avoid compounding quantization errors incurred by a ring all-reduce with many workers. We observe that MuLoCo outperforms DiLoCo in absolute terms and that both achieve lossless compression up to 4-bits. When compressing to 2-bits there is a substantial performance degradation for Linear quantization but not statistical. In Figure~\ref{fig:appendix_quantization} we report additional experiments where quantization is applied individually to each row of a tensor. We find that this improves the performance of linear quantization at 2-bits and in the statistical case has near identical performance to global quantization.

\finding{Both MuLoCo and DiLoCo achieve lossless 4-bit quantization. (Fig.~\ref{fig:quantized_compression})}

\textbf{Global Quantization v.s. Row-wise quantization.} With simple and efficient implementations in mind, we decided to study both global and row-wise quantization. Global quantization is simple to implement and has minimal metadata overhead; however, it does require computing statistics across the entire tensor, which can be inefficient for large models or when parameters are sharded. Row-wise quantization was chosen for its improved performance characteristics and its affinity for parallelization. While the row-wise approach has the requirement to store metadata for each row, this avoids the need to synchronize metadata and tensor value information during the reduction since everything can be sent at once. It is also amenable to highly parallelized dequantize-reduce-quantize operations on GPUs since each row can be processed independently.  

\textbf{Top-$k$ Sparsification} Figure~\ref{fig:topk_streaming} reports the performance of MuLoCo and DiLoCo with and without error-feedback under top-$k$ sparsification retaining $0.5\%,1\%,2.5\%,5\%,10\%,20\%,$ and $50\%$ of entries. We observe that MuLoCo consistently outperforms DiLoCo and that Error feedback helps prevent performance degradation. However, the degradation in performance can be substantial at larger sparsities. Given the additional overhead of communicating the index, the true compression ratios of vanilla top-$k$ sparsification are higher than the reported sparsity, making the more competitive sparsity ratios of 5-10\% less appealing than using 2-bit quantization with an all-to-all reduce scatter. Nevertheless, if very strong compression is the goal, practitioners can significantly benefit from using MuLoCo with error feedback over DiLoCo.

\subsection{Streaming MuLoCo}
\label{sec:streaming}


In this section, we investigate MuLoCo's performance when using partitioned communication~\citep{streamingdiloco}. Specifically, in our K=8 worker and H=30 communication step setting, we partition the model's parameters into thirds and take outer steps for partitions every 10 steps. This has the effect of reducing the peak communication volume by a factor equal to the number of partitions. Figure~\ref{fig:topk_streaming} (b) reports validation loss curves during training for DiLoCo, MuLoCo, and their streaming counterparts. We observe that both the streaming and non-streaming methods reach the same final validation loss, demonstrating that MuLoCo is compatible with streaming.
\finding{MuLoCo is compatible with the partitioned communication of streaming DiLoCo.}

\begin{figure}[h]
    \centering
    \begin{minipage}[c]{0.5\textwidth}
        \centering
        \includegraphics[width=\linewidth]{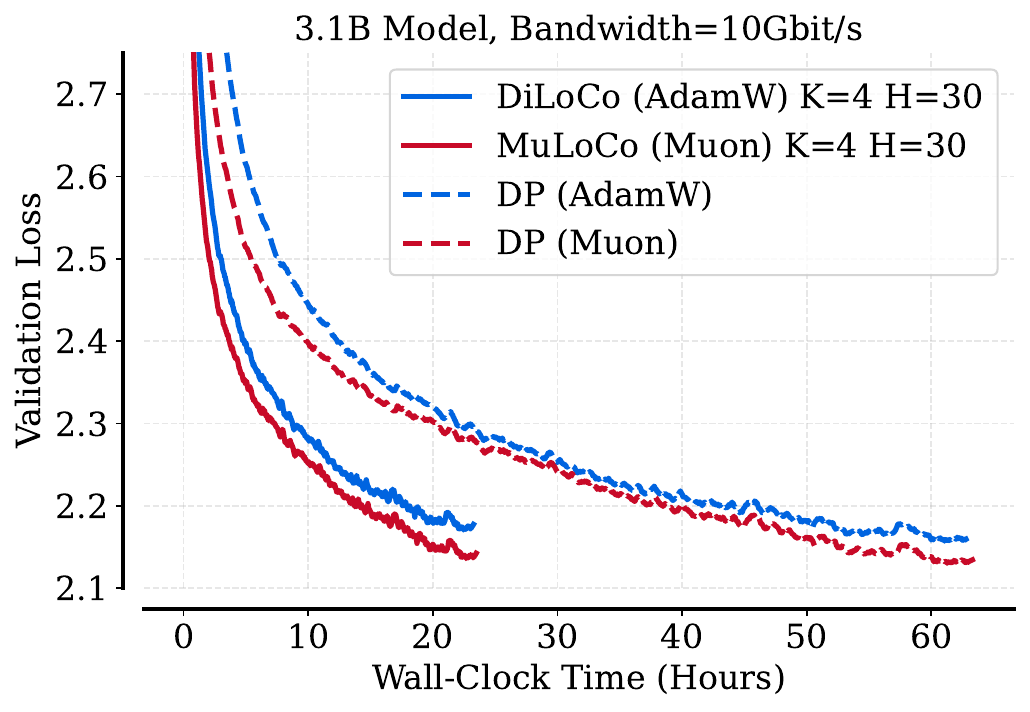}
    \end{minipage}%
    \qquad
    \begin{minipage}[c]{0.45\textwidth}
        \centering
        \small
        \begin{adjustbox}{max width=\linewidth}
        \begin{tabular}{@{}lccc@{}}
        \toprule
        \textbf{Metric} & \textbf{DiLoCo} & \textbf{MuLoCo} & \textbf{$\Delta$ (\%)} \\
        \midrule
        End-to-End Time (s) & 2.82 & 2.85 & +0.96 \\
        MFU (\%) & 43.04 & 42.63 & $-$0.95 \\
        TFLOPS/GPU & 425.7 & 421.7 & $-$0.95 \\
        Tokens/s/GPU & 23,250 & 23,029 & $-$0.95 \\
        Training Time (H) & 23.26 & 23.49 & +0.96 \\\midrule
\textbf{Memory Complexity} & \multirow{2}{*}{4} & \multirow{2}{*}{3} & \multirow{2}{*}{$-$25} \\
\textbf{(Parameter copies)} &  &  &  \\
        \bottomrule
        \end{tabular}
        \end{adjustbox}
    \end{minipage}
    \caption{\textbf{Wall-clock training and system-level performance comparison.} \textbf{Left:} Idealized wall-clock training time under bandwidth constraints (10 Gbit/s) for MuLoCo, DiLoCo, and their DP baselines. DiLoCo completes training slightly faster due to its lower per-step overhead, but MuLoCo's faster convergence and better worker scaling lead to better performance at any training duration. Both methods reach lower loss much faster than their DP baselines due to communication efficiency. \textbf{Right:} System metrics (median over first 500 steps). Configuration: 3.1B model, 2M token batch ($1024 \times 2048$), $K=4$ workers, $H=30$, $4 \times 8$ H100 GPUs. MuLoCo has negligible overhead ($<1\%$) relative to DiLoCo.}
    \label{fig:wallclock}
\end{figure}

\subsection{Wall-clock training time}\label{sec:wallclock}
In this section, we (1) validate the practicality of using Muon as the inner optimizer for DiLoCo, given its more expensive optimizer step, and (2) illustrate the utility of MuLoCo for training under bandwidth constraints. Using the distributed implementation of~\cite{ahn2025dion} in combination with HSDP within TorchTitan~\citep{liang2025torchtitan}, we train a 3.1B model across four 8$\times$H100 nodes using FSDP within nodes and data parallelism across nodes. Table~\ref{fig:wallclock} reports the end-to-end step time, MFU, TFLOPS/GPU, throughput, and training time in a chinchilla optimal regime. We find that the optimizer has negligible training time overhead: $+0.96\%$. In figure~\ref{fig:wallclock} we report wall-clock training curves, demonstrating that Muon's overhead is negligible (MuLoCo dominates DiLoCo) and showing the significant benefits of using MuLoCo over Muon DP in a bandwidth-constrained environment with 10 Gbit/s networks (23 v.s. 60+ hours). 

\finding{MuLoCo has negligible wall-clock overhead relative to DiLoCo. (Tab.~\ref{fig:wallclock})}

\section{Results---Scaling up}\label{sec:results-scaling}
In this section, we investigate whether MuLoCo's performance improvements over DiLoCo hold as model and dataset size are increased under a chinchilla optimal prescription~\citep{chinchilla}. To accomplish this we extensively tune MuLoCo, DiLoCo, and their respective data parallel baselines on 150M-3.1B language model pre-training tasks (see Tab.~\ref{tab:model_details}) and fit power laws to their final evaluation loss. We use a communication interval of $H=30$ steps for these experiments and do not use any compression or streaming. Hyperparameters of all optimizers are extensively tuned at each worker count, $K\in\{1,2,4,8,16\}$. These include the weight decay ($\lambda$), inner learning rate ($\eta_\text{in}$), global batch size (B), outer learning rate ($\eta_\text{out}$), and outer momentum ($\mu$).  We refer the reader to section~\ref{sec:emp-eval} for further details of our hyperparameter tuning process.

The following subsections present the data collected from this study, highlighting the most important results: (\ref{sec:scalingcompute}) projected performance improvement of MuLoCo and DiLoCo at scale, (\ref{sec:scaling-batch-size}) MuLoCo's larger optimal and critical batch sizes, and (\ref{sec:scaling-15b}) evaluation results for training 15B models.

\begin{generaltakeaway}{Takeaways:}
    \begin{enumerate}
        \item \textbf{MuLoCo's improved worker scaling relative to DiLoCo is maintained at scale.} At $K>2$ MuLoCo continues to outperform DiLoCo at scale \textit{even} when normalized by their respective DP baselines.
        \item \textbf{MuLoCo's performance improves relative to data-parallel with scale.} At K=16 workers, the performance gap between MuLoCo and DP Muon shrinks, suggesting that the former will approach the latter's performance at larger scales while being more parallelizable. 
        \item \textbf{MuLoCo has larger critical batch sizes (CBS) than DiLoCo.} Our results at 3.1B scale demonstrate that MuLoCo has a consistently larger CBS than DiLoCo at any K.
        \item \textbf{At iso-loss, K=1 MuLoCo has better training-time-efficiency than DP Muon, K=1 DiLoCo, and DP AdamW.} Fitting scaling laws to performance at CBS, we find that K=1 MuLoCo requires, within observed data, up to $4.5 \times$ fewer sequential FLOPs ($\approx$ training time) to reach the same loss as DP AdamW.
    \end{enumerate}
\end{generaltakeaway}

\subsection{Extrapolating MuLoCo's performance when scaling compute.}\label{sec:scalingcompute}
In this section, we inspect whether MuLoCo's improved performance at higher worker counts extends to larger scales and whether MuLoCo's performance at larger worker counts improves with scale as was found for DiLoCo~\citep{scalingdiloco}.

We first evaluate the fit of three candidate compute scaling laws: (i)~$L(C) = aC^\alpha$, a simple power law; (ii)~$L(C) = aC^\alpha + c$, with a per-optimizer irreducible loss; and (iii)~$L(C) = aC^\alpha + L_0$, with a joint irreducible loss $L_0$ shared across all 12 optimizer/worker combinations while each retains independent $a$ and $b$. 

\textbf{Fitting scaling laws.}  We fit the parameters of all scaling laws by minimizing a Huber loss~\citep{huberloss} in log-space. Specifically, given predictions $\hat{L}i$ and targets $L_i$, we minimize $\sum_i H_\delta(\log \hat{L}i - \log L_i)$ where $H_\delta$ is the Huber loss function with $\delta = 0.001$. This provides robustness to outliers while remaining smooth near zero. Optimization is performed using L-BFGS~\citep{lbfgs} with a maximum of 15,000 iterations per run. To avoid poor local minima, each fit is repeated from 512 random initializations, and the solution with
  the lowest objective is retained. For joint irreducible loss fitting, a three-phase grid search is used: a coarse sweep over 200 candidate $L_\mathrm{irr}$ values (each evaluated with 64 restarts per
  combination), a zoom phase refining the neighborhood of the best candidate, and a final phase re-fitting all combinations at the selected $L_\mathrm{irr}$ with the full 512 restarts.

\begin{table}[h]
    \centering
    \small
    \caption{\textbf{Comparison of compute scaling law forms.} We hold out the 3.1B scale from fitting for each power law. We report the mean absolute log-space residual across all 12 optimizer/worker combinations. We find that adding an irreducible loss substantially improves extrapolation to unseen scales.}
    \label{tab:lc_comparison}
    \begin{tabular}{l|c|cc}
        \toprule
        \multirow{2}{*}{Functional Form} & Train Residual & \multicolumn{2}{c}{Evaluation Residual} \\
        & 150M--1.7B & 3.1B & 15B \\
        \midrule
        $L(C) = AC^\alpha$ & 0.013 & 0.056 & 0.160 \\
        $L(C) = AC^\alpha + c$ & \textbf{0.001} & \textbf{0.011} & \textbf{0.053} \\
        $L(C) = AC^\alpha + L_\mathrm{irr}$ & \textbf{0.001} & \textbf{0.011} & 0.057 \\
        \bottomrule
    \end{tabular}
\end{table}

\begin{figure*}[t]
    \centering
     \subfloat[MuLoCo scaling law fit in compute]{%
        \includegraphics[width=0.48\linewidth]{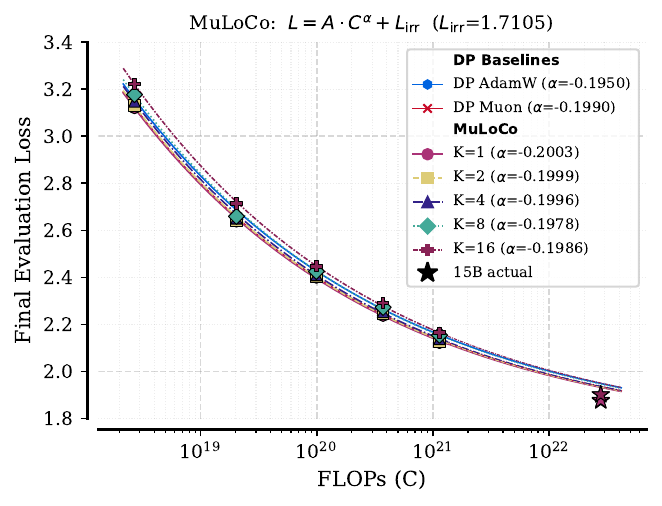}}
        \hfill
    \subfloat[DiLoCo scaling law fit in compute]{%
        \includegraphics[width=0.48\linewidth]{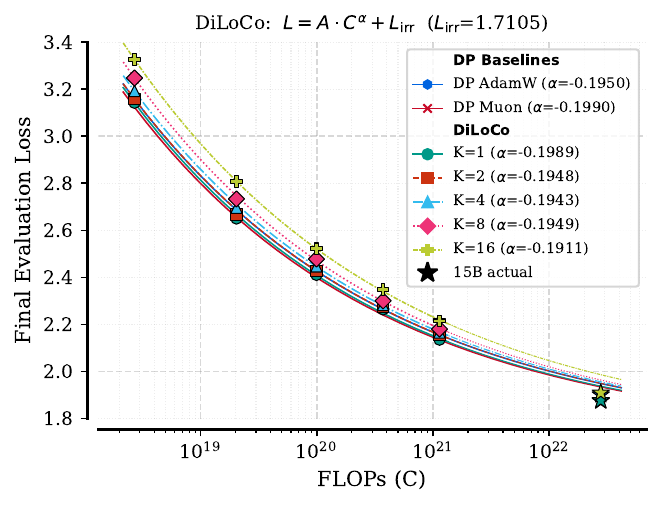}}
   \caption{\textbf{Power-law fits in compute with a shared irreducible loss.}
    We fit power laws of the form $L(C)=A\,C^{\alpha}+L_{\mathrm{irr}}$ to the final evaluation loss as a function of training compute $C$, using a \emph{joint} (shared) irreducible loss $L_{\mathrm{irr}}$ across methods.
    The fitted exponents $\alpha$ are larger for MuLoCo than DiLoco, indicating better performance with scale. We also observe that the fit underestimates the final loss attained by our 15B runs.}
    \label{fig:compute_powerlaws_joint_lirr}
\end{figure*}

\textbf{Evaluating functional forms for L(C).} Table~\ref{tab:lc_comparison} reports residuals from the different scaling laws fit to $150M-1.7B$ datapoints. We hold out 3.1B scale to evaluate the performance of each method. We find that adding an irreducible loss substantially improves extrapolation, reducing the mean 3.1B residual from $0.056$ to $0.011$. Turning to the 15B residuals, we observe that using an irreducible loss also improves extrapolation at this scale, but that the source of error changes: it leads to under-reporting 15B performance when $L(C) = aC^\alpha$ would have over-reported it. We also found that the extrapolation is marginally improved at 15B scale when fitting individual irreducible losses. However, given that $15$B was not swept to optimality and that the assumption of a shared irreducible loss is a reasonable one, we opt to use the power law with a joint irreducible loss.

\begin{figure}[t]
    \centering
    \includegraphics[width=\linewidth]{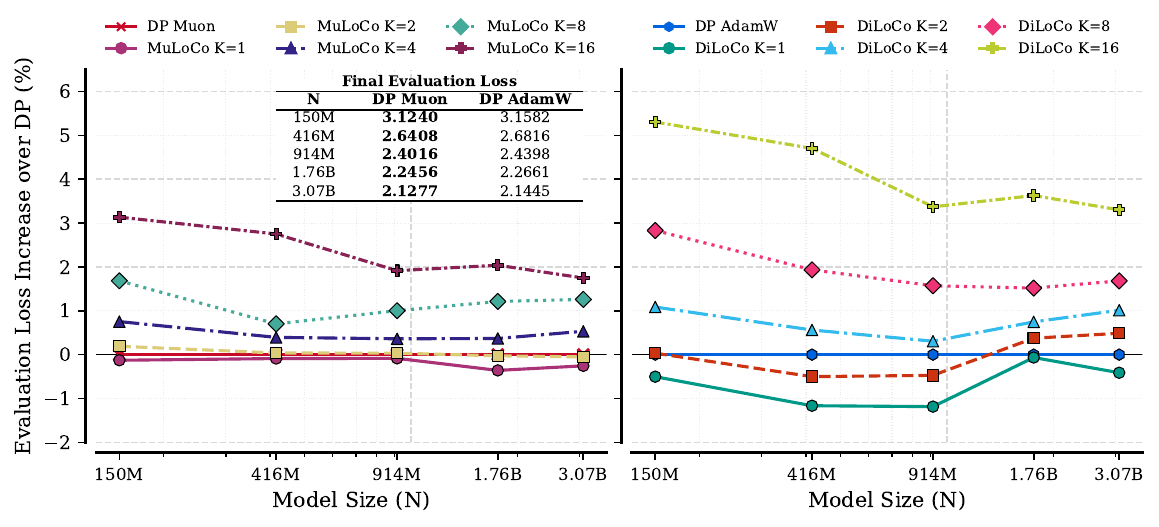}
    \caption{\textbf{Relative performance vs. data-parallel baselines as model size and worker count scale.}
    We report the percent increase in final evaluation loss over data-parallel (DP) for MuLoCo (left) and DiLoCo (right) across model sizes and worker counts $K\in\{1,2,4,8,16\}$. MuLoCo’s performance relative to DP improves at larger worker counts, and we observe the relative gap shrinks with scale for $K=16$ workers. Note that MuLoCo is normalized by \emph{Muon} DP, while DiLoCo is normalized by \emph{AdamW} DP.}
    \label{fig:workerscalingmodelsize}
\end{figure}

\textbf{Final L(C).} We now re-fit the power law with a joint irreducible loss to all five training scales (150M-3.1B).  Table~\ref{tab:lc_joint_params} of the appendix reports the fitted parameters, while Figure~\ref{fig:compute_powerlaws_joint_lirr} plots the corresponding curves. We find that an irreducible loss of $L_\text{irr} = 1.711$ achieves the best fit. The scaling exponents cluster around $\alpha \approx -0.20$ ($\alpha \in [-0.200, -0.191]$), with MuLoCo having generally larger $\alpha$ values than DiLoCo across the board, indicating stronger scaling with compute. When evaluating the extrapolation to $15$B we observe that, as we found above, the power laws consistently underreport performance at larger scale, suggesting that an irreducible loss of $L_\mathrm{irr} = 1.711$ is likely too large in practice. Given this finding, in figure~\ref{fig:exponents_vs_irr_normalized} of the appendix, we fit compute scaling laws at fixed $L_{\mathrm{irr}}$ values shared across all methods. For each $L_{\mathrm{irr}}$, we report the resulting scaling exponent $\alpha$ normalized by the exponent of the corresponding DP baseline. We find that at lower irreducible losses, which are likely to be more realistic than $L_\text{irr} = 1.711$, MuLoCo at higher worker counts is projected reach and eventually outperform data parallel baseline.

\finding{The scaling exponents of Muon-based methods are smaller, demonstrating their stronger scaling.}
\textbf{Performance at different worker counts.} Figure~\ref{fig:workerscalingmodelsize} reports the performance of MuLoCo and DiLoCo normalized by their respective data parallel baselines as a function of model size. First, we find that performance improvements of MuLoCo over DiLoCo, even when normalized by their respective data-parallel baselines, hold at larger scales. Second, we interpret our results within the context of~\citet{scalingdiloco} who found that, at all worker counts ($K$), DiLoCo follows a trend of strict performance improvement relative to DP AdamW as model size is increased. In our experiments, we find the same trend for K=16 MuLoCo and K=16 DiLoCo, suggesting that these optimizers should continue to improve relative to data parallel with scale. However, we don't observe the same strict trend of monotonic performance improvement with model size. While we are uncertain about the exact cause of this discrepancy, we conjecture that it may be caused by one or many of the following (listed in no particular order): differences in hyperparameter tuning, differences in pre-training data, differences in how evaluation loss was calculated, our choice to additionally tune the outer optimizer's momentum, our decision to decay the learning rate to $1/10$, and other experimental discrepancies. 

\finding{At K=16 MuLoCo consistently improves with scale relative to the DP Muon.}
\finding{MuLoCo's improved performance over DiLoCo at $K>2$ \textit{even} when normalized by their respective data-parallel baselines is maintained at scale.}


\subsection{On MuLoCo's large critical batch size}\label{sec:scaling-batch-size}
Maintaining strong performance at large batch sizes is very important for efficiently training the largest models. Training at these batch sizes requires leveraging a larger number of devices in parallel, which can easily span poorly interconnected regions, such as across datacenters or across racks within a datacenter. In such situations, communication-efficient optimizers can be very beneficial. However, \textit{a prerequisite is to be able to leverage large batch sizes to begin with}. Our results in this section will demonstrate that MuLoCo has significant benefits over DiLoCo in this regard.

\begin{figure*}[ht]
    \centering
    \includegraphics[width=0.95\textwidth]{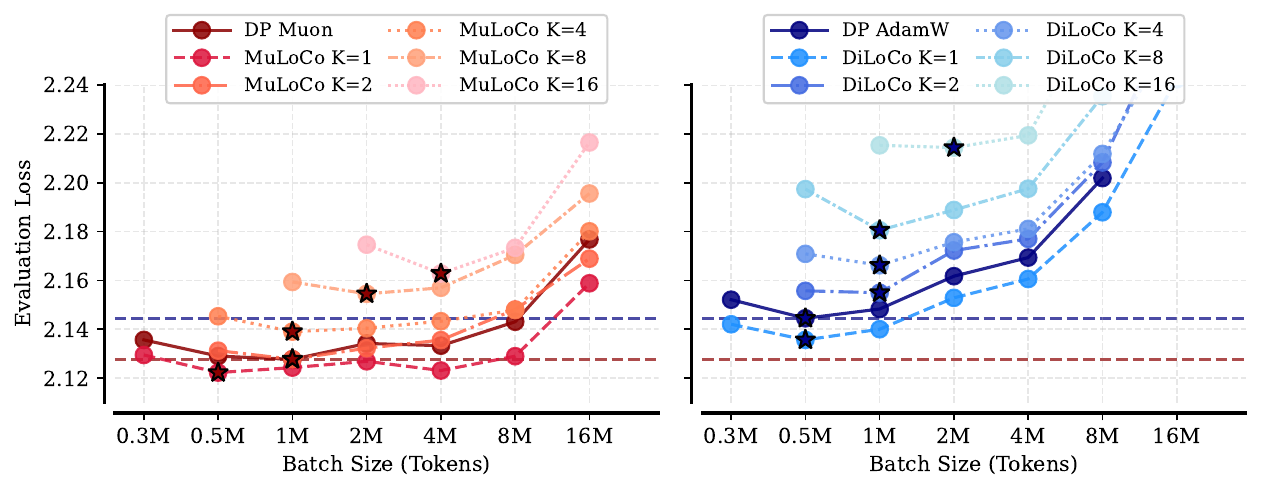}
    \caption{\textbf{MuLoCo has much larger critical batch sizes than DiLoCo.}
    We train 3.1B-parameter models and \emph{tune hyperparameters separately at each batch size} for every method, reporting the \emph{final evaluation loss} at the end of training. We observe that MuLoCo has larger optimal batch sizes at $K>1$ and a much larger critical batch size at all worker counts, enabling substantially more parallelizable training (potentially large enough to require cross-datacenter training). In contrast, DiLoCo degrades rapidly as batch size increases, leading to a much smaller usable batch-size regime.}
    \label{fig:batchsize_3p1b_joint}
\end{figure*}

\textbf{Critical batch size at 3.1B.} Figure~\ref{fig:batchsize_3p1b_joint} reports the final evaluation loss of 3.1B models tuned at each batch size. We observe that AdamW degrades substantially as the batch size increases, while MuLoCo maintains strong performance relative to its optimal batch size until $\sim8$M tokens at all worker counts. MuLoCo's increased critical batch sizes result in increasingly parallelizable training. At the largest scales, taking full advantage of this parallelism to reduce training time will likely require training across datacenters, \textit{leading to communication bottlenecks for DP Muon, but \textbf{not} for MuLoCo.} Even when communication is not a bottleneck, MuLoCo at K=1 still offers a notable advantage over Muon DP, maintaining stronger performance at larger batch sizes. This is particularly noticeable at $B=8$M where MuLoCo K=1 matches the optimal performance of Muon DP, achieved at a much smaller $\bopt=1$M.

\finding{MuLoCo's critical batch size at 3.1B scale is larger than DiLoCo's at all worker counts ($K$).}

\textbf{Extrapolating the Critical Batch Size.} Given that the critical batch size is known to scale with training data~\citep{bergsma2025power,cbszhang}, we fit power laws $\bcrit(D) = aD^\alpha$ to extrapolate the critical batch size ($\bcrit$) as training data (D) is increased. With $\bopt$ being the optimal batch size, we take $\bcrit$ to be the largest batch size such that $L(\bcrit)\leq 0.01 L(\bopt)$. We exclude MuLoCo and DiLoCo at $K\neq 1$ in these experiments as their batch sizes were not sufficiently swept to determine $\bcrit$ at each scale. The power laws are fit using L-BFGS as described in the previous section. Figure~\ref{fig:cbs_and_efficiency} (a) plots the power law fit for Muon DP, AdamW DP, MuLoCo K=1, and DiLoCo K=1. We observe that Muon DP and MuLoCo have much larger scaling exponents than DiLoCo and AdamW DP. This demonstrates that their critical batch sizes will grow with scale at a faster rate. We also find that MuLoCo and DiLoCo both have larger prefactors than their data-parallel baselines, meaning they will have a slightly larger critical batch size at each scale.
\finding{MuLoCo K=1 is projected to have larger critical batch sizes than DP Muon at scale.}

\newcommand{\Bopt}{B_{\mathrm{opt}}}
\newcommand{\Lcrit}{L_{\mathrm{irr}}}
\newcommand{\Bcrit}{B_{\mathrm{crit}}}
\newcommand{\Ladamw}{L_{\mathrm{AdamW}}}
\newcommand{\Lopt}{L_{\mathrm{opt}}}
\newcommand{\Tadamw}{T_{\mathrm{AdamW}}}
\newcommand{\Topt}{T_{\mathrm{opt}}}
\newcommand{\Cadamw}{C_{\mathrm{AdamW}}}
\newcommand{\Copt}{C_{\mathrm{opt}}}

\begin{figure*}[h!]
    \centering
    \subfloat[Iso-loss training-time efficiency]{%
        \includegraphics[width=0.48\linewidth]{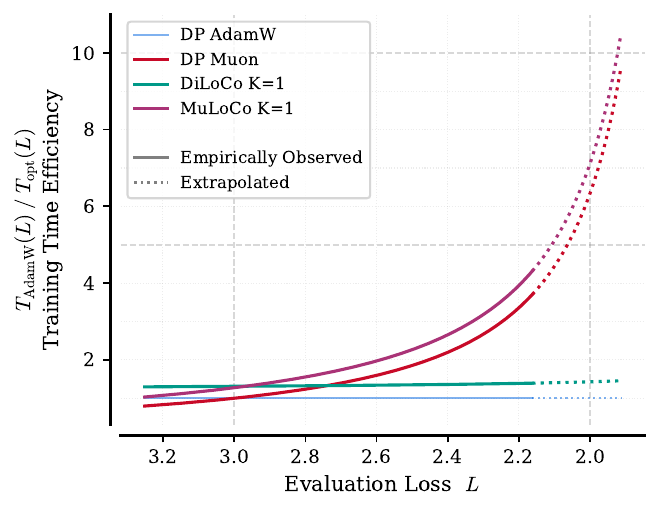}
    }
    \hfill
    \subfloat[Critical batch-size scaling]{%
        \includegraphics[width=0.48\linewidth]{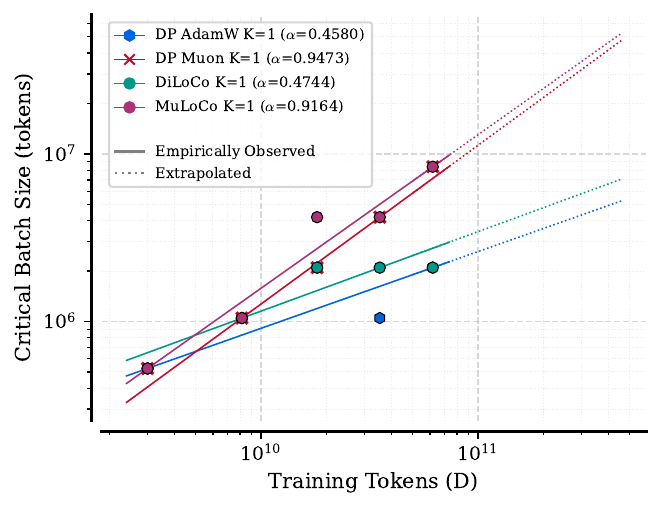}
    }
    \caption{\textbf{MuLoCo achieves substantial training-time gains at iso-loss due to its large critical batch sizes.}
    (left) Reports the iso-loss training-time efficiency of different optimizers relative to DP AdamW. The ratio $\Tadamw(L)/\Topt(L)$ measures how many more sequential FLOPs AdamW requires to reach the same loss~$L$, assuming sufficient parallel and well-interconnected accelerators to realize the speedup from larger critical batch sizes. At observed scales, MuLoCo reaches the same loss as AdamW in roughly ${\sim}4.5\times$ fewer sequential FLOPs.
    (right) Power-law fits of the critical batch size (CBS) versus training tokens show that MuLoCo’s CBS is larger than Muon data-parallel (DP) and significantly larger than AdamW DP and DiLoCo, indicating a wider regime of near-lossless batch scaling.
    \emph{Dotted curves indicate extrapolation beyond the range of available data.}}
    \label{fig:cbs_and_efficiency}
\end{figure*}


\textbf{Training time efficiency at iso-loss.}
The largest-scale pre-training runs often take over a month of training time to complete. They are primarily bottlenecked by the number of sequential gradient descent steps required for strong performance, which is directly determined by batch size used. While the strongest possible final performance is achieved at the optimal batch size ($\Bopt$), to maintain reasonable training time costs for the largest runs, practitioners generally train at the critical batch size ($\Bcrit$), since with enough training data $\Bcrit>>\Bopt$. Therefore, optimizers should not only be compared w.r.t their best possible performance but also based on their training time to loss at $\Bcrit$.  A reasonable proxy for training time is $T \propto C / B_{\mathrm{crit}}(C)$, the amount of FLOPs that must be computed sequentially when training at $\Bcrit$~\citep{bergsma2025power}. In Figure~\ref{fig:cbs_and_efficiency} (a), we directly compare the iso-loss training-time efficiency of different optimizers relative to DP AdamW, finding that K=1 MuLoCo has the strongest training-time advantage.

To accomplish this, for each optimizer $m$ we fit two scaling laws from our training runs: a loss curve $L_m(C) = a_m C^{b_m} + L_0$ with a shared irreducible loss $L_0$ across all $K{=}1$ methods, and a critical batch size scaling law $B_{\mathrm{crit},m}(D) = A_m D^{\alpha_m}$, where $D$ denotes training tokens and we use the Chinchilla-optimal relationship $D = 20N$, $C = 6ND$. Given a target loss $L$, we invert the loss scaling law to obtain the compute $C_m(L)$ required by each optimizer, and then evaluate the corresponding training-time proxy $T_m(L) = C_m(L) / B_{\mathrm{crit},m}(C_m(L))$. The iso-loss training-time efficiency is then defined as $\Tadamw(L)/\Topt(L)$: a value greater than one indicates that optimizer $m$ reaches the same loss as AdamW in fewer sequential FLOPs.

This metric captures the practical advantage of optimizers with larger critical batch sizes: they can absorb more data-parallel workers without diminishing returns, training faster in wall-clock time for a fixed target loss. Equivalently, the iso-loss training-time ratio decomposes into a compute-efficiency term and a parallelism term:
\begin{equation}
    \frac{\Tadamw(L)}{\Topt(L)} \;=\; 
    \underbrace{\frac{\Cadamw(L)}{\Copt(L)}}_{\text{compute savings}} 
    \;\times\; 
    \underbrace{\frac{\Bcrit^{(m)}\bigl(C_m(L)\bigr)}
    {\Bcrit^{(\mathrm{AdamW})}\bigl(\Cadamw(L)\bigr)}}_{\text{parallelism advantage}}\,.
\end{equation}
We find that K=1 MuLoCo reaches the same loss as DP AdamW in roughly ${\sim}4.5\times$ fewer sequential steps at observed scales. This advantage is much larger than its compute-only savings because MuLoCo combines improved compute scaling with a substantially larger critical batch size.

\finding{At iso-loss, K=1 MuLoCo reaches the same loss as DP AdamW in roughly ${\sim}5\times$ fewer sequential steps at observed scales.}

\textbf{Pareto optimal training-time performance tradeoff.}
The iso-loss analysis above shows that MuLoCo can reach a target loss in substantially fewer sequential FLOPs than DP AdamW by combining improved compute efficiency with a larger critical batch size. However, this comparison is not FLOP-matched: because MuLoCo is itself more FLOP-efficient, it generally reaches lower loss than AdamW at the same compute budget before any training-time advantage from larger batch sizes is considered. We therefore also examine the FLOP-matched setting in Figure~\ref{fig:worker-scaling} (b), which compares final evaluation loss as a function of FLOPs per batch size, a proxy for sequential training time. At the 3.1B scale, MuLoCo K=1 lies on the Pareto frontier: it achieves stronger final performance than AdamW DP, Muon DP, and DiLoCo K=1, while also supporting the largest critical batch size. Notably, at its critical batch size, MuLoCo matches the performance of DP Muon at its optimal batch size and outperforms AdamW and DiLoCo at all batch sizes. Thus, among the optimizers compared, MuLoCo K=1 provides the best training-time-performance tradeoff in the FLOP-matched regime.

\finding{In a FLOP-matched comparison, K=1 MuLoCo lies on the Pareto frontier between final performance and training time, achieving the strongest loss while supporting the largest critical batch size.}

\subsection{Scaling up to 15B}\label{sec:scaling-15b}
In this section, we use the datapoints from our scaling study to extrapolate optimal hyperparameters to 15B scale, train with these hyperparameters, and evaluate the performance of the resulting models. While reporting the optimal performance is not guaranteed for each method since we did not sweep at this scale, these experiments provide us with an understanding of the performance attainable from different optimizers, an understanding of how large batch training and communication overhead contributed to wall-clock training time, and an increased confidence in the scalability of our method.

\textbf{Extrapolating optimal hyperparameters.} Using the data collected in our scaling experiments, we fit three new power laws: (1) $\lambda(N) = aN^\alpha$ (weight decay), (2) $\eta(N) = aN^\alpha$ (learning rate), and (3) $\bopt(D) = aD^\alpha$ (optimal batch size). Following our tuning protocol from section~\ref{sec:emp-eval}, the weight decay power law is fit to the optimal value found at $B=1$M and extrapolated in this regime. The optimal value found is subsequently rescaled according to the extrapolated optimal batch size. For learning rate and batch size we simply fit power laws to their optimal values for each optimizer. It should be noted that our power laws for with $\bopt$ were fit with K=1 MuLoCo, having $\bopt=4$M at 3.1B scale. However, further sweeping later revealed that this was a local minimum (see Fig.~\ref{fig:batchsize_3p1b_joint}). This means that K=1 MuLoCo's batch size used at 15B scale is likely closer to the $\bcrit$ than $\bopt$ and that K=1 MuLoCo's optimal final performance is likely slightly under-reported. Further details in addition to the optimal values used are reported in~\ref{sec:apdx:15b-tuning} of the appendix.

\begin{table}[ht]
\centering
\caption{\textbf{Zero-shot LM Evaluation Results for K=1 and K=16 optimizers at 15B scale.} We extrapolate optimal hyperparameters for K=1 and K=16 optimizers and train 15B parameter models with them. We report accuracy for MMLU, WinoGrande, and normalized accuracy for HellaSwag, PIQA, ARC-E, ARC-C, OBQA. }
\label{tab:lm_eval_complete_15b}
\begin{adjustbox}{width=\textwidth}
\begin{tabular}{cl|c|ccccccc|c}
\toprule
\textbf{Model Size} & \textbf{Optimizer} & \textbf{Eval. Loss} & \textbf{MMLU} & \textbf{HellaSwag} & \textbf{WinoGrande} & \textbf{PIQA} & \textbf{ARC-E} & \textbf{ARC-C} & \textbf{OBQA} & \textbf{Mean Acc.} \\
\midrule
15B & DP AdamW & 1.8873 & 57.20\% & 77.69\% & 71.19\% & 81.39\% & 78.11\% & 52.30\% & 45.40\% & 66.18\% \\
15B & DP Muon & \textbf{1.8642 }& 59.14\% & 79.05\% & 71.19\% & 81.12\% & 77.23\% & 53.07\% & 46.80\% & \textbf{66.80}\% \\\midrule
15B & DiLoCo K=1 & 1.8914 & 59.80\% & 79.06\% & 71.43\% & 81.34\% & 77.99\% & 52.56\% & 46.40\% & 66.94\% \\
15B & MuLoCo K=1 & \textbf{1.8839} & 59.49\% & 79.06\% & 71.11\% & 81.61\% & 78.11\% & 52.73\% & 46.60\% & \textbf{66.96}\% \\\midrule
15B & DiLoCo K=16 & \textbf{1.9062} & 54.15\% & 77.59\% & 69.30\% & 81.72\% & 77.65\% & 52.56\% & 46.80\% & 65.68\% \\
15B & MuLoCo K=16 & 1.9168 & 56.96\% & 77.18\% & 70.72\% & 80.74\% & 76.60\% & 53.16\% & 46.20\% & \textbf{65.94}\% \\
\bottomrule
\end{tabular}
\end{adjustbox}
\end{table}

\textbf{Evaluating final performance.} Table~\ref{tab:lm_eval_complete_15b} reports the final loss values obtained for models trained with different optimizers in addition to their zero-shot performance across MMLU, HellaSwag, WinoGrande, PIQA, ARC-E, ARC-C, and OBQA. Across optimizers, we find that all methods converge to \textit{nearly} the same average downstream accuracy, with MuLoCo ($K{=}1$), DiLoCo ($K{=}1$), and data-parallel Muon performing marginally stronger, while the $K{=}16$ variants and DP AdamW perform slightly worse. Importantly, as shown in the per-step view of Figure~\ref{fig:15b_validation_combined}, MuLoCo ($K{=}1$) trains at the largest batch size (17M) while reaching nearly identical final evaluation loss and evaluation performance to DiLoCo ($K{=}1$) and DP Muon both trained for many more steps. This demonstrates that MuLoCo $K=1$ can leverage more parallelism without sacrificing performance. A similar advantage holds for MuLoCo ($K{=}16$), which reached comparable performance to DiLoCo ($K{=}16$) and DP AdamW while also having a larger batch size than both.

\textbf{Evaluating wall-clock training time when accounting for batch size, optimizer step times, and communication overhead.} We also collected system-level metrics during training, enabling wall-clock analyses under networking constraints (Figure~\ref{fig:bandwidth_wall_clock}). In bandwidth-constrained regimes (10 Gbit/s), the communication-efficient optimizers dominate: both $K{=}16$ methods train substantially faster than $K{=}1$ MuLoCo, despite their smaller batch sizes, with MuLoCo ($K{=}16$) emerging as the fastest configuration due to its larger extrapolated optimal batch size. As bandwidth increases (800 to 12,800 Gbit/s), communication becomes less of a bottleneck and leveraging more compute in parallel becomes more important. In the high bandwidth regime, the ability of MuLoCo ($K{=}1$) to leverage very large batch sizes becomes advantageous, leading it ultimately train the fastest.

\finding{When training under bandwidth constraints, K=16 MuLoCo has the fastest wall-clock training time, while K=1 MuLoCo is fastest in high-bandwidth settings.  }

\begin{figure*}[ht]
    \centering
    \includegraphics[width=0.98\linewidth]{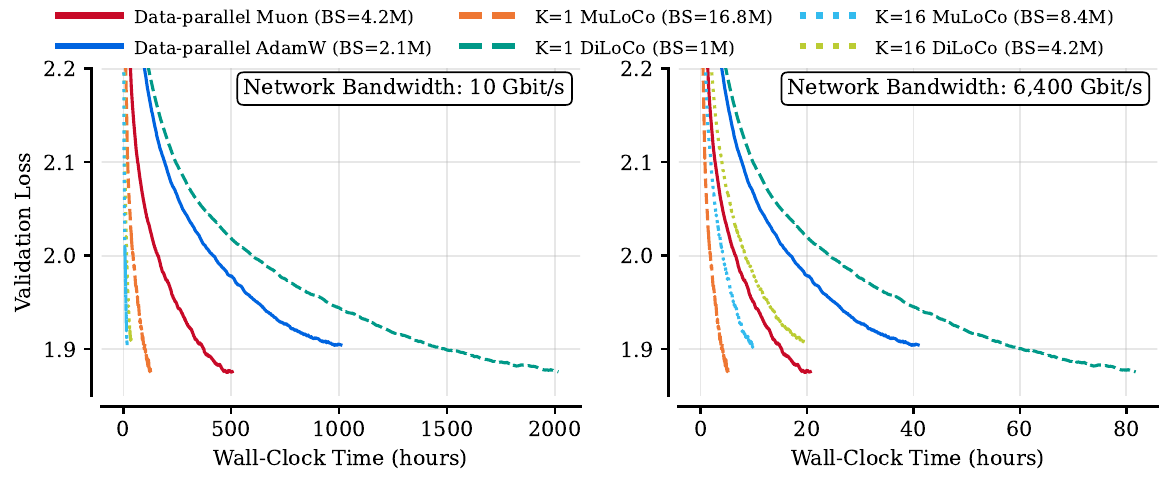}
    \caption{\textbf{Idealized wall-clock training time for 15B runs under low- and high-bandwidth networking.}
    We estimate training time by combining (i) network communication time, (ii) optimizer step time, and (iii) forward/backward (FW/BW) compute time.
    The left panel reports validation loss versus wall-clock time assuming a low-bandwidth network, while the right panel reports the same quantities in a datacenter setting with high bandwidth.
    In the low-bandwidth regime, \textbf{K=16 MuLoCo} achieves the fastest time-to-loss and dominates the DiLoCo baselines; in the datacenter regime, \textbf{K=1 MuLoCo} achieves the fastest time-to-loss.
    In both cases, MuLoCo’s larger batch sizes translate into faster wall-clock training to comparable loss than corresponding DiLoCo configurations.}
    \label{fig:15b_wallclock_low_vs_dc}
\end{figure*}

\section{Limitations}
We have made our best effort to conduct a sound empirical study. However, as with most empirical works, we could not exhaustively run all relevant experiments, leading to natural limitations from settings that were not tried. Listed in no particular order, we acknowledge that out study has the following limitations. Firstly, our exploration of MuLoCo's performance at scale is limited to experiments training 150M-15B dense transformers in a 20TPP setting. Therefore, the scaling trends we predict may not hold for models that are several orders of magnitude larger. Secondly, our study is limited to studying dense transformers with no exploration of other popular foundation model architectures (e.g., MoEs, SSMs, or Hybrid architectures). Thirdly, we do not explore outer optimizers beyond Nesterov SGD. Finally, we do not prove convergence of the method. 
\section{Conclusion}
In conclusion, we have compared the performance of DiLoCo to MuLoCo across a range of workloads relevant to distributed optimization and as training is scaled up. We established that MuLoCo outperforms DiLoCo under long communication intervals, compressed communication, when using streaming updates, and even in wall-clock training time. Beyond improving over DiLoCo in absolute terms, we even found that MuLoCo's performance relative to DP Muon scales better than DiLoCo's performance relative to DP AdamW as the number of workers increases. Furthermore, we found that MuLoCo's critical batch size is larger than DiLoCo's at all worker counts, leading to faster wall-clock training times and possibly warranting training across datacenters to take advantage of the additional data parallelism. Finally, we found that K=1 MuLoCo has a Pareto-optimal performance-training-time tradeoff among all optimizers in our study. Overall, our results demonstrate that MuLoCo is a flexible distributed optimization algorithm. With its large critical batch sizes, stronger performance than DP Muon at K=1, and its improved worker scaling over DiLoCo, MuLoCo has significant advantages for large-scale training in both low-bandwidth and high-bandwidth settings.



\section*{Acknowledgements} 
We are particularly grateful to Lin Xiao for providing feedback on our manuscript and guidance on our experiments throughout. We thank Jack Rae and Tom Gunter for providing practical guidance regarding our scaling law and critical batch size experiments. We thank Andrei Mircea and Ekaterina Lobacheva for insightful discussions about our pseudogradient analysis experiments. We thank Howard Huang and Tushar Jain for their advice in resolving TorchTitan and TorchFT bugs. We would also like to thank Zeyuan Allen-Zhu, Konstantin Mishchenko, Ashok Cutkosky, Ziqui Bu, Alice Yang, Lisa jin, Andrey Gromov, Maissam Barkeshli, and Sanae Lotfi for fruitful comments and research discussions.  

We acknowledge support FRQNT New Scholar [\emph{E.B.}], the FRQNT Doctoral (B2X) scholarship [\emph{B.T.}], the Canada CIFAR AI Chair Program [\emph{I.R.}], and the Canada Excellence Research Chairs Program in Autonomous AI [\emph{I.R.}].

\bibliographystyle{assets/plainnat}
\bibliography{tmlr}

\clearpage
\newpage
\beginappendix

\tableofcontents

\clearpage
\section{MuLoCo Algorithm Block}
We present non-compressed and compressed versions of MuLoCo in Algorithms~\ref{algo:muloco} and~\ref{algo:muloco-ef}, respectively. The compressed version of the algorithm is used in the compression experiments of section~\ref{sec:compression}, while Algorithm~\ref{algo:muloco} is used elsewhere.

\begin{minipage}[t]{0.47\textwidth}
\begin{algorithm}[H]
\caption{MuLoCo (No Compression).}
\label{algo:muloco}
\footnotesize
\begin{algorithmic}[1]
  \STATE \textbf{Input:} $N$ (communication steps), $K$ (workers), $H$ (local steps), $t$ (local step counter), data shards $\{\mathcal{D}_1,\ldots,\mathcal{D}_K\}$, initial model $\theta^{(0)}$, outer momentum $\mu$, outer LR $\eta_{\text{out}}$

  \STATE $u^{(0)} \gets 0$;\;\; $\theta_i^{(1)} \gets \theta^{(0)} \ \ \forall i$;\;\;$t \gets 0$

  \FOR{$n = 1 \ldots N$ }
    \FOR{$i = 1 \ldots K$ \textbf{in parallel} }
      \FOR{$h = 1 \ldots H$ }
        \STATE $t \gets t + 1$; $\;\;(x,y) \sim \mathcal{D}_i$; $\;\;\mathcal{L} \gets f_\theta(x,y;\theta_i^{(t)})$
        \STATE $\theta_i^{(t)} \gets \texttt{Muon}(\theta_i^{(t)}, \nabla \mathcal{L})$ 
      \ENDFOR
      \STATE $\Delta_i^{(t)} \gets \theta^{(t-H)} - \theta_i^{(t)}$ \; \textit{(parameter delta)}
    \ENDFOR
    \STATE $\Delta^{(t)} \gets \frac{1}{K} \sum_{i=1}^{K} \Delta_i^{(t)}$ \; \textit{(all-reduce outer delta)}
    \STATE $u^{(t)} \gets \mu u^{(t-H)} + \eta_{\text{out}} \Delta^{(t)}$ \; \textit{(Outer Momentum)}
    \STATE $\theta^{(t)} \gets \theta^{(t-1)} - \mu u^{(t)} - \eta_{\text{out}} \Delta^{(t)}$ \; \textit{(Outer Update)}
  \ENDFOR
\end{algorithmic}
\end{algorithm}
\end{minipage}%
\hfill
\begin{minipage}[t]{0.50\textwidth}
\begin{algorithm}[H]
\caption{MuLoCo with compressed communication and optional EF.}
\label{algo:muloco-ef}
\footnotesize
\begin{algorithmic}[1]
  \STATE \textbf{Input:} $N$ (communication steps), $K$ (workers), $H$ (local steps), $t$ (local step counter), data shards $\{\mathcal{D}_1,\ldots,\mathcal{D}_K\}$, initial model $\theta^{(0)}$, {\color{darkred}compressor $\gC$}, outer momentum $\mu$, outer LR $\eta_{\text{out}}$, {\color{darkred}error feedback flag \texttt{EF}}

  \STATE $u^{(0)} \gets 0$;\;\; $\theta_i^{(1)} \gets \theta^{(0)} \ \ \forall i$;\;\;$t \gets 0$
  \color{darkred}%
  \IF{\texttt{EF}}
  \STATE $\gE_i^{(0)} \gets 0 \ \ \forall i$
  \ENDIF
  \color{black}%

  \FOR{$n = 1 \ldots N$ }
    \FOR{$i = 1 \ldots K$ \textbf{in parallel} }
      \FOR{$h = 1 \ldots H$ }
        \STATE $t \gets t + 1$; $\;\;(x,y) \sim \mathcal{D}_i$; $\;\;\mathcal{L} \gets f_\theta(x,y;\theta_i^{(t)})$
        \STATE $\theta_i^{(t)} \gets \texttt{Muon}(\theta_i^{(t)}, \nabla \mathcal{L})$ 
      \ENDFOR
      \STATE $\Delta_i^{(t)} \gets \theta^{(t-H)} - \theta_i^{(t)}$ \; \textit{(parameter delta)}
      \color{darkred}%
      \IF{\texttt{EF}}
      \STATE $\gE_i^{(t)} \gets \beta\,\gE_i^{(t-H)} + \Delta_i^{(t)}$
      \STATE $\Delta_i^{(t)} \gets \gC\!\big(\gE_i^{(t)}\big)$ \; \textit{(compress EF)}
      \STATE $\gE_i^{(t+1)} \gets \gE_i^{(t)} - \Delta_i^{(t)}$
    \ELSE
      \STATE $\Delta_i^{(t)} \gets \gC\!\big(\Delta_i^{(t)}\big)$ \; \textit{(compress parameter delta)}
    \ENDIF
      \color{black}%
    \ENDFOR
    \STATE $\Delta^{(t)} \gets \frac{1}{K} \sum_{i=1}^{K} \Delta_i^{(t)}$ \; \textit{(all-reduce outer delta)}
    \STATE $u^{(t)} \gets \mu u^{(t-H)} + \eta_{\text{out}} \Delta^{(t)}$ \; \textit{(Outer Momentum)}
    \STATE $\theta^{(t)} \gets \theta^{(t-1)} - \mu u^{(t)} - \eta_{\text{out}} \Delta^{(t)}$ \; \textit{(Outer Update)}
  \ENDFOR
\end{algorithmic}
\end{algorithm}
\end{minipage}

\section{Proof of Proposition 4.2}\label{apdx:proof}

\paragraph{Notation.}
For a matrix $A \in \mathbb{R}^{m \times n}$, we denote by $\sigma_1(A)\ge\dots\ge\sigma_r(A)\ge 0$ its singular values, with $r := \min\{m,n\}$. The Frobenius inner product is $\langle A,B\rangle_F := \mathrm{Tr}(A^\top B)$, and the Frobenius norm is $\|A\|_F := \sqrt{\langle A,A\rangle_F}$. The nuclear norm is $\|A\|_\ast := \sum_{j=1}^r \sigma_j(A)$.

\begin{proposition}[Pseudogradient nuclear norm depends on optimizer step alignment]
Fix $m,n\in\mathbb{N}$ and let $r:=\min\{m,n\}$. Consider a pseudogradient of the form
\[
\Psi \;=\;\frac{1}{K}\sum_{k=1}^K\sum_{h=1}^H \alpha_{t+h}\,\psi^{(t+h,k)}\in\mathbb{R}^{m\times n},
\qquad \alpha_{t+h}\ge 0.
\]
Let $\Psi = U\Sigma V^\top$ be its singular value decomposition and define the orthonormal factor $\Psi^\star := U V^\top.$ Let the cosine similarity between a given optimizer step and $\Psi^\star$ be
\[
\rho^{(t+h,k)} \;:=\;
\frac{\langle \psi^{(t+h,k)},\Psi^\star\rangle_F}{\|\psi^{(t+h,k)}\|_F\;\|\Psi^\star\|_F}.
\]
Then the nuclear norm of $\Psi$ satisfies
\[
\|\Psi\|_\ast
=
\frac{\sqrt{r}}{K}\sum_{k=1}^K\sum_{h=1}^H \rho^{(t+h,k)}\alpha_{t+h}\,\|\psi^{(t+h,k)}\|_F.
\]
\end{proposition}

\proof{
Let $\Psi = U\Sigma V^\top$ be the singular value decomposition of $\Psi$, and recall the definition of its \emph{orthonormal factor} $\Psi^\star := UV^\top$.

We first show that $\langle \Psi,\Psi^\star\rangle_F = \|\Psi\|_\ast$. Indeed,
\begin{align}
\langle \Psi,\Psi^\star\rangle_F
&= \mathrm{Tr}\!\left(\Psi^\top \Psi^\star\right)
= \mathrm{Tr}\!\left((U\Sigma V^\top)^\top (UV^\top)\right) \nonumber \\
&= \mathrm{Tr}\!\left((V\Sigma U^\top)(UV^\top)\right)
= \mathrm{Tr}\!\left(V\Sigma (U^\top U) V^\top\right) \nonumber \\
&= \mathrm{Tr}\!\left(V\Sigma V^\top\right)
= \mathrm{Tr}(\Sigma)
= \sum_{j=1}^r \sigma_j(\Psi)
= \|\Psi\|_\ast. \label{eq:initbound_novonneumann}
\end{align}

By linearity of the Frobenius inner product, we can expand
\begin{align}
\langle\Psi,\Psi^\star\rangle_F
&= \left\langle \frac{1}{K}\sum_{k=1}^K\sum_{h=1}^H\alpha_{t+h}\psi^{(t+h,k)},\;\Psi^\star\right\rangle_F \nonumber \\
&= \frac{1}{K}\sum_{k=1}^K\sum_{h=1}^H \alpha_{t+h}\,\langle \psi^{(t+h,k)},\Psi^\star\rangle_F. \label{eq:lin_expand}
\end{align}

Substituting the definition of cosine similarity,
\[
\rho^{(t+h,k)} \;:=\;
\frac{\langle \psi^{(t+h,k)},\Psi^\star\rangle_F}{\|\psi^{(t+h,k)}\|_F\;\|\Psi^\star\|_F},
\]
we obtain
\begin{align}
\langle\Psi,\Psi^\star\rangle_F
&= \frac{1}{K}\sum_{k=1}^K\sum_{h=1}^H\rho^{(t+h,k)}\alpha_{t+h}\|\psi^{(t+h,k)}\|_F\|\Psi^\star\|_F. \label{eq:cos_sub}
\end{align}

Finally, since $\Psi^\star = UV^\top$ has singular values equal to $1$ for $j\le r$, we have
\[
\|\Psi^\star\|_F = \sqrt{\sum_{j=1}^r \sigma_j(\Psi^\star)^2} = \sqrt{r}.
\]
Combining this with \eqref{eq:initbound_novonneumann}, \eqref{eq:cos_sub} yields
\[
\|\Psi\|_\ast
= \langle\Psi,\Psi^\star\rangle_F
= \frac{\sqrt{r}}{K}\sum_{k=1}^K\sum_{h=1}^H \rho^{(t+h,k)}\alpha_{t+h}\,\|\psi^{(t+h,k)}\|_F,
\]
\qed
}
\begin{corollary}[Muon pseudogradient]
In the setting of Proposition~\ref{thm:pseudogradient-spectrum-lowerbound}, assume that the inner-step optimizer updates, $\psi^{(t+h,k)}$, are produced by Newton Shulz Iteration, making them orthonormal;
then
\[
\|\Psi\|_\ast
=
\frac{r}{K}\sum_{k=1}^K\sum_{h=1}^H \rho^{(t+h,k)}\alpha_{t+h}.
\]
\end{corollary}
\proof{This follows from proposition~\ref{thm:pseudogradient-spectrum-lowerbound} since $\psi^{(t+h,k)}$ is orthonormal we have $\|\psi^{(t+h,k)}\|_F=\sqrt{r}. \qed$}

\begin{corollary}[AdamW pseudogradient]
In the setting of Proposition~\ref{thm:pseudogradient-spectrum-lowerbound}, for general inner-step matrices $\psi^{(t+h,k)}$ (e.g., AdamW effective steps), one always has
\[
\|\Psi\|_\ast
=
\frac{\sqrt{r}}{K}\sum_{k=1}^K\sum_{h=1}^H \rho^{(t+h,k)}\alpha_{t+h}\,\|\psi^{(t+h,k)}\|_F.
\]
\end{corollary}
\proof{This is a restatement of proposition~\ref{thm:pseudogradient-spectrum-lowerbound} to illustrate that it applies to AdamW pseudogradients. $\qed$}

\section{Extended Results}
This section reviews complementary and additional results to the main manuscript: (\ref{sec:apdx:comp}) additional experiments regarding compressed communication, (\ref{sec:apdx:scalinglaws}) additional scaling law results, and (\ref{sec:apdx:15b}) additional results detailing our 15B training runs.

\subsection{MuLoCo vs DiLoCo: Pseudogradient Compression}\label{sec:apdx:comp}
\begin{figure*}[h]
    \centering
    \subfloat[Linear quantization]{\includegraphics[width=0.45\linewidth]{fig/quant_plots_linear.pdf}}
    \qquad
    \subfloat[Statistical quantization]{\includegraphics[width=0.45\linewidth]{fig/quant_plots_statistical.pdf}}\\[-2pt]

    \subfloat[Row-wise linear quantization]{\includegraphics[width=0.45\linewidth]{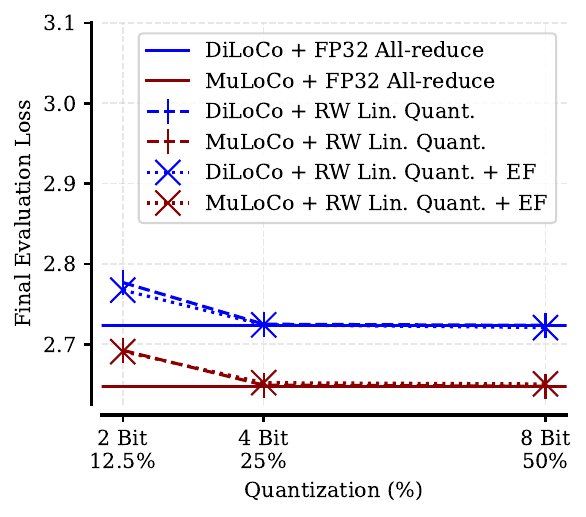}}
    \qquad
    \subfloat[Row-wise statistical quantization]{\includegraphics[width=0.45\linewidth]{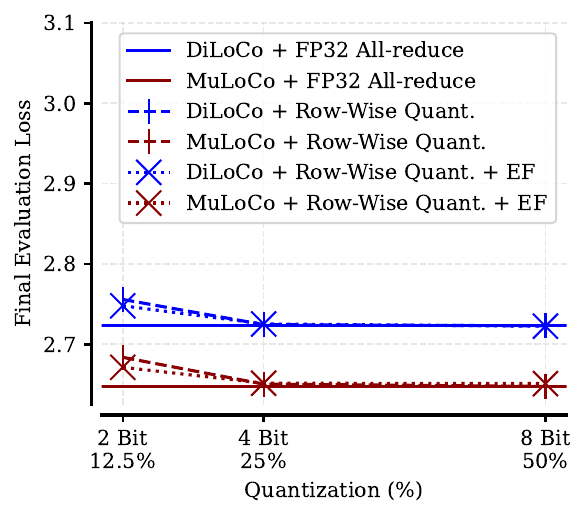}}

    \caption{\textbf{Quantization ablations.} We compare MuLoCo and DiLoCo under 8-bit (50\%), 4-bit (25\%), and 2-bit (12.5\%) quantized communication, with and without error feedback (EF). Across all settings, 4-bit quantization is effectively lossless and 2-bit quantization remains strong, \emph{except} for global linear quantization, which shows a clear degradation at 2 bits. In all cases, MuLoCo outperforms DiLoCo, and enabling EF provides little additional benefit.}
    \label{fig:appendix_quantization}
\end{figure*}

\begin{figure}[htbp]
        \vspace{0pt}
        \centering
        \includegraphics[width=0.5\linewidth]{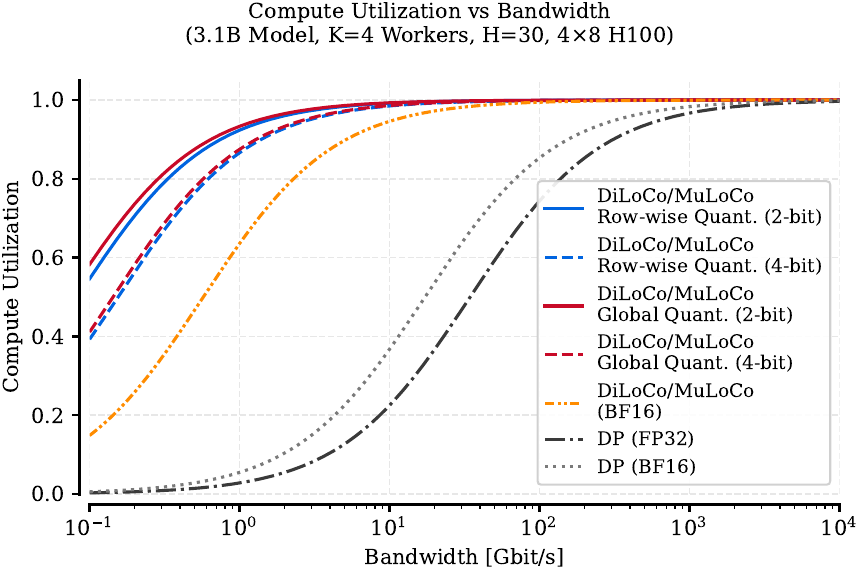}
        \captionof{figure}{\textbf{Compute Utilization as a function of network bandwidth when training a 3.1B model with different optimizers and compression schemes.} The plot was using timings from Table~\ref{fig:wallclock} and assumes that any time not spent communicating is used for computation. We observe that DiLoCo and MuLoCo baselines require two orders of magnitude less bandwidth to achieve $99\%+$ compute utilization than DP approaches. 
        }
        \label{fig:computeutil}
\end{figure}

Figure~\ref{fig:appendix_quantization} reports the performance for all quantization methods, including row-wise versions of statistical and linear quantization. We observe that across all settings, 4-bit quantization is effectively lossless and 2-bit quantization remains strong, \emph{except} for global linear quantization, which shows a clear degradation at 2 bits. In all cases, MuLoCo outperforms DiLoCo, and enabling EF provides little additional benefit. In figure~\ref{fig:computeutil} we observe that even quantizing to 4-bits can have a substantial advantage when communication is the bottleneck. As demonstrated in our experiments, this comes at no cost to performance.

\textbf{Assumptions around the collective used.} For our quantization experiments, we assume that collective communication is implemented using a bandwidth-efficient all-to-all reduce-scatter and subsequent ring all-gather~\citep{zeropp}. This choice is made instead of (1) using a bandwidth-expensive all-gather of full quantized parameters and (2) using a ring for the reduce-scatter. Compared to (1), the all-to-all allows us to avoid bandwidth growth in the number of workers, while relative to (2) it avoids compounding quantization errors during the ring-reduction, which is also exacerbated as the number of workers is increased. When using an all-to-all collective for the reduce scatter, all quantized tensors can be reduced in high precision on the same device before being re-quantized for the final all-gather. This required two total quantizations: (1) a quantization before the all-to-all, and (2) a quantization after the high-precision reduction before the all-gather. All our quantization experiments use two quantizations. For our top-$k$ experiments, we assume an all-gather across all parameters and, therefore, only sparsify the tensor once immediately before communication.

\textbf{Compression results in table format} For the reader's convenience, we provide our quantization and top-k results in Tables~\ref{tab:quantization_losses} and ~\ref{tab:topk_losseas}, respectively.

\begin{table}[htbp]
\centering
\caption{Final evaluation loss for top-$k$ sparsification.}
\label{tab:topk_losseas}
\begin{tabular}{lcccc}
\toprule
 & \multicolumn{2}{c}{DiLoCo} & \multicolumn{2}{c}{MuLoCo} \\
\cmidrule(lr){2-3} \cmidrule(lr){4-5}
Top-K & w/o EF & w/ EF & w/o EF & w/ EF \\
\midrule
FP32 (baseline) & \multicolumn{2}{c}{2.7238} & \multicolumn{2}{c}{2.6478} \\
\midrule
0.5\% & 3.3660 & 3.1646 & 3.0221 & 2.9052 \\
1\% & 3.2231 & 3.0641 & 2.9662 & 2.8633 \\
2.5\% & 3.0644 & 2.9402 & 2.9211 & 2.7988 \\
5\% & 2.9630 & 2.8695 & 2.8417 & 2.7580 \\
10\% & 2.8712 & 2.8224 & 2.7777 & 2.7230 \\
25\% & 2.7883 & 2.7529 & 2.7084 & 2.6822 \\
50\% & 2.7437 & 2.7316 & 2.6668 & 2.6566 \\
\bottomrule
\end{tabular}
\end{table}

\begin{table}[htbp]
\centering
\caption{Final evaluation loss for different quantization strategies.}
\label{tab:quantization_losses}
\begin{tabular}{llcccc}
\toprule
 & & \multicolumn{2}{c}{DiLoCo} & \multicolumn{2}{c}{MuLoCo} \\
\cmidrule(lr){3-4} \cmidrule(lr){5-6}
Compressor & Bits & w/o EF & w/ EF & w/o EF & w/ EF \\
\midrule
FP32 (baseline) & - & \multicolumn{2}{c}{2.7238} & \multicolumn{2}{c}{2.6478} \\
\midrule
Linear & 2-bit & 2.9819 & 2.8695 & 2.7751 & 2.7543 \\
 & 4-bit & 2.7268 & 2.7241 & 2.6551 & 2.6544 \\
 & 8-bit & 2.7223 & 2.7244 & 2.6478 & 2.6512 \\
\midrule
Statistical & 2-bit & 2.7576 & 2.7503 & 2.6938 & 2.6742 \\
 & 4-bit & 2.7255 & 2.7229 & 2.6505 & 2.6531 \\
 & 8-bit & 2.7214 & 2.7210 & 2.6474 & 2.6517 \\
\midrule
RW Linear & 2-bit & 2.7769 & 2.7675 & 2.6925 & 2.6908 \\
 & 4-bit & 2.7250 & 2.7242 & 2.6492 & 2.6522 \\
 & 8-bit & 2.7236 & 2.7208 & 2.6478 & 2.6507 \\
\midrule
RW Statistical & 2-bit & 2.7558 & 2.7480 & 2.6838 & 2.6714 \\
 & 4-bit & 2.7245 & 2.7253 & 2.6504 & 2.6516 \\
 & 8-bit & 2.7229 & 2.7225 & 2.6476 & 2.6511 \\
\bottomrule
\end{tabular}
\end{table}
\clearpage

\subsection{Extended Scaling Law Experiments}\label{sec:apdx:scalinglaws}
The following section presents further analysis and reports on greater details of our scaling law experiments. Table~\ref{tab:lc_joint_params} reports the parameters of scaling laws fit in section~\ref{sec:results-scaling}. We also report the final loss values of all methods in Table~\ref{tab:loss_comparison_vertical} and the zero-shot evaluation benchmark scores for 3.1B models in Table~\ref{tab:lm_eval_complete}.

\begin{table}[h]
    \centering
    \small
    \caption{\textbf{Parameters for $L(C) = aC^\alpha + L_0$} with joint $L_0 = 1.711$, fitted on all five training scales (150M--3.1B). Residuals are $|\log L_\mathrm{actual} - \log L_\mathrm{pred}|$.}
    \label{tab:lc_joint_params}
    \begin{tabular}{llrrrr}
        \toprule
        Method & $K$ & $a$ & $\alpha$ & Train & 15B \\
        \midrule
        DP AdamW & 1 & 5677 & $-$0.195 & 0.003 & 0.025 \\
        DP Muon & 1 & 6584 & $-$0.199 & 0.002 & 0.032 \\
        \addlinespace
        DiLoCo & 1 & 6620 & $-$0.199 & 0.004 & 0.032 \\
         & 2 & 5647 & $-$0.195 & 0.003 & --- \\
         & 4 & 5640 & $-$0.194 & 0.002 & --- \\
         & 8 & 6015 & $-$0.195 & 0.002 & --- \\
         & 16 & 5372 & $-$0.191 & 0.001 & 0.040 \\
        \addlinespace
        MuLoCo & 1 & 6927 & $-$0.200 & 0.002 & 0.028 \\
         & 2 & 6852 & $-$0.200 & 0.003 & --- \\
         & 4 & 6851 & $-$0.200 & 0.002 & --- \\
         & 8 & 6467 & $-$0.198 & 0.004 & --- \\
         & 16 & 6906 & $-$0.199 & 0.002 & 0.026 \\
        \midrule
        \textbf{Mean} & & & & \textbf{0.002} & \textbf{0.030} \\
        \bottomrule
    \end{tabular}
\end{table}

\begin{table}[ht]
    \centering
    \small
    \caption{\textbf{MuLoCo vs DiLoCo scaling study.} Percentages show relative change from respective DP baselines. Bold = best (lowest) loss in each column.}
    \label{tab:loss_comparison_vertical}
    \begin{adjustbox}{max width=\linewidth}
    \begin{tabular}{cl|lllll|l}
    \toprule
     & Method & \textbf{150M} & \textbf{416M} & \textbf{914M} & \textbf{1.8B} & \textbf{3.1B} & \textbf{15B} (No Sweep) \\
    \midrule
    \multirow{2}{*}{DP} & Muon & 3.124 & 2.641 & 2.402 & 2.246 & 2.128 & \textbf{1.875} \\
     & AdamW & 3.158 & 2.682 & 2.440 & 2.266 & 2.145 & 1.901 \\[2mm]
    \multirow{2}{*}{$K{=}1$} & MuLoCo & \textbf{3.120} ($-$0.1\%) & \textbf{2.638} ($-$0.1\%) & \textbf{2.400} ($-$0.1\%) & \textbf{2.238} ($-$0.4\%) & \textbf{2.122} ($-$0.3\%) & 1.878 (+0.2\%) \\
     & DiLoCo & 3.142 ($-$0.5\%) & 2.650 ($-$1.2\%) & 2.411 ($-$1.2\%) & 2.265 ($-$0.1\%) & 2.136 ($-$0.4\%) & 1.877 ($-$1.3\%) \\[2mm]
    \multirow{2}{*}{$K{=}2$} & MuLoCo & 3.130 (+0.2\%) & 2.642 (+0.0\%) & 2.402 (+0.0\%) & 2.245 (+0.0\%) & 2.127 ($-$0.1\%) & -- \\
     & DiLoCo & 3.159 (+0.0\%) & 2.668 ($-$0.5\%) & 2.428 ($-$0.5\%) & 2.275 (+0.4\%) & 2.155 (+0.5\%) & -- \\[2mm]
    \multirow{2}{*}{$K{=}4$} & MuLoCo & 3.148 (+0.8\%) & 2.651 (+0.4\%) & 2.410 (+0.4\%) & 2.254 (+0.4\%) & 2.139 (+0.5\%) & -- \\
     & DiLoCo & 3.192 (+1.1\%) & 2.697 (+0.6\%) & 2.447 (+0.3\%) & 2.283 (+0.7\%) & 2.166 (+1.0\%) & -- \\[2mm]
    \multirow{2}{*}{$K{=}8$} & MuLoCo & 3.177 (+1.7\%) & 2.659 (+0.7\%) & 2.426 (+1.0\%) & 2.273 (+1.2\%) & 2.155 (+1.3\%) & -- \\
     & DiLoCo & 3.248 (+2.8\%) & 2.734 (+1.9\%) & 2.478 (+1.6\%) & 2.300 (+1.5\%) & 2.181 (+1.7\%) & -- \\[2mm]
    \multirow{2}{*}{$K{=}16$} & MuLoCo & 3.222 (+3.1\%) & 2.713 (+2.8\%) & 2.448 (+1.9\%) & 2.291 (+2.0\%) & 2.165 (+1.7\%) & 1.902 (+1.4\%) \\
     & DiLoCo & 3.326 (+5.3\%) & 2.808 (+4.7\%) & 2.522 (+3.4\%) & 2.348 (+3.6\%) & 2.215 (+3.3\%) & 1.910 (+0.4\%) \\
    \bottomrule
    \end{tabular}
    \end{adjustbox}
\end{table}

\begin{table}[ht]
\centering
\caption{\textbf{Zero-shot LM Evaluation Results all optimizers at 3.1B scale and DP, K=1, and K=16 at 15B scale.} We report accuracy for MMLU, WinoGrande and normalized accuracy for HellaSwag, PIQA, ARC-E, ARC-C, OBQA. We additionally report mean evaluation accuracy across all available tasks.}
\label{tab:lm_eval_complete}
\begin{adjustbox}{width=\textwidth}
\begin{tabular}{llcccccccc}
\toprule
\textbf{Model Size} & \textbf{Configuration} & \textbf{MMLU} & \textbf{HellaSwag} & \textbf{WinoGrande} & \textbf{PIQA} & \textbf{ARC-E} & \textbf{ARC-C} & \textbf{OBQA} & \textbf{Mean Acc.} \\
\midrule
3.1B & DP AdamW  & 29.76\% & 66.98\% & 60.62\% & 76.33\% & 71.17\% & 42.66\% & 41.40\% & 55.56\% \\
3.1B & DP Muon  & 37.24\% & 68.18\% & 62.95\% & 77.49\% & 71.19\% & 42.03\% & 42.47\% & 57.36\%\\
3.1B & DiLoCo K=1  & 35.34\% & 67.58\% & 64.09\% & 77.58\% & 69.36\% & 41.81\% & 41.80\% & 56.79\% \\
3.1B & DiLoCo K=2  & 27.46\% & 66.48\% & 61.40\% & 76.66\% & 69.23\% & 41.30\% & 40.80\% & 54.76\% \\
3.1B & DiLoCo K=4  & 30.01\% & 65.84\% & 61.64\% & 76.82\% & 70.66\% & 41.30\% & 41.80\% & 55.44\% \\
3.1B & DiLoCo K=8  & 28.18\% & 65.45\% & 60.38\% & 76.33\% & 70.08\% & 42.92\% & 39.40\% & 54.68\% \\
3.1B & DiLoCo K=16 & 27.38\% & 63.03\% & 59.27\% & 75.95\% & 66.41\% & 41.72\% & 39.60\% & 53.34\% \\
3.1B & MuLoCo K=1  & 37.10\% & 68.60\% & 63.54\% & 77.69\% & 71.42\% & 42.58\% & 42.60\% & 57.65\% \\
3.1B & MuLoCo K=2  & 37.52\% & 68.35\% & 64.17\% & 77.69\% & 72.18\% & 42.24\% & 41.80\% & 57.71\% \\
3.1B & MuLoCo K=4  & 35.43\% & 67.49\% & 61.01\% & 77.75\% & 70.20\% & 41.72\% & 43.40\% & 56.71\% \\
3.1B & MuLoCo K=8  & 35.29\% & 66.82\% & 60.85\% & 77.31\% & 70.45\% & 43.00\% & 41.40\% & 56.45\% \\
3.1B & MuLoCo K=16  & 29.50\% & 65.46\% & 60.38\% & 75.90\% & 68.01\% & 39.68\% & 42.40\% & 54.48\% \\
\midrule
15B & DP AdamW  & 57.20\% & 77.69\% & 71.19\% & 81.39\% & 78.11\% & 52.30\% & 45.40\% & 66.18\% \\
15B & DP Muon  & 59.14\% & 79.05\% & 71.19\% & 81.12\% & 77.23\% & 53.07\% & 46.80\% & 66.80\% \\
15B & DiLoCo K=1  & 59.80\% & 79.06\% & 71.43\% & 81.34\% & 77.99\% & 52.56\% & 46.40\% & 66.94\% \\
15B & DiLoCo K=16  & 54.15\% & 77.59\% & 69.30\% & 81.72\% & 77.65\% & 52.56\% & 46.80\% & 65.68\% \\
15B & MuLoCo K=1 & 59.49\% & 79.06\% & 71.11\% & 81.61\% & 78.11\% & 52.73\% & 46.60\% & 66.96\% \\
15B & MuLoCo K=16  & 56.96\% & 77.18\% & 70.72\% & 80.74\% & 76.60\% & 53.16\% & 46.20\% & 65.94\% \\
\bottomrule
\end{tabular}
\end{adjustbox}
\end{table}

\begin{figure}[t]
    \centering
    \includegraphics[width=\linewidth]{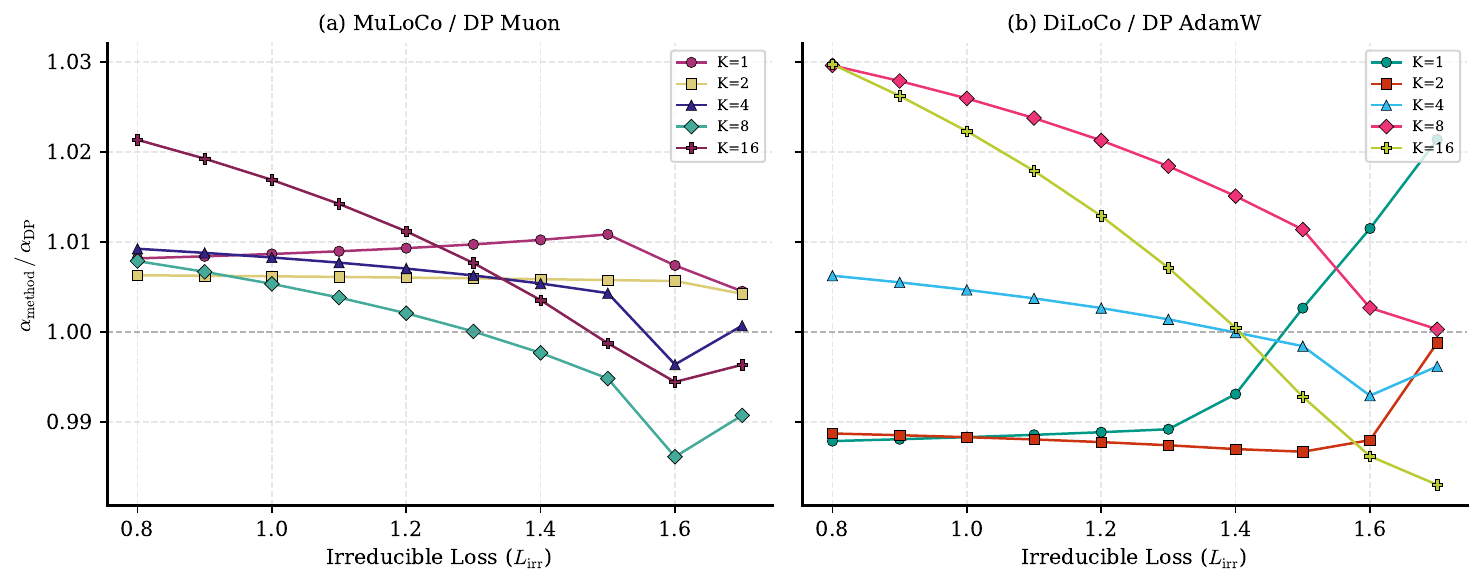}
    \caption{\textbf{Scaling exponents as a function of the assumed irreducible loss.}
    We fit compute scaling laws of the form $L(C)=aC^{\alpha}+L_{\mathrm{irr}}$ to our 150M-3.1B datapoints while holding $L_{\mathrm{irr}}$ fixed to a predetermined value that is shared across all methods. For each $L_{\mathrm{irr}}$, we report the resulting scaling exponent $\alpha$ \emph{normalized by the corresponding data-parallel (DP) baseline} (y-axis: $\alpha_{\text{method}}/\alpha_{\text{DP}}$), so values above $1$ indicate better than DP scaling while values below $1$ indicate worse scaling than DP.
    The left panel shows MuLoCo normalized by DP Muon, while the right panel shows DiLoCo normalized by DP AdamW.
    At lower irreducible losses, the fits suggest that higher-worker MuLoCo configurations (larger $K$) become increasingly competitive and will catch up to and surpass DP performance at scale.}
    \label{fig:exponents_vs_irr_normalized}
\end{figure}

\clearpage
\begin{figure*}[t]
    \centering
    \subfloat[Compute efficiency at optimal batch size]{%
        \includegraphics[width=0.48\linewidth]{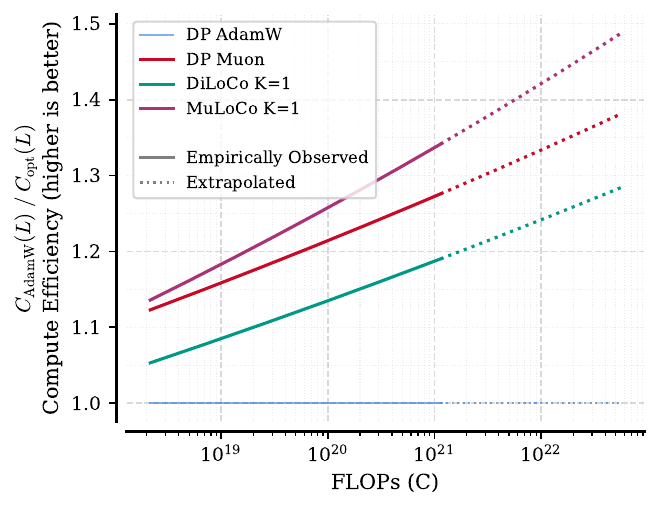}
    }
    \hfill
    \subfloat[Normalized performance at iso training time]{%
        \includegraphics[width=0.48\linewidth]{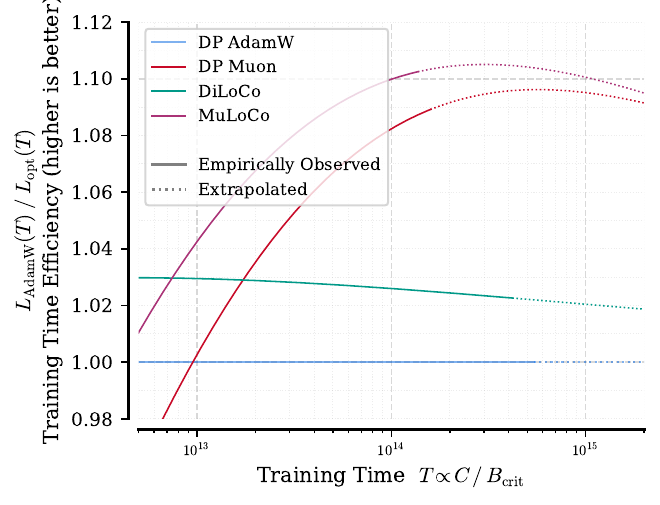}
    }
    \caption{\textbf{MuLoCo improves both compute efficiency and training-time-normalized performance.}
    (a) We report iso-loss compute efficiency, $\Cadamw(L)/\Copt(L)$, using scaling laws fit at each method's optimal batch size. Values above one indicate that optimizer $m$ reaches the same loss as AdamW using fewer FLOPs.
    (b) We report normalized performance at fixed training time, where training time is measured as FLOPs divided by the critical batch size. This metric is evaluated at $\Bcrit$ and assumes enough accelerators are scaled in proportion to the critical batch size to realize the corresponding reduction in sequential computation. Under this assumption, MuLoCo achieves an ${\sim}8$--$10\%$ loss improvement over AdamW for the same training-time budget.
    Solid curves indicate the empirically observed regime, while dotted curves indicate extrapolation beyond the range of available data.}
    \label{fig:apdx_compute_and_time_efficiency}
\end{figure*}

\textbf{Compute efficiency and training-time-normalized performance.}
Figure~\ref{fig:apdx_compute_and_time_efficiency} provides two complementary views of the efficiency gains implied by our scaling laws. The compute-efficiency plot uses the same compute--loss scaling-law methodology as Sec.~\ref{sec:scalingcompute}, fit across the extensively swept 150M--3.1B models, but evaluates each optimizer at its optimal batch size. For a target loss $L$, we invert the fitted loss curve for each optimizer to obtain the compute required to reach that loss, $C_m(L)$, and report $\Cadamw(L)/\Copt(L)$. Values above one therefore, indicate that optimizer $m$ reaches the same loss using fewer FLOPs than AdamW. In contrast, the training-time-normalized performance plot evaluates each method at its critical batch size, using $T \propto C/\Bcrit(C)$ as a proxy for the amount of sequential computation required. This asks a different question: given the same training-time budget, how much better is the final loss if enough accelerators are used in parallel to scale with the method's critical batch size? We find that MuLoCo improves over AdamW under both views: it is more compute-efficient at the optimal batch size ($\sim 1.35 \times$), and when training at $\Bcrit$, it can achieve an ${\sim}8$--$10\%$ loss improvement over AdamW for the same training-time budget.


\clearpage
\subsection{Extended 15B results}\label{sec:apdx:15b}
\begin{figure*}[t]
    \centering
    \begin{minipage}{\textwidth}
        \centering
        \includegraphics[width=\textwidth]{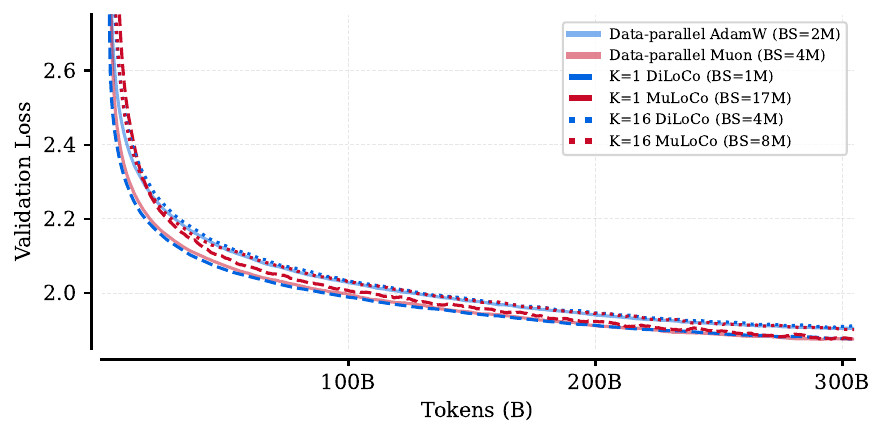}
    \end{minipage}
    
    \vspace{0.5em}
    
    \begin{minipage}{\textwidth}
        \centering
        \includegraphics[width=\textwidth]{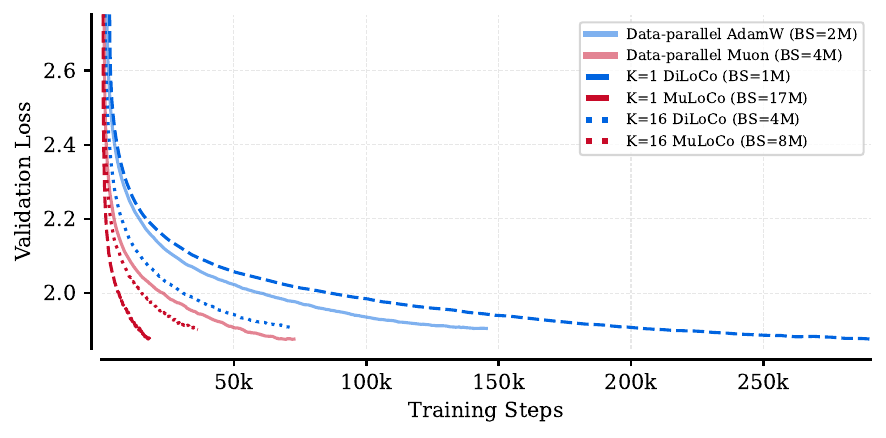}
    \end{minipage}
    
    \caption{
    \textbf{Validation loss for 15B pre-training runs.} We extrapolate optimal hyperparameters to 15B scale and train models with different optimizers.
    \emph{Top:} validation loss as a function of training tokens.
    \emph{Bottom:} validation loss as a function of optimization steps.
    Despite using a \textbf{16$\times$ larger batch size}, MuLoCo ($K=1$, BS=17M) reaches nearly the same final loss as DiLoCo ($K=1$, BS=1M), demonstrating its strong capability for large-batch training. 
    For $K=16$, DiLoCo narrows the gap to data-parallel AdamW, but both remain notably worse than data-parallel Muon, MuLoCo K=1, and DiLoCo K=1. 
    Despite reaching nearly the same loss as DP AdamW and DiLoCo K=16, MuLoCo $K=16$ can leverage larger batch sizes.
    }
    \label{fig:15b_validation_combined}
\end{figure*}

\begin{figure}[t]
    \centering
    
    \begin{subfigure}{0.8\linewidth}
        \centering
        \includegraphics[width=\linewidth]{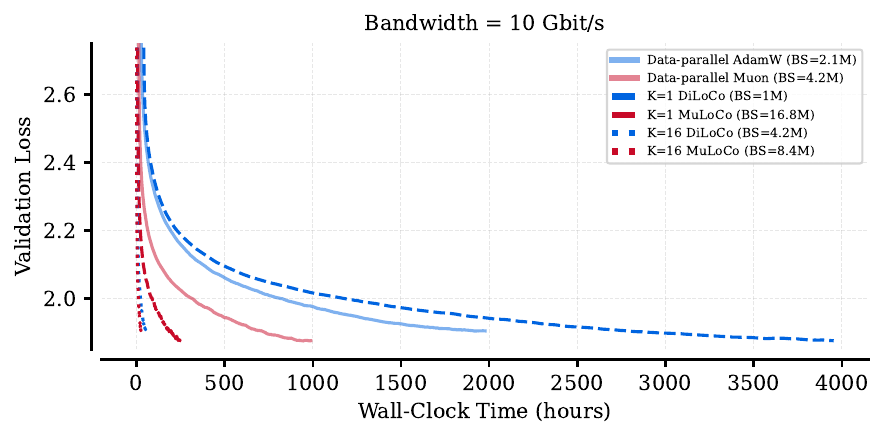}
    \end{subfigure}
    
    
    \begin{subfigure}{0.8\linewidth}
        \centering
        \includegraphics[width=\linewidth]{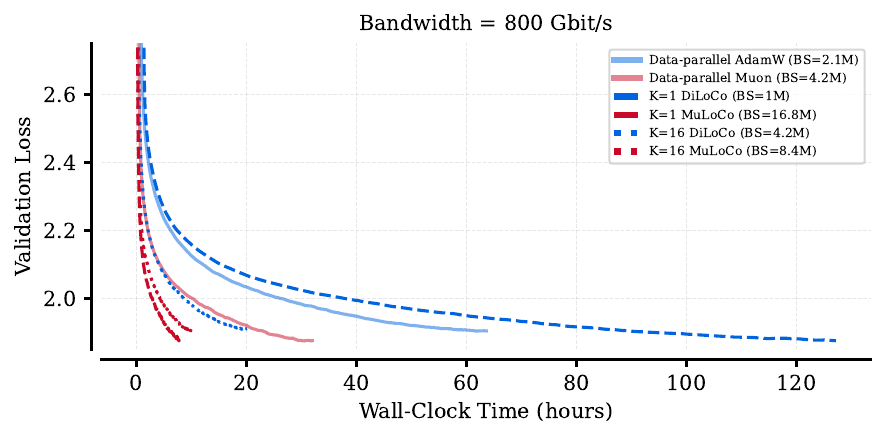}
    \end{subfigure}
    
    
    \begin{subfigure}{0.8\linewidth}
        \centering
        \includegraphics[width=\linewidth]{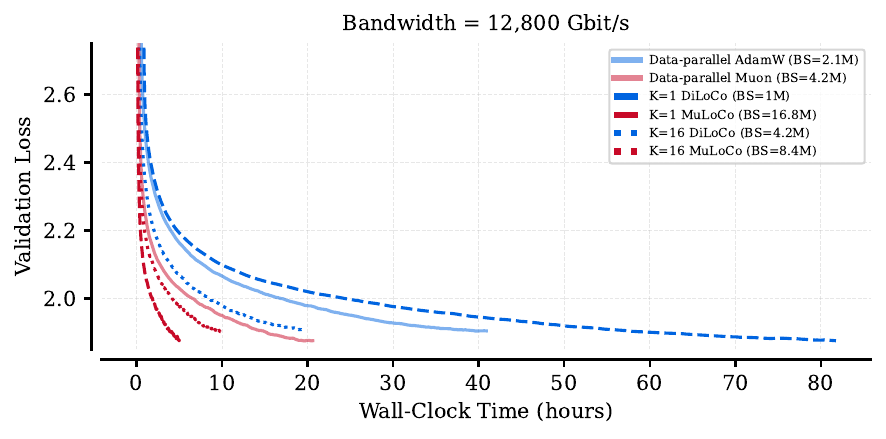}
    \end{subfigure}

    \caption{\textbf{Training time under bandwidth constraints.}
    We compare validation loss versus wall-clock time across network bandwidths of 10, 800, and 12{,}800 Gbit/s under the assumption that the cluster size is scaled proportional to the batch size.
    Both batch size and communication time critically influence overall wall-clock training time. 
    In the bandwidth-constrained regime (10 Gbit/s), batch size still matters, but reducing communication frequency dominates: the $K{=}16$ optimizers train substantially faster than $K{=}1$ MuLoCo despite using smaller batch sizes. 
    As network speed increases, communication becomes less of a bottleneck and larger batch sizes become more advantageous. 
    At 12{,}800 Gbit/s, we observe that $K{=}1$ MuLoCo (with its larger batch size) trains faster, illustrating the shifting trade-off between communication cost and statistical efficiency.}
    \label{fig:bandwidth_wall_clock}
\end{figure}

The following section reports on a number of additional results from our 15B experiments. First we present timings for the 15B training runs with Muon and AdamW (Table~\ref{tab:system_metrics_15b}). We find that the additional overhead of Muon's optimizer step is negligible at 15B scale. Using these metrics, we provide idealized wall-clock training estimates for different bandwidth networks, assuming the GPUs are scaled proportionally to batch size (Table~\ref{tab:idealized_wallclock_15b} and Figures~\ref{fig:bandwidth_wall_clock}). Finally, we also present validation loss curves for these training runs as a function of training tokens and training steps (Figure~\ref{fig:15b_validation_combined}).

\begin{table}[th]
\centering
\vspace{0pt}
\begin{adjustbox}{width=0.5\linewidth}
\begin{tabular}{@{}lccc@{}}
\toprule
\textbf{Metric} & \textbf{AdamW} & \textbf{Muon} & \textbf{Relative $\Delta$ (\%)} \\
\midrule
End-to-End (s) & 0.9753 & 0.9832 & +0.81 \\
MFU (\%) & 39.89 & 39.57 & $-$0.80 \\
TFLOPS/GPU & 394.54 & 391.37 & $-$0.80 \\
Throughput (tokens/s/GPU) & 4,200 & 4,166 & $-$0.81 \\\midrule
{Memory Complexity} & \multirow{2}{*}{4} & \multirow{2}{*}{3} & \multirow{2}{*}{$-$25} \\
{(Parameter copies)} &  &  &  \\
\bottomrule
\end{tabular}
\end{adjustbox}
\captionof{table}{\textbf{System-level performance comparison of training with AdamW and Muon at 15B scale.} 
End-to-end step time when training at local batch size 2 and global batch size $1024$ with 16 FSDP ranks and 32 data parallel ranks. Metrics correspond to the median over the first 500 steps. Relative $\Delta$ computed as $\frac{\text{Muon} - \text{AdamW}}{\text{AdamW}} \times 100$. We observe that the additional overhead of Muon's optimizer step is negligible at 15B scale.}
\label{tab:system_metrics_15b}
\end{table}

\begin{table}[h]
    \centering
    \small
    \caption{\textbf{Idealized wall-clock training time under bandwidth constraints.} We report idealized wall-clock times (in hours) estimated using the 15B system measurements in \ref{tab:system_metrics_15b}. The estimates factor in global batch size, communication overhead, optimizer step time, and forward/backward (FW/BW) compute time. We observe that communication-efficient optimizers dominate at low bandwidth, while larger batch sizes dominate under high bandwidth.}
    \label{tab:idealized_wallclock_15b}
    \begin{adjustbox}{max width=\linewidth}
    \begin{tabular}{@{}lrrrrrr@{}}
        \toprule
        \textbf{Method} & \multicolumn{6}{c}{\textbf{Training Time (hours)}} \\
        \cmidrule(l){2-7}
         & \textbf{10 Gbit/s} & \textbf{100 Gbit/s} & \textbf{400 Gbit/s} & \textbf{1,600 Gbit/s} & \textbf{3,200 Gbit/s} & \textbf{6,400 Gbit/s} \\
        \midrule
        Data-parallel AdamW (BS=2.1M) & 1008.2 & 136.3 & 63.6 & 45.4 & 42.4 & 40.9 \\
        Data-parallel Muon (BS=4.2M) & 504.3 & 68.3 & 32.0 & 22.9 & 21.4 & 20.6 \\
        K=1 DiLoCo (BS=1M) & 2016.3 & 272.5 & 127.2 & 90.8 & 84.8 & 81.8 \\
        K=1 MuLoCo (BS=16.8M) & 126.1 & 17.1 & 8.0 & 5.7 & 5.3 & 5.2 \\
        K=16 DiLoCo (BS=4.2M) & 35.8 & 21.3 & 20.1 & 19.8 & 19.7 & 19.7 \\
        K=16 MuLoCo (BS=8.4M) & 18.0 & 10.7 & 10.1 & 10.0 & 9.9 & 9.9 \\
        \bottomrule
    \end{tabular}
    \end{adjustbox}
\end{table}

\clearpage

\section{Extended pseudogradient analysis}

\begin{figure*}[h]
    \centering
    \subfloat[DiLoCo: Individual workers]{%
        \includegraphics[width=0.49\linewidth]{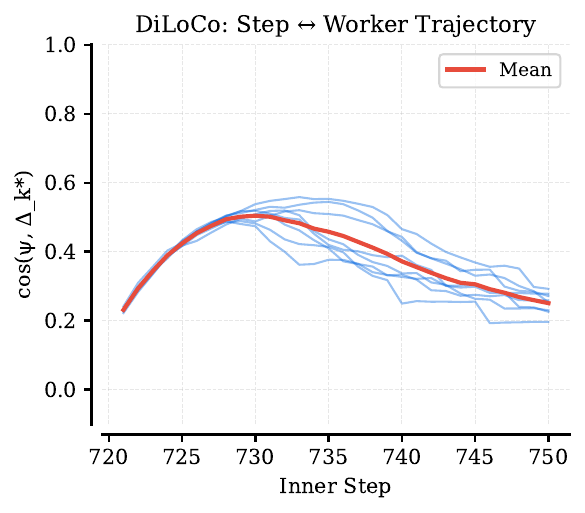}}
    \hfill
    \subfloat[MuLoCo: Individual workers]{%
        \includegraphics[width=0.49\linewidth]{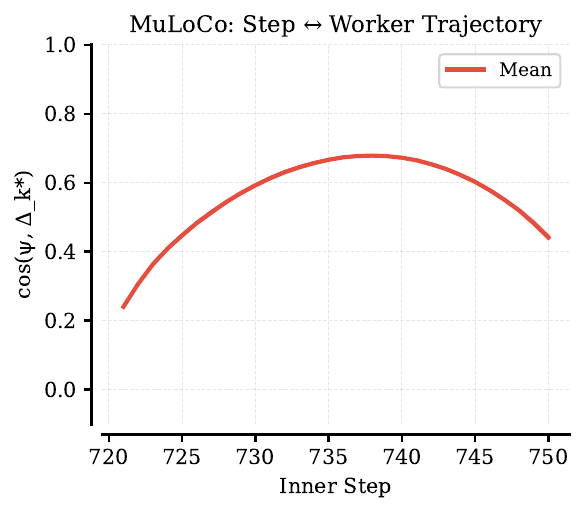}} \\[-2pt]
    \subfloat[DiLoCo: Pseudogradient]{%
        \includegraphics[width=0.49\linewidth]{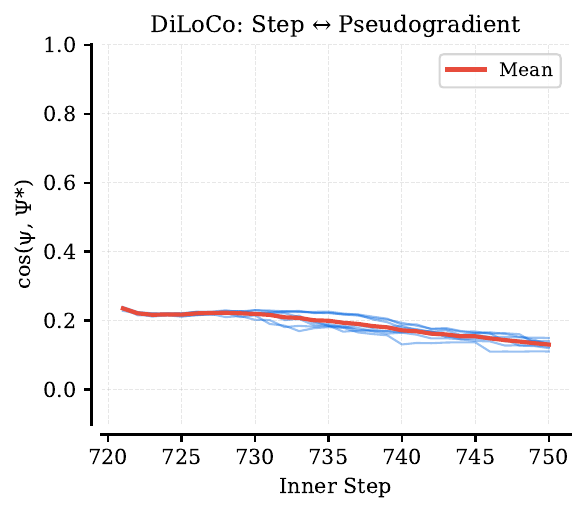}}
    \hfill
    \subfloat[MuLoCo: Pseudogradient]{%
        \includegraphics[width=0.49\linewidth]{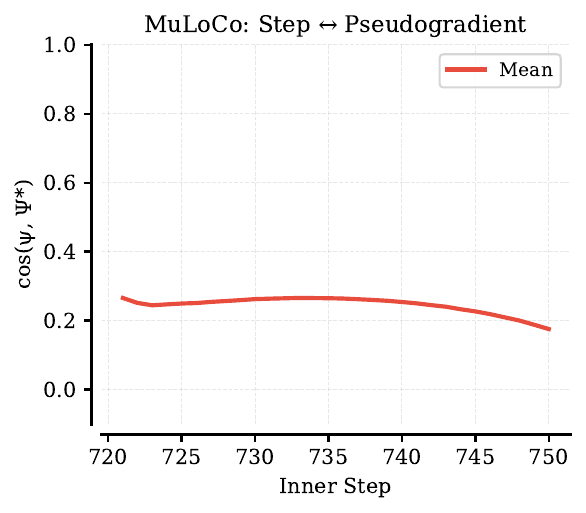}}
    \caption{\textbf{Worker-level variability in step alignment for DiLoCo vs.\ MuLoCo.} Each plot overlays the trajectories of \emph{individual workers} together with their \emph{mean} (labeled “Mean”). For MuLoCo (right column), the mean curve lies almost exactly on top of the individual-worker curves, making them largely indistinguishable. In contrast, DiLoCo (left column) exhibits visibly separated worker trajectories, indicating substantially higher variance across workers. Top row shows individual-worker trajectories; bottom row shows the corresponding pseudogradient views.}
    \label{fig:step_variance_appendix}
\end{figure*}

\section{Optimal hyperparameters }

\subsection{Hyperparameter Sweep Details}\label{app:hp_sweep_details}

As described in Section~\ref{sec:emp-eval}, we conduct extensive hyperparameter sweeps at each model scale for all optimizers and worker counts. In total, our scaling study comprises over 2,200 individual pre-training runs across six model scales (150M--15B), four optimizers (DP AdamW, DP Muon, DiLoCo, MuLoCo), and five worker counts ($K\in\{1,2,4,8,16\}$). We summarize the sweep ranges, tuning strategy, and optimal hyperparameters found below.

\paragraph{Sweep ranges.}
Table~\ref{tab:hp_sweep_ranges} summarizes the hyperparameter search spaces used across our scaling study. All learning rates are swept on a multiplicative grid of powers of $\sqrt{2}$ (i.e., each value is $\sqrt{2}\times$ the previous), while batch sizes are swept on powers of 2. Weight decay values are rescaled according to~\citet{scalewd} as the batch size changes: for DP methods we rescale by the global batch size, and for DiLoCo/MuLoCo we rescale by the per-worker batch size ($B/K$). The outer optimizer hyperparameters ($\eta_\text{out}$, $\mu$) are extensively swept at the 416M scale and reused at other scales, following~\citet{scalingdiloco}'s finding that they remain roughly constant across model sizes.

\begin{table}[t]
\centering
\small
\caption{\textbf{Hyperparameter search spaces.} Ranges are aggregated across all model scales and worker counts. They correspond to the ranges that we needed to sweep in order to ensure the optimal HP value was an interior point of the values tried. Learning rates follow a $\sqrt{2}$ grid; batch sizes follow a powers-of-2 grid. Weight decay is rescaled with batch size per~\citet{scalewd}. Outer optimizer HPs are swept at 416M and reused at larger scales.}
\label{tab:hp_sweep_ranges}
\begin{adjustbox}{max width=\textwidth}
\begin{tabular}{lcccc}
\toprule
\textbf{Hyperparameter} & \textbf{DP AdamW} & \textbf{DP Muon} & \textbf{DiLoCo} & \textbf{MuLoCo} \\
\midrule
Inner LR ($\eta_\text{in}$) & $[1.7\text{e-}4,\, 6.3\text{e-}2]$ & $[3.9\text{e-}3,\, 3.5\text{e-}1]$ & $[3.5\text{e-}4,\, 8.8\text{e-}2]$ & $[4.9\text{e-}4,\, 1.8\text{e-}1]$ \\
Weight decay ($\lambda$) & $[6.3\text{e-}5,\, 1\text{e-}1]$ & $[1\text{e-}4,\, 1\text{e-}1]$ & $[1.6\text{e-}5,\, 1.7\text{e-}1]$ & $[5\text{e-}5,\, 8\text{e-}2]$ \\
Global batch size ($B$) & $\{32$--$4096\}$ & $\{32$--$8192\}$ & $\{8$--$8192\}$ & $\{8$--$8192\}$ \\
Outer LR ($\eta_\text{out}$) & --- & --- & $\{0.4$--$1.1\}$ & $\{0.4$--$1.1\}$ \\
Outer momentum ($\mu$) & --- & --- & $\{0.5$--$0.9\}$ & $\{0.5$--$0.9\}$ \\
\midrule
Total runs (all scales) & 339 & 260 & 822 & 808 \\
\bottomrule
\end{tabular}
\end{adjustbox}
\end{table}

\begin{figure}[h]
    \centering
    \includegraphics[width=0.5\linewidth]{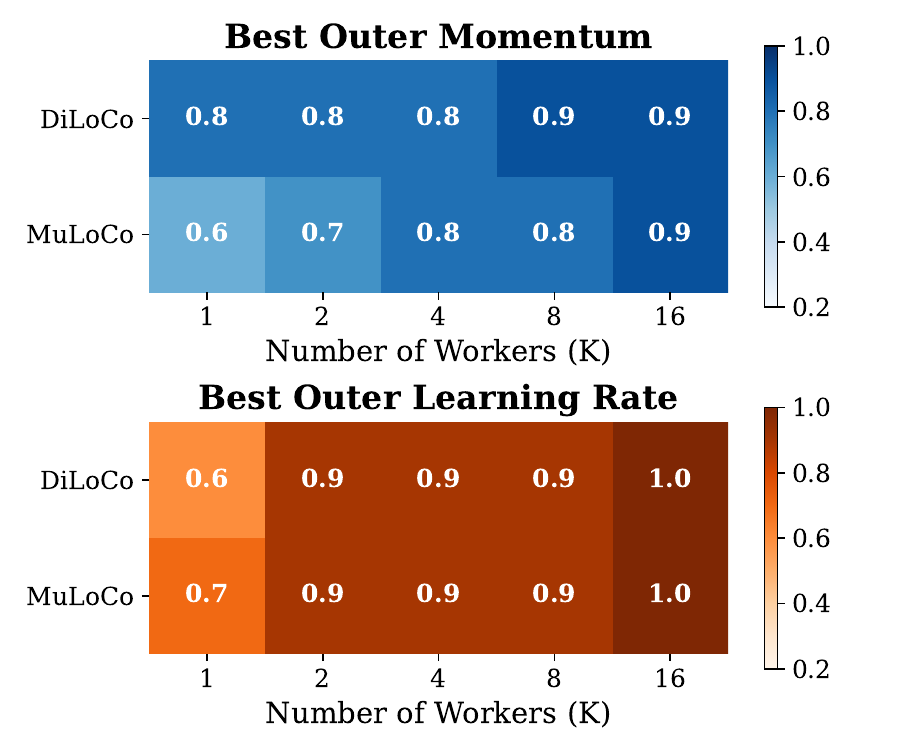}
    \caption{\textbf{At low worker counts MuLoCo's outer momentum should be decreased.} When tuning the outer optimizer hyperparameters for MuLoCo and DiLoCo, we found that MuLoCo favors lower momentum coefficients at lower worker counts relative to DiLoCo.}
    \label{fig:outer_opt_grid}
\end{figure}

\paragraph{Optimal hyperparameters at each scale.}
Tables~\ref{tab:hp_optimal_dp}--\ref{tab:hp_optimal_muloco} report the optimal hyperparameters found at each model scale. Several trends emerge:
\begin{itemize}
    \item \textbf{Learning rate decreases with scale} for AdamW-based methods (DP AdamW, DiLoCo), dropping roughly $30\times$ from 150M to 3.1B. In contrast, Muon-based methods (DP Muon, MuLoCo) maintain relatively stable learning rates across scales.
    \item \textbf{Weight decay is tied to batch size} via the rescaling rule of~\citet{scalewd}. Within a fixed scale, $\lambda$ decreases as $K$ increases because the per-worker batch size ($B/K$) decreases.
    \item \textbf{Batch size grows with scale} for all methods. DP methods and $K{=}1$ workers generally use $B=32$--$256$ at 150M--3.1B. At higher $K$, the per-worker batch size decreases while the total token batch size ($B \times K \times 2048$) remains roughly constant within a scale.
    \item \textbf{MuLoCo favors lower outer momentum at small K.} As shown in Figure~\ref{fig:outer_opt_grid}, MuLoCo uses $\mu=0.6$ at $K=1$ versus $\mu=0.8$ for DiLoCo. At $K\geq4$, both converge to $\mu=0.8$--$0.9$.
    \item \textbf{Outer learning rate increases with K} for both DiLoCo and MuLoCo, rising from $\eta_\text{out}=0.6$--$0.7$ at $K=1$ to $\eta_\text{out}=1.0$ at $K=16$.
\end{itemize}

\begin{table}[t]
\centering
\small
\caption{\textbf{Optimal hyperparameters for data-parallel baselines} at each model scale. $\eta_\text{in}$: inner learning rate; $\lambda$: weight decay; $B$: global batch size (sequences); $B_\text{tok}$: global token batch size ($= B \times 2048$). Since $K=1$ for DP methods, $B/K = B$.}
\label{tab:hp_optimal_dp}
\begin{tabular}{llrrrr}
\toprule
\textbf{Scale} & \textbf{Method} & $\boldsymbol{\eta_\text{in}}$ & $\boldsymbol{\lambda}$ & $\boldsymbol{B}$ & $\boldsymbol{B_\text{tok}}$ \\
\midrule
150M & DP AdamW & 4.42e-2 & 1.25e-4 & 64 & 128K \\
     & DP Muon  & 1.25e-1 & 1.25e-3 & 64 & 128K \\
\midrule
416M & DP AdamW & 1.10e-2 & 3.00e-2 & 512 & 1M \\
     & DP Muon  & 8.84e-2 & 2.50e-3 & 512 & 1M \\
\midrule
914M & DP AdamW & 5.52e-3 & 5.00e-3 & 256 & 512K \\
     & DP Muon  & 3.13e-2 & 5.00e-3 & 256 & 512K \\
\midrule
1.76B & DP AdamW & 1.38e-3 & 1.00e-1 & 512 & 1M \\
      & DP Muon  & 3.13e-2 & 1.00e-3 & 512 & 1M \\
\midrule
3.07B & DP AdamW & 1.38e-3 & 5.00e-4 & 256 & 512K \\
      & DP Muon  & 2.21e-2 & 1.00e-3 & 512 & 1M \\
\midrule
15.2B & DP AdamW & 1.73e-4 & 2.00e-4 & 1024 & 2M \\
      & DP Muon  & 1.10e-2 & 4.00e-4 & 2048 & 4M \\
\bottomrule
\end{tabular}
\end{table}

\begin{table}[t]
\centering
\small
\caption{\textbf{Optimal hyperparameters for DiLoCo.} At each model scale and worker count $K$. $B/K$: per-worker batch size (sequences); $B$: global batch size; $B_\text{tok}$: global token batch size ($= B \times 2048$). All runs use $H=30$ synchronization steps and SGD with Nesterov momentum as the outer optimizer.}
\label{tab:hp_optimal_diloco}
\begin{adjustbox}{max width=\textwidth}
\begin{tabular}{clrrrrrrrr}
\toprule
\textbf{Scale} & $\boldsymbol{K}$ & $\boldsymbol{\eta_\text{in}}$ & $\boldsymbol{\lambda}$ & $\boldsymbol{B/K}$ & $\boldsymbol{B}$ & $\boldsymbol{B_\text{tok}}$ & $\boldsymbol{\eta_\text{out}}$ & $\boldsymbol{\mu}$ \\
\midrule
\multirow{5}{*}{150M}
 & 1  & 2.21e-2 & 2.50e-4  & 128 & 128  & 256K & 0.6 & 0.8 \\
 & 2  & 4.42e-2 & 1.25e-4  & 64  & 128  & 256K & 0.9 & 0.8 \\
 & 4  & 2.21e-2 & 6.25e-5  & 32  & 128  & 256K & 0.9 & 0.8 \\
 & 8  & 4.42e-2 & 3.13e-5  & 16  & 128  & 256K & 0.9 & 0.9 \\
 & 16 & 2.21e-2 & 3.13e-5  & 16  & 256  & 512K & 1.0 & 0.9 \\
\midrule
\multirow{5}{*}{416M}
 & 1  & 1.10e-2 & 5.00e-3  & 256 & 256  & 512K & 0.6 & 0.8 \\
 & 2  & 1.10e-2 & 2.50e-3  & 128 & 256  & 512K & 0.9 & 0.8 \\
 & 4  & 1.10e-2 & 1.25e-3  & 64  & 256  & 512K & 0.9 & 0.8 \\
 & 8  & 7.81e-3 & 1.25e-3  & 64  & 512  & 1M   & 0.7 & 0.9 \\
 & 16 & 2.76e-3 & 6.25e-4  & 32  & 512  & 1M   & 1.0 & 0.9 \\
\midrule
\multirow{5}{*}{914M}
 & 1  & 5.52e-3 & 5.00e-3  & 256 & 256  & 512K & 0.6 & 0.8 \\
 & 2  & 3.91e-3 & 2.50e-3  & 128 & 256  & 512K & 0.9 & 0.8 \\
 & 4  & 3.91e-3 & 1.25e-3  & 64  & 256  & 512K & 0.9 & 0.8 \\
 & 8  & 2.76e-3 & 1.25e-3  & 64  & 512  & 1M   & 0.9 & 0.9 \\
 & 16 & 1.95e-3 & 6.25e-4  & 32  & 512  & 1M   & 1.0 & 0.9 \\
\midrule
\multirow{5}{*}{1.76B}
 & 1  & 1.95e-3 & 1.00e-3  & 512 & 512  & 1M   & 0.6 & 0.8 \\
 & 2  & 1.38e-3 & 5.00e-4  & 256 & 512  & 1M   & 0.9 & 0.8 \\
 & 4  & 1.95e-3 & 2.50e-4  & 128 & 512  & 1M   & 0.9 & 0.8 \\
 & 8  & 1.38e-3 & 1.25e-4  & 64  & 512  & 1M   & 0.9 & 0.9 \\
 & 16 & 1.38e-3 & 6.25e-5  & 32  & 512  & 1M   & 1.0 & 0.9 \\
\midrule
\multirow{5}{*}{3.07B}
 & 1  & 1.38e-3 & 5.00e-4  & 256 & 256  & 512K & 0.6 & 0.8 \\
 & 2  & 6.91e-4 & 5.00e-4  & 256 & 512  & 1M   & 0.9 & 0.8 \\
 & 4  & 1.38e-3 & 2.50e-4  & 128 & 512  & 1M   & 0.9 & 0.8 \\
 & 8  & 1.38e-3 & 1.25e-4  & 64  & 512  & 1M   & 0.9 & 0.9 \\
 & 16 & 6.91e-4 & 1.25e-4  & 64  & 1024 & 2M   & 1.0 & 0.9 \\
\bottomrule
\end{tabular}
\end{adjustbox}
\end{table}

\begin{table}[t]
\centering
\small
\caption{\textbf{Optimal hyperparameters for MuLoCo.} At each model scale and worker count $K$. $B/K$: per-worker batch size (sequences); $B$: global batch size; $B_\text{tok}$: global token batch size ($= B \times 2048$). All runs use $H=30$ synchronization steps and SGD with Nesterov momentum as the outer optimizer.}
\label{tab:hp_optimal_muloco}
\begin{adjustbox}{max width=\textwidth}
\begin{tabular}{clrrrrrrrr}
\toprule
\textbf{Scale} & $\boldsymbol{K}$ & $\boldsymbol{\eta_\text{in}}$ & $\boldsymbol{\lambda}$ & $\boldsymbol{B/K}$ & $\boldsymbol{B}$ & $\boldsymbol{B_\text{tok}}$ & $\boldsymbol{\eta_\text{out}}$ & $\boldsymbol{\mu}$ \\
\midrule
\multirow{5}{*}{150M}
 & 1  & 6.25e-2 & 1.25e-3  & 64  & 64   & 128K & 0.7 & 0.6 \\
 & 2  & 8.84e-2 & 1.25e-3  & 64  & 128  & 256K & 0.9 & 0.7 \\
 & 4  & 8.84e-2 & 6.25e-4  & 32  & 128  & 256K & 0.9 & 0.8 \\
 & 8  & 1.25e-1 & 6.25e-4  & 32  & 256  & 512K & 0.9 & 0.8 \\
 & 16 & 6.25e-2 & 3.13e-4  & 16  & 256  & 512K & 1.0 & 0.9 \\
\midrule
\multirow{5}{*}{416M}
 & 1  & 4.42e-2 & 5.00e-3  & 256 & 256  & 512K & 0.7 & 0.6 \\
 & 2  & 4.42e-2 & 5.00e-3  & 256 & 512  & 1M   & 0.9 & 0.7 \\
 & 4  & 4.42e-2 & 2.50e-3  & 128 & 512  & 1M   & 0.9 & 0.8 \\
 & 8  & 6.25e-2 & 1.25e-3  & 64  & 512  & 1M   & 0.9 & 0.8 \\
 & 16 & 4.42e-2 & 6.25e-4  & 32  & 512  & 1M   & 1.0 & 0.9 \\
\midrule
\multirow{5}{*}{914M}
 & 1  & 2.21e-2 & 5.00e-3  & 256 & 256  & 512K & 0.7 & 0.6 \\
 & 2  & 2.21e-2 & 2.50e-3  & 128 & 256  & 512K & 0.9 & 0.7 \\
 & 4  & 2.21e-2 & 1.25e-3  & 64  & 256  & 512K & 0.9 & 0.8 \\
 & 8  & 3.13e-2 & 1.25e-3  & 64  & 512  & 1M   & 0.9 & 0.8 \\
 & 16 & 3.13e-2 & 1.25e-3  & 64  & 1024 & 2M   & 1.0 & 0.9 \\
\midrule
\multirow{5}{*}{1.76B}
 & 1  & 3.13e-2 & 2.00e-3  & 1024 & 1024 & 2M   & 0.7 & 0.6 \\
 & 2  & 2.21e-2 & 5.00e-4  & 256  & 512  & 1M   & 0.9 & 0.7 \\
 & 4  & 2.21e-2 & 2.50e-4  & 128  & 512  & 1M   & 0.9 & 0.8 \\
 & 8  & 1.56e-2 & 1.25e-4  & 64   & 512  & 1M   & 0.9 & 0.8 \\
 & 16 & 1.56e-2 & 1.25e-4  & 64   & 1024 & 2M   & 1.0 & 0.9 \\
\midrule
\multirow{5}{*}{3.07B}
 & 1  & 1.56e-2 & 5.00e-4  & 256  & 256  & 512K & 0.7 & 0.6 \\
 & 2  & 1.56e-2 & 5.00e-4  & 256  & 512  & 1M   & 0.9 & 0.8 \\
 & 4  & 1.56e-2 & 2.50e-4  & 128  & 512  & 1M   & 0.9 & 0.8 \\
 & 8  & 1.56e-2 & 2.50e-4  & 128  & 1024 & 2M   & 0.9 & 0.8 \\
 & 16 & 1.56e-2 & 2.50e-4  & 128  & 2048 & 4M   & 1.0 & 0.9 \\
\bottomrule
\end{tabular}
\end{adjustbox}
\end{table}

\paragraph{Outer optimizer hyperparameters.}
For both DiLoCo and MuLoCo, we observe a consistent pattern across all model sizes: (1) the outer learning rate increases from $0.6$--$0.7$ at $K=1$ to $1.0$ at $K=16$, and (2) the outer momentum increases from $0.6$--$0.8$ at $K=1$ to $0.9$ at $K\geq8$. Notably, MuLoCo uses a lower outer momentum ($\mu=0.6$) than DiLoCo ($\mu=0.8$) at $K=1$, which is consistent with Muon's optimizer steps already being well-aligned (see Section~\ref{sec:inner-opt}), reducing the benefit of additional momentum. These values are extensively tuned at the 416M scale (Table~\ref{tab:hp_sweep_ranges}) and reused at all other scales. For a visualization of optimal values, see Figure~\ref{fig:outer_opt_grid}

\paragraph{Weight decay rescaling.}
At each model scale, we first determine the optimal weight decay $\lambda^*$ for DP AdamW and DP Muon at a fixed batch size of $B=$ 1M tokens (approximately $B=512$ sequences). When sweeping other batch sizes, we rescale $\lambda^*$ following~\citet{scalewd}. For DiLoCo and MuLoCo, the weight decay is rescaled according to the per-worker batch size ($B/K$), which explains the systematic decrease of $\lambda$ with increasing $K$ observed in Tables~\ref{tab:hp_optimal_diloco} and~\ref{tab:hp_optimal_muloco}.

\clearpage
\subsection{Extropolating HPs to 15B scale}\label{sec:apdx:15b-tuning}
\begin{figure*}[ht]
    \centering
    \subfloat[Batch Size (K=1 Worker)]{\includegraphics[width=0.45\linewidth]{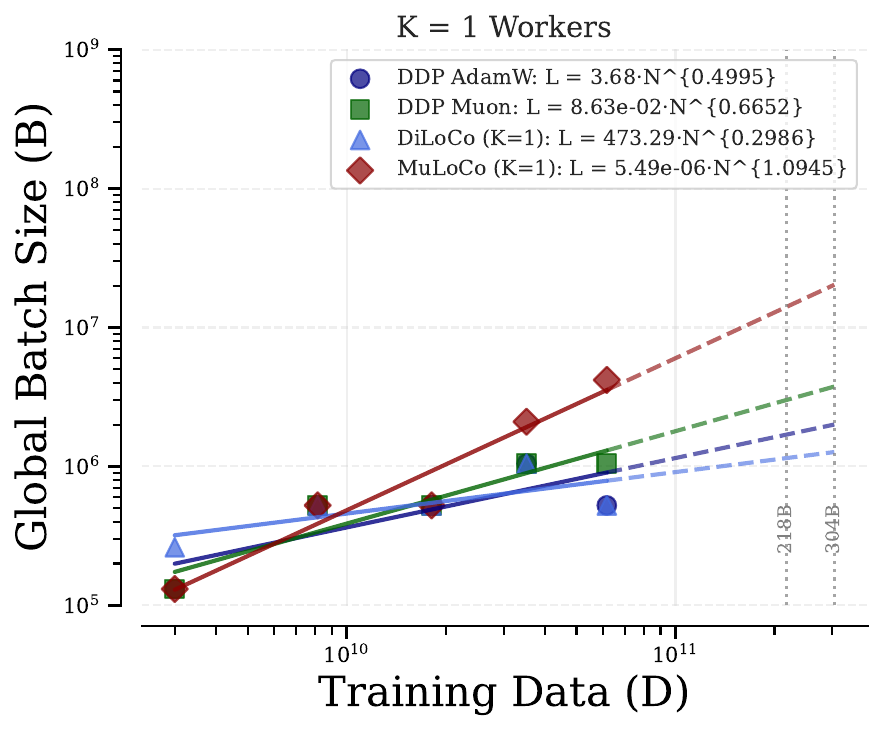}}\qquad
    \subfloat[Batch Size (K=16 Workers)]{\includegraphics[width=0.45\linewidth]{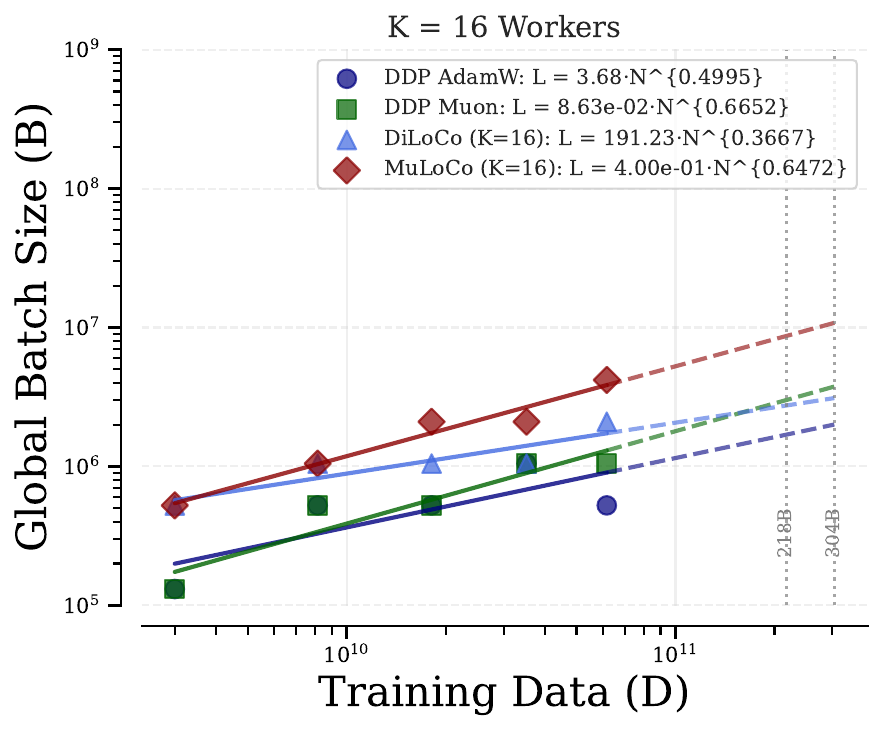}}\\
    \subfloat[Learning Rate (K=1 Worker)]{\includegraphics[width=0.45\linewidth]{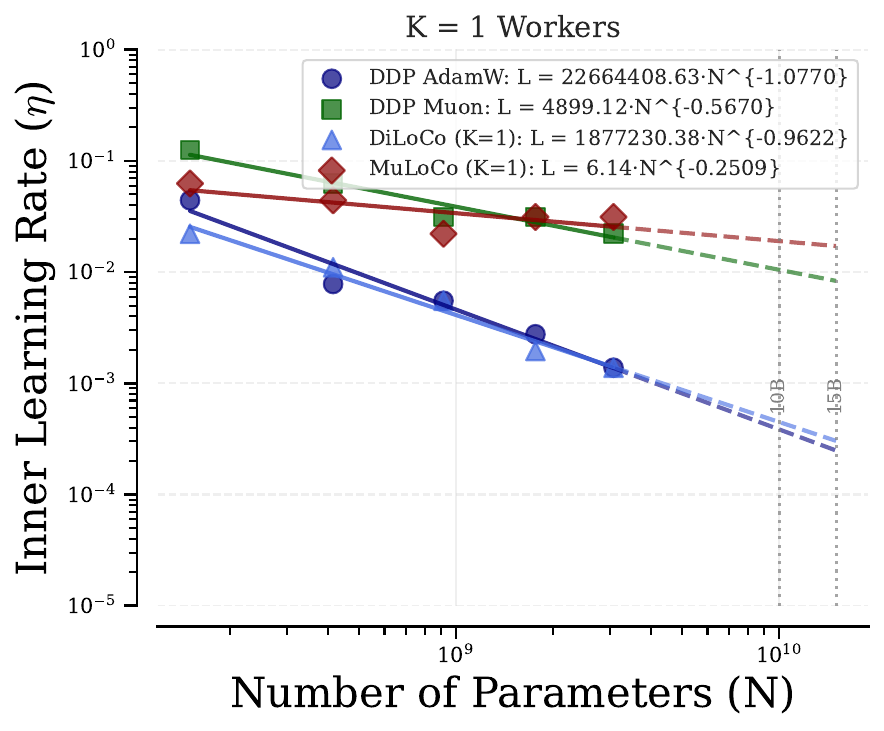}}\qquad
    \subfloat[Learning Rate (K=16 Workers)]{\includegraphics[width=0.45\linewidth]{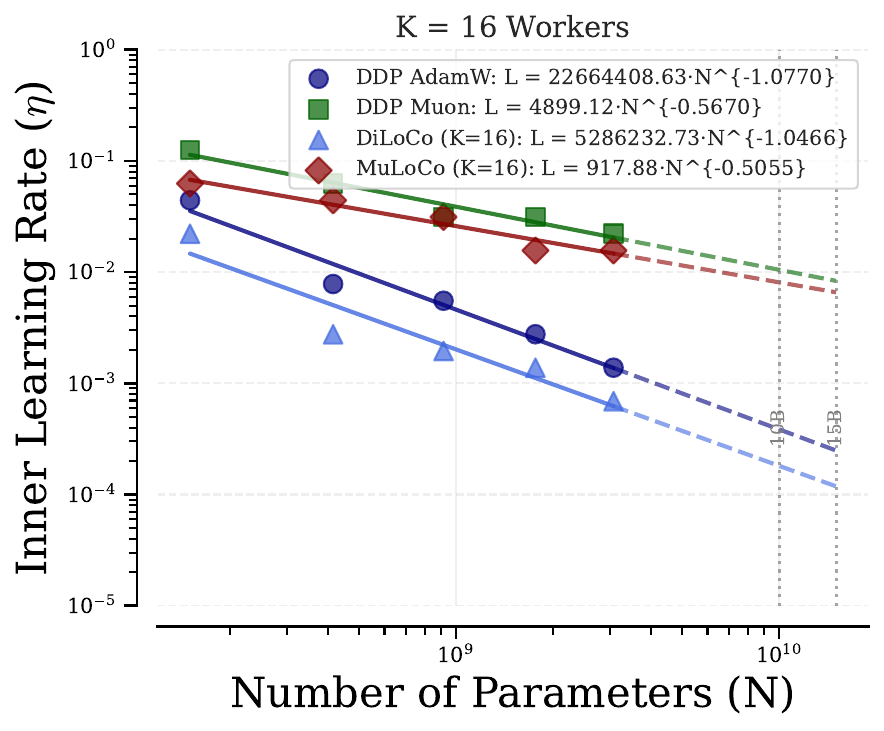}}
    \caption{\textbf{Extrapolating optimal batch size ($B^*$) and inner learning rate ($\eta_\mathrm{in}$) for different optimization algorithms.} We fit a power law in tokens (D) to the optimal batch size and in parameters (N) for the optimal learning rates at each scale.  \textbf{**NOTE 1:} These were fit before realizing that B=4M was a local minimum for K=1 MuLoCo at 3.1B scale, so the optimal batch size is slightly larger. We still chose to report these fits since they were used in our 15B experiments. \textbf{**NOTE 2:} The legend in these plots shows an powerlaw with misleading variables. It should be $B=AD^\alpha$ for (a,b) $\eta_\mathrm{in}=AN^\alpha$ for (c,d). The y and x axes are correct. }
    \label{fig:bsextrapolation} \vspace{-15pt}
\end{figure*}

At the 15B scale, running full hyperparameter sweeps is computationally prohibitive. Instead, we extrapolate optimal weight decay, inner learning rate, and batch size from power law fits to smaller scales. To promote better accelerator utilization, we use the nearest power-of-2 batch size to the optimal extrapolated value. Depending on whether we rounded the batch size up or down, we adjust the learning rate up or down, respectively to its nearest $\sqrt{2}$ power. For weight decay, we used $1e-4$ for both AdamW and Muon at $B=1$M since the extrapolated values were close enough, and this matches our other experiments. The hyperparameters used are reported in table~\ref{tab:hp_15B}. It should be noted that our power laws with $B^*$ were fit with K=1 MuLoCo, having $B^*=4$M at 3.1B scale. However, further sweeping later revealed that this was a local minimum (see Fig.~\ref{fig:batchsize_3p1b_joint}). This means that K=1 MuLoCo's batch size is likely closer to the CBS than $B^*$ at 15B scale and that its optimal final performance is likely slightly under-reported. The power law fits used for $B^*$ and $\eta$ are reported in Figure.~\ref{fig:bsextrapolation}.

\begin{table}[h]
\centering
\small
\caption{\textbf{Hyperparameters for 15B runs.} These are extrapolated from power law fits to smaller scales (150M--3.07B) with minor manual adjustments. $B/K$: per-worker batch size (sequences); $B$: global batch size (sequences); $B_\text{tok}$: global token batch size ($= B \times 2048$).}
\label{tab:hp_15B}
\begin{adjustbox}{max width=\textwidth}
\begin{tabular}{llrrrrrrr}
\toprule
\textbf{Method} & $\boldsymbol{K}$ & $\boldsymbol{\eta_\text{in}}$ & $\boldsymbol{\lambda}$ & $\boldsymbol{B/K}$ & $\boldsymbol{B}$ & $\boldsymbol{B_\text{tok}}$ & $\boldsymbol{\eta_\text{out}}$ & $\boldsymbol{\mu}$ \\
\midrule
DP AdamW & 1  & 1.73e-4 & 2.00e-4 & 1024 & 1024 & 2M   & ---  & ---  \\
DP Muon  & 1  & 1.10e-2 & 4.00e-4 & 2048 & 2048 & 4M   & ---  & ---  \\
\midrule
DiLoCo   & 1  & 3.45e-4 & 1.00e-4 & 512  & 512  & 1M   & 0.6  & 0.8  \\
DiLoCo   & 16 & 3.45e-4 & 2.50e-5 & 128  & 2048 & 4M   & 1.0  & 0.9  \\
\midrule
MuLoCo   & 1  & 1.56e-2 & 1.60e-3 & 8192 & 8192 & 16M  & 0.7  & 0.6  \\
MuLoCo   & 16 & 1.10e-2 & 5.00e-5 & 256  & 4096 & 8M   & 1.0  & 0.9  \\
\bottomrule
\end{tabular}
\end{adjustbox}
\end{table}

\clearpage

\section{A Robuts Evaluation Loss Estimate}\label{sec:evalloss}

\begin{figure}[h]
    \centering
    \includegraphics[width=\linewidth]{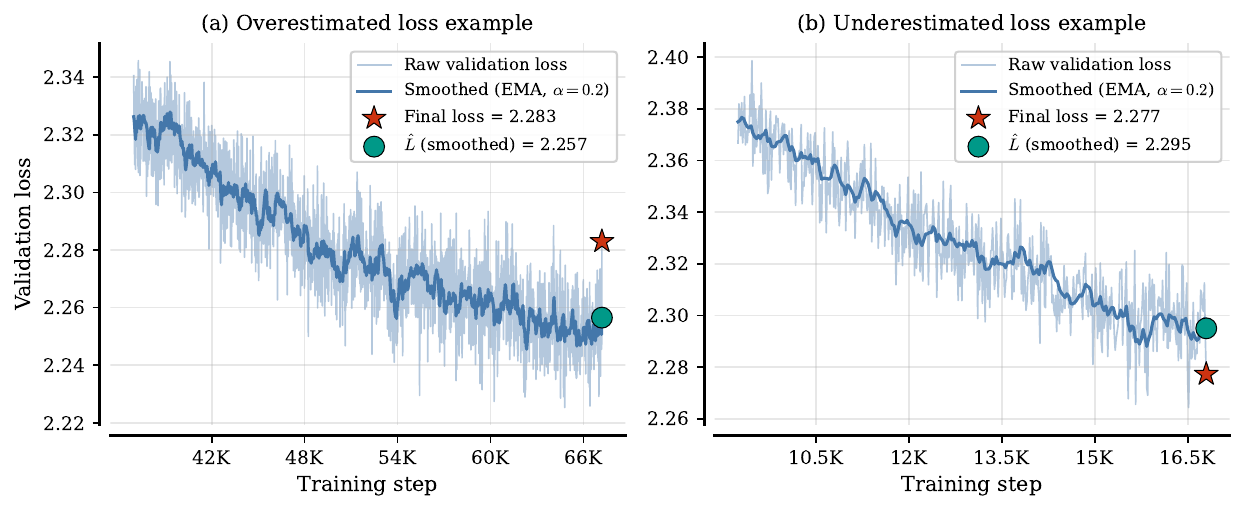}
    \caption{\textbf{Motivation for reporting a smoothed evaluation loss $\hat{L}$ rather than the final logged validation loss.}
    Each panel shows a real validation-loss trajectory near the end of training (raw points) together with the time-weighted EMA used in Sec.~\ref{sec:evalloss} (with $\alpha=0.2$ and filtering to synchronization boundaries at $H=30$).
    \textbf{Left:} an \emph{overestimated} final loss, where the last evaluated batch happens to be unusually difficult (the final logged loss exceeds $\hat{L}$).
    \textbf{Right:} an \emph{underestimated} final loss, where the last evaluated batch is unusually easy (the final logged loss is below $\hat{L}$).
    These examples illustrate that directly using the final recorded validation loss can introduce noise-driven bias in run comparisons and scaling-law fits, whereas $\hat{L}$ provides a more robust final performance estimate.}
    \label{fig:ema_motivation}
\end{figure}

Throughout this work, we evaluate each training run by computing a smoothed evaluation loss $\hat{L}$ from its validation loss trajectory. Directly using the final recorded validation loss is unreliable for comparing runs and fitting scaling laws, as validation losses are logged at discrete intervals and exhibit non-negligible noise, overestimating or underestimating final performance (see Figure~\ref{fig:ema_motivation}). This is likely caused by sampling different final evaluations depending on the number of training iterations for the run in question.  We therefore adopt a time-weighted exponential moving average (EMA) that accounts for the temporal spacing between measurements and provides a smooth, robust estimate of the final performance.

Concretely, each run records a validation loss trajectory $\{(\ell_i, t_i)\}_{i=1}^{N}$, where $\ell_i$ is the validation loss (cross-entropy on 2M token batches from the NemotronCC held-out split) measured at training step $t_i$. Since DiLoCo and MuLoCo synchronize worker parameters every $H$ inner steps, we first filter the trajectory to retain only measurements taken at synchronization boundaries, i.e., steps $t_i$ satisfying $t_i \bmod H = 0$ (with $H=30$ throughout this work). This filtering ensures that all methods---including data-parallel baselines---are evaluated at steps where the model parameters correspond to a post-synchronization state rather than an intermediate local state. For data-parallel methods, which effectively synchronize every step, this subsampling still produces a dense set of measurements.

After filtering, we obtain a subsequence $\{(\ell_j, t_j)\}_{j=1}^{M}$ and apply a time-weighted EMA. Unlike a standard EMA that assigns a fixed weight $\alpha$ to each new observation, we adjust the effective smoothing factor based on the elapsed time $\Delta t_j = t_j - t_{j-1}$ between consecutive measurements. The smoothed sequence $\{s_j\}_{j=1}^{M}$ is computed as
\begin{equation}
    s_1 = \ell_1, \qquad s_j = \tilde{\alpha}_j \, \ell_j + (1 - \tilde{\alpha}_j) \, s_{j-1}, \quad j = 2, \ldots, M,
\end{equation}
where the adaptive smoothing coefficient is
\begin{equation}\label{eq:ema_alpha}
    \tilde{\alpha}_j = 1 - \exp\!\Bigl(-\frac{\alpha \, \Delta t_j}{H}\Bigr).
\end{equation}
Here $\alpha > 0$ is a base smoothing parameter. When measurements are evenly spaced at every $H$ steps ($\Delta t_j = H$), the adaptive coefficient reduces to the constant $\tilde{\alpha} = 1 - e^{-\alpha}$. The evaluation loss for the run is then $\hat{L} = s_M$, the final value of the smoothed sequence.

We use $\alpha = 0.2$ throughout, which yields $\tilde{\alpha} \approx 0.181$ at the nominal spacing of $H=30$ steps. This provides substantial smoothing (an effective window spanning roughly the last $1/\tilde{\alpha} \approx 5$--$6$ synchronization intervals) while remaining responsive enough to track the loss trajectory near convergence. All scaling law fits, hyperparameter comparisons, and optimal run selections reported in this paper use this smoothed evaluation loss $\hat{L}$.

\end{document}